\documentclass{article}

\usepackage{microtype}
\usepackage{graphicx}
\usepackage{subfigure}
\usepackage{booktabs} %

\usepackage{hyperref}

\usepackage[preprint]{arxiv2023}

\usepackage{amsmath}
\usepackage{amssymb}
\usepackage{mathtools}
\usepackage{amsthm}

\usepackage[capitalize,noabbrev]{cleveref}

\theoremstyle{plain}

\theoremstyle{definition}

\theoremstyle{remark}

\usepackage[textsize=tiny]{todonotes}
\usepackage[T1]{fontenc}
\usepackage{amsfonts}
\usepackage{pifont}
\usepackage{multirow}
\usepackage{colortbl}
\usepackage{nameref}
\usepackage{bm}
\usepackage{enumitem}
\usepackage{threeparttable}

\def\eqref#1{equation~\ref{#1}}
\def\Eqref#1{Equation~\ref{#1}}

\def\1{\bm{1}}

\def\vs{{\bm{s}}}

\def\mI{{\bm{I}}}

\DeclareMathAlphabet{\mathsfit}{\encodingdefault}{\sfdefault}{m}{sl}
\SetMathAlphabet{\mathsfit}{bold}{\encodingdefault}{\sfdefault}{bx}{n}

\newcommand{\blue}[1]{#1}

\newcommand{\e}[1]{{\small $#1$}}
\newcommand{\algname}{\textsc{CoPaint }}
\newcommand{\algnamens}{\textsc{CoPaint}}

\icmltitlerunning{Towards Coherent Image Inpainting Using Denoising Diffusion Implicit Models}

\begin{document}

\twocolumn[
\icmltitle{Towards Coherent Image Inpainting Using Denoising Diffusion Implicit Models
}

\icmlsetsymbol{equal}{*}

\begin{icmlauthorlist}
\icmlauthor{Guanhua Zhang}{equal,ucsb}
\icmlauthor{Jiabao Ji}{equal,ucsb}
\icmlauthor{Yang Zhang}{ibm}
\icmlauthor{Mo Yu}{ibm2}
\icmlauthor{Tommi Jaakkola}{mit}
\icmlauthor{Shiyu Chang}{ucsb}
\end{icmlauthorlist}

\icmlaffiliation{ucsb}{UC Santa Barbara}
\icmlaffiliation{ibm}{MIT-IBM Watson AI Lab}
\icmlaffiliation{ibm2}{IBM Research during the project’s involvement}
\icmlaffiliation{mit}{MIT CSAIL}

\icmlcorrespondingauthor{Guanhua Zhang}{guanhua@ucsb.edu}
\icmlcorrespondingauthor{Jiabao Ji}{jiabaoji@ucsb.edu}

\icmlkeywords{Machine Learning, ICML}

\vskip 0.3in
]

\printAffiliationsAndNotice{\icmlEqualContribution} %

\begin{abstract}
Image inpainting refers to the task of generating a complete, natural image based on a partially revealed reference image. Recently, many research interests have been focused on addressing this problem using fixed diffusion models. These approaches typically directly replace the revealed region of the intermediate or final generated images with that of the reference image or its variants. However, since the unrevealed regions are not directly modified to match the context, it results in \emph{incoherence} between revealed and unrevealed regions. To address the incoherence problem, a small number of methods introduce a rigorous Bayesian framework, but they tend to introduce mismatches between the generated and the reference images due to the approximation errors in computing the posterior distributions. In this paper, we propose \algnamens, which can coherently inpaint the whole image without introducing mismatches. \algname also uses the Bayesian framework to jointly modify both revealed and unrevealed regions, but approximates the posterior distribution in a way that allows the errors to gradually drop to zero throughout the denoising steps, thus strongly penalizing any mismatches with the reference image. Our experiments verify that \algname can outperform the existing diffusion-based methods under both objective and subjective metrics. The codes are available at \url{https://github.com/UCSB-NLP-Chang/CoPaint/}.

\end{abstract}

\section{Introduction}
Image inpainting refers to the problem of generating a natural, complete image based on a partially revealed reference image. In recent years, researchers have increasingly focused on using diffusion models, a class of generative models that convert noise images into natural images through a series of denoising steps, to solve this problem. One popular approach is to use a fixed, generic diffusion model that has been pre-trained for image generation. This eliminates the need for retraining the diffusion model, making the process more efficient and versatile.

\begin{figure}[t]
    \begin{tabular}{cccc}
        \hspace{-1.5mm}
        \includegraphics[width=0.23\linewidth]{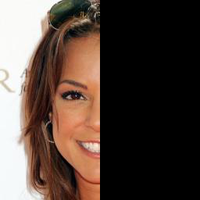} \hspace{-5mm} & 
        \includegraphics[width=0.23\linewidth]{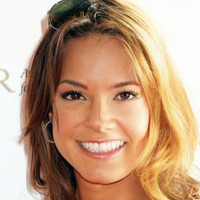} \hspace{-5mm} & 
        \includegraphics[width=0.23\linewidth]{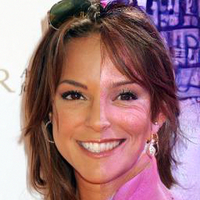} \hspace{-5mm} & 
        \includegraphics[width=0.23\linewidth]{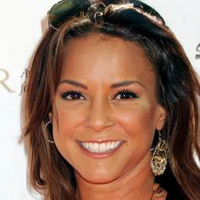} \\
        {{\footnotesize (a) \text{Input}}} & 
        {{\footnotesize(b) \textsc{Blended}}} &
        {{\footnotesize (c) \textsc{Ddrm}}} & 
        {{\footnotesize (d) \textsc{CoPaint}}}
    \end{tabular}
    \caption{{\footnotesize Inpainted images by \textsc{Blended} (b), \textsc{Ddrm} (c) and our proposed method \textsc{CoPaint-TT} (d). Image are generated conditioned on the given masked input (a) with a fixed diffusion model.
    }}
    \label{fig:intro-coherence}
\end{figure}

However, despite their promising performance, such methods are susceptible to the \emph{incoherence problem}. Specifically, these methods often impose the inpainting constraints based on some form of replacement operations, \emph{e.g.}, directly replacing the revealed portion of the predicted image with that of the reference image~\cite{ddnm}, or replacing the revealed portion of the intermediate denoising results with a corrupted version of the reference images~\cite{blended, repaint}. Yet the pixels of the unrevealed region, which should also be modified to match the context of the revealed region, are not directly modified~\cite{trippeDiffusion2022}. 
As a result, these methods can easily lead to discontinuity or incoherence between the revealed and unrevealed regions in the generated images.  For example, Figure~\ref{fig:intro-coherence} shows some incoherent inpainting results of a half-masked portrait image. The result in (b) has unmatched hair colors and styles between the left and right halves, and the result in (c) has a clear discontinuity in the middle resulting from different skin tones.

To address the incoherence problem, a small number of methods like \textsc{Dps}~\citep{dps} and \textsc{Resampling}~\citep{trippeDiffusion2022} use a more rigorous Bayesian framework, which casts the inpainting problem as sampling the images from the posterior distributions, conditional on the inpainting constraint. Since the posterior distribution differs from the prior distribution in both the revealed and unrevealed pixels, these methods can ensure that the entire image is coherently modified. However, since the posterior distribution is often very hard to compute, these methods would resort to approximations or Monte Carlo methods, which would introduce errors in satisfying the inpainting constraints. In short, it remains an unresolved problem how to ensure coherence during generation while strictly enforcing inpainting constraints.

In this paper, we propose \algnamens, a simple inpainting algorithm that addresses the incoherence problem without violating the inpainting constraints. \algname also adopts the Bayesian framework to coherently modify the entire images but introduces a new solution to address the challenges in computing and sampling from the posterior distribution. Specifically, \algname derives an approximated posterior distribution for the intermediate images, whose maximum a posteriori (MAP) samples become equivalent to directly minimizing the errors in the inpainting constraint, referred to as the \emph{inpainting errors}.  To make the computation of the inpainting errors tractable at each intermediate denoising step, we use the one-step estimation of the final generated image instead of directly computing the final generation. Although this would introduce further approximation errors, we can show that the errors would gradually decrease as the denoising process proceeds. Notably, at the final step, all the approximation errors can be made zero. 

Our experimental evaluations on \texttt{CelebA-HQ}  and \texttt{ImageNet} with various shapes of the revealed region verify that \algname has better inpainting quality and coherence than existing diffusion-model-based approaches under both objective and subjective metrics. For example, \algname achieves an average of 19\% relative reduction in \texttt{LPIPS} compared to \textsc{RePaint} \cite{repaint}, our most competitive baseline, while consuming 31\% less computation budget on \texttt{ImageNet} dataset.

\section{Related Work}
\label{sec:related_work}
Image inpainting is a long-lasting research question in computer vision, aiming at completing a degraded image naturally and coherently \citep{Xiang2022DeepLF,Shah2022OverviewOI}.
In recent years, various deep learning techniques have been suggested for the task of inpainting~\citep{Reddy2022ExplorationOI}, with a majority of them built upon auto-encoder~\citep{pathak2016context, Vo2018StructuralI, liu2018image, Iizuka2017GloballyAL, Song2018SPGNetSP, Guo2019ProgressiveII, Xiao2018DeepIG, Hong2019DeepFN, Nazeri2019EdgeConnectSG,Liu2020CorrectionTR}, VAE~\citep{Zheng2019PluralisticIC,Zhao2020UCTGANDI,Zhao2021LargeSI,Peng2021GeneratingDS}, GAN~\citep{pathak2016context, Vo2018StructuralI, liu2018image, Iizuka2017GloballyAL, Song2018SPGNetSP, Guo2019ProgressiveII, Xiao2018DeepIG, Hong2019DeepFN,Weng2022ASO} or auto-regressive transformer~\citep{Yu2021DiverseII,Wan2021HighFidelityPI} structures.
Despite achieving notable successes in inpainting,  these methods are primarily based on supervised learning, \emph{i.e.}, the networks require to be trained on specific degradation types. 
As a result, these approaches require large computational resources and may not be well-suited for scenarios that were not encountered during training, leading to poor generalization performance~\citep{Xiang2022DeepLF}.
More recently, diffusion model-based approaches are gaining increasing popularity due to their exceptional results in image generation \citep{sohl-dicksteinDeep2015,hoDenoising,yangDiffusion2022,BondTaylor2021UnleashingTP, Chung2022ImprovingDM, Batzolis2022NonUniformDM, Bansal2022ColdDI, Liu2022DelvingGI, Ku2022IntelligentPP, Benton2022FromDD, Horwitz2022ConffusionCI, Horita2022ASD, Li2022SDMSD}.
Besides, these methods enjoy the advantage of being able to perform inpainting without the need for degradation-specific training~\citep{Song2019GenerativeMB}.
In this section, we will review the current literature on diffusion-based inpainting. These methods can broadly be divided into two categories: supervised and unsupervised methods~\citep{ddrm}.

\paragraph{Supervised diffusion inpainting}
Supervised diffusion inpainting approaches involve training a diffusion model for the specific task of inpainting, taking into account the particular degradation types.
\textsc{Palette}~\citep{sahariaPalette2022,Saharia2021ImageSV} feeds the degraded image to the diffusion model at each time step of the diffusion process for training a diffusion inpainting model. 
Similar methods are also used by \textsc{Glide}~\citep{nicholGLIDE2022}, where a text-conditional diffusion model is fine-tuned for the inpainting task.
\textsc{Latent Diffusion}~\citep{rombachHighResolution2022} incorporates an autoencoding model for compressing the image space, and then the spatially aligned conditioning information is concatenated with the input of the model.
By contrast, \textsc{Ccdf}~\citep{chungComeCloserDiffuseFaster2022} adopts a non-expansive mapping for aggregating the degradation operation during training.
A ``predict-and-refine'' conditional diffusion model is proposed by \citet{whangDeblurring2021}, where a diffusion model is trained to refine the output of a deterministic predictor.
However, all these methods require degradation-specific training, which could be computationally expensive and may not generalize well to unseen degradation operators.

\paragraph{Unsupervised diffusion inpainting}
Different from supervised methods, unsupervised diffusion inpainting aims at utilizing pre-trained diffusion models for the inpainting task without any model modification.
Our proposed method also falls into this category.
As an early work, \citet{Song2019GenerativeMB} proposes to modify the DDPM sampling process by spatially blending the noisy version of the degraded image in each time step of the denoising process.
A similar idea is adopted by \textsc{BlendedDiffusion} for text-driven inpainting~\citep{blended}.
\textsc{Ddrm}~\citep{ddrm} defines a new posterior diffusion process whose marginal probability is proved to be consistent with DDPM~\citep{hoDenoising}.
Roughly speaking, the proposed denoising process is equivalent to blending the degraded image in a weighted-sum manner in each time step.
Despite the high efficiency of these methods, the images generated by the simple blending-based methods are often not harmonizing in the recovered part~\citep{repaint}.

To address the issue, the authors of \textsc{RePaint}~\citep{repaint} proposed a resampling strategy.
Specifically, a ``time travel'' operation is introduced, where images from the current \e{t} time step are first blended with the noisy version of the degraded image, and then used to generate images in the \e{t+1} time step using a one-step forward process, thereby reducing the visual inconsistency caused by blending.
\citet{trippeDiffusion2022} further proves that a simple blending-based method would introduce irreducible approximation error in the generation process.
A particle filtering-based method, named \textsc{Resampling}, is then proposed, where for time step \e{t}, each generated image is resampled based on its probability of generating the revealed part of the degraded image in the \e{t-1} time step.
\citet{pokleDeep2022} look at diffusion models in a deep equilibrium (DEQ) perspective and propose a DEQ method for inverting DDIM to save memory consumption.
\textsc{Ddnm}~\citep{ddnm} 
introduces a new blending mechanism, where the degraded image is directly incorporated in each time step without noise.
Another recent work \textsc{Dps}~\citep{dps} addresses the inpainting problem via approximation of the posterior sampling in a similar manner with classifier-free guided diffusion~\citep{dhariwalDiffusion2021}. 
Specifically, they use the approximated gradient of the posterior likelihood as a mean shift for images generated at each time step of the denoising process. 
Different from these methods, we introduce a Bayesian framework to jointly modify both revealed and unrevealed parts of images by maximizing the posterior in each time step along the denoising process and thus enjoying better coherence for the inpainted part.

\vspace{-1mm}
\section{Background and Notations}
\label{sec:background}
In this section, we will provide a brief overview of the diffusion model frameworks and notations that will be used in this paper. Note that we will only cover just enough details for the purpose of explaining our proposed approach. We would recommend readers refer to the original papers cited for complete details and derivations.

Denote \e{\bm X_0} as a random vector of the natural images (vectorized). DDIMs \cite{songDenoising2022} 
try to recover the distribution of \e{\bm X_0} through a set of intermediate variables, \emph{e.g.,} \e{\bm X_{1:T}}, which are progressively corrupted versions of \e{\bm X_0}. There are two processes in a DDIM framework, a \emph{forward diffusion process}, which defines how \e{\bm X_0} is corrupted into \e{\bm X_T}, and a \emph{reverse denoising process}, which governs how to recover \e{\bm X_0} from \e{\bm X_T} based on the forward process.

The forward diffusion process of DDIMs follows that of the denoising diffusion probabilistic models (DDPMs)~\citep{hoDenoising, sohl-dicksteinDeep2015}, which is a Markov process that progressively adds Gaussian noises to the intermediate variables, \emph{i.e.,}
\begin{equation}
    \small
    \begin{aligned}
      & q(\bm X_{1:T} | \bm X_0) = \prod_{t=1}^T q(\bm X_t|\bm X_{t-1}),\\
      & q(\bm X_t|\bm X_{t-1}) = \mathcal{N}(\bm X_t; \sqrt{\alpha_t} \bm X_{t-1}, \beta_t \mI),
    \end{aligned}
\end{equation}
where \e{\alpha_{1:T}} and \e{\beta_{1:T}} define the scaling and variance schedule with \e{\alpha_t=1-\beta_t}. It can be easily shown that, with an appropriate scaling and variance schedule and a sufficiently large \e{T}, \e{\bm X_T} approaches the standard Gaussian distribution.

For the reverse diffusion process, DDIMs introduce another distribution \e{q_\sigma}, called the inference distribution, that has a matching conditional distribution of each individual intermediate variable to \e{q}. Specifically 
\begin{equation}
    \small
    \begin{aligned}
        & q_\sigma(\bm X_{1:T} | \bm X_0) = q_\sigma(\bm X_T | \bm X_0) \prod_{t=T}^2 q_\sigma(\bm X_{t-1} | \bm X_t, \bm X_0), \\
        & q_\sigma(\bm X_T| \bm X_0) = \mathcal{N}(\bm X_T; \sqrt{\bar{\alpha}_T} \bm{X}_0, (1-\bar{\alpha}_T)\bm I), \\
        & q_\sigma(\bm X_{t-1} | \bm X_t, \bm X_0) = \mathcal{N}(\bm X_{t-1}; \bm \mu_t, \sigma_t^2 \bm I),
        \label{eq:q_sigma}
    \end{aligned}
\end{equation}
where \e{\bar{\alpha}=\prod^t_{i=1}\alpha_i} and \e{\sigma_t^2} is a free hyperparameter, and
\begin{equation}
    \small
    \bm \mu_t = \sqrt{\bar{\alpha}_{t-1}} \bm X_0 + \sqrt{1-\bar{\alpha}_{t-1} - \sigma_t^2} \frac{\bm X_t - \sqrt{\bar{\alpha}_t} \bm X_0}{\sqrt{1-\bar{\alpha}_t}}.
\end{equation}
It can be shown that as long as \e{\sigma_t^2 \in [0, 1-\bar{\alpha}_t], \forall t}. \e{q_{\sigma}} and \e{q} have matching distributions: \e{q_\sigma(\bm X_t | \bm X_0) = q(\bm X_t | \bm X_0), \forall t}.

The denoising process is derived from \e{q_\sigma} by replacing \e{\bm X_0} with an estimated value of \e{\bm X_0}, \emph{i.e.,} 
\begin{equation}
\small
\begin{split}
    p_{\theta}(\bm X_T) &= \mathcal{N}(\bm X_T; \bm 0, \bm I)\\ p_{\theta}(\bm X_{t-1} | \bm X_t) &= q_\sigma(\bm X_{t-1} | \bm X_t, \hat{\bm X}^{(t)}_0),
\end{split}
\label{eq:p_theta}
\end{equation}
where 
\begin{equation}
\small
    \hat{\bm X}^{(t)}_0 = \bm f_\theta^{(t)} (\bm X_t)
    \label{eq:fast_gen}
\end{equation}
is produced by a (reparameterized) neural network that predicts \e{\bm X_0} from \e{\bm X_t} by minimizing the mean squared error.

\begin{figure}
    \centering
    \includegraphics[width=0.95\linewidth]{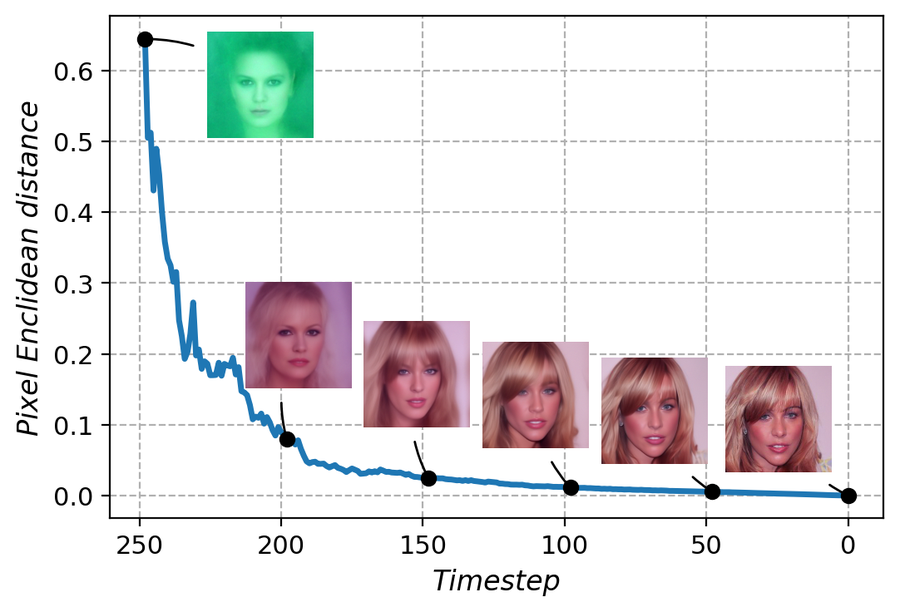}
    \caption{The trajectory of the gap between \e{\bm f_\theta^{(t)} (\tilde{\bm X}_t)} and \e{\tilde{\bm X}_0} along the unconditional diffusion denoising process. We report the pixel-wise averaged Euclidean distance between the two.  }
    \label{fig:mse_traj}
\end{figure}

\Eqref{eq:fast_gen} provides a way of estimating the final generation as a deterministic function of \e{\tilde{\bm X}_t}. In particular, \e{\bm f_\theta^{(t)} (\tilde{\bm X}_t)} is generated by feeding to the inference network once, and thus can be regarded as a compute-efficient approximation of the final generation. We will refer to it as \emph{one-step generation}. As shown in Figure~\ref{fig:mse_traj}
, the gap between \e{\bm f_\theta^{(t)} (\tilde{\bm X}_t)} and \e{\tilde{\bm X}_0} typically gets smaller as \e{t} gets smaller. As we will show, one-step generation is central to our algorithm because it permits direct control over the final generation through the intermediate variables.

\section{The \algname Algorithm} \label{sec:method}

\subsection{Problem Formulation}
The image inpainting problem aims to generate a natural, complete image given a partially revealed image, such that the generated image is identical to the given image in the revealed regions. Formally, denote \e{\bm r(\cdot)} as an operator that outputs a revealed subset of the input dimensions, and \e{\bm s_0} as the revealed portion of the given reference image. Then the goal of image inpainting is to generate a natural image under the following inpainting constraint
\begin{equation}
\small
    \mathcal{C}: \bm r(\tilde{\bm X}_0) = \bm s_0,
    \label{eq:constraint}
\end{equation}
which we denote as event \e{\mathcal{C}} for notation brevity. In this paper, we focus on the scenario where the diffusion model is pretrained and fixed, \emph{i.e.,} \e{\bm f_\theta^{(t)}} is fixed for all \e{t}.

As discussed, many existing diffusion-model-based approaches only replace the revealed region of the generated intermediate or final images \emph{i.e.,} \e{\bm r(\bm X_t)} or \e{\bm r(\bm X_0)}, 
to directly impose the inpainting constraint, whereas the generation of the remaining unrevealed region is not directly modified to match the context. Thus the resulting generated images could easily suffer from incoherence between the revealed and unrevealed regions. In the following, we will explain how we propose to jointly optimize both regions.

\subsection{A Prototype Approach}
\label{subsec:prototype}

We will start with a prototype approach. Consider the simplest form of DDIM, where \e{\sigma_t = 0, \forall t}. In other words, the denoising process becomes a deterministic process with respect to \e{\tilde{\bm X}_T}. As a result, the inpainting constraint on \e{\tilde{\bm X}_0} in \Eqref{eq:constraint} can translate to that on \e{\tilde{\bm X}_T}, so the image inpainting problem boils down to determining an appropriate \e{\tilde{\bm X}_T} based on the following posterior distribution:
\begin{equation}
    \small
    \begin{aligned}
        p_\theta\big(\tilde{\bm X}_T \big| \mathcal{C} \big) 
        \propto& p_\theta (\tilde{\bm X}_T) \cdot p_\theta \big(\bm r(\tilde{\bm X}_0) = \bm s_0 \big| \tilde{\bm X}_T \big) \\
        =& p_\theta (\tilde{\bm X}_T) \cdot \delta\big(\bm r(\tilde{\bm X}_0) = \bm s_0\big).
    \end{aligned}
    \label{eq:XT_posterior}
\end{equation}
According to Equations~\ref{eq:q_sigma} and \ref{eq:p_theta},  \e{p_\theta (\tilde{\bm X}_T)} is a standard Gaussian distribution. To clarify, \e{p_\theta \big(\bm r(\tilde{\bm X}_0) = \bm s_0 \big| \tilde{\bm X}_T \big)} denotes the probability density function of \e{\bm r(\tilde{\bm X}_0)} evaluated at \e{\bm s_0}, conditional on the value of \e{\tilde{\bm X}_T}. Since \e{\tilde{\bm X}_T} is given and \e{\tilde{\bm X}_0} is a deterministic function of \e{\tilde{\bm X}_T}, \e{p_\theta \big(\bm r(\tilde{\bm X}_0) = \bm s_0 \big| \tilde{\bm X}_T \big)} becomes a dirac delta function \e{\delta(\cdot)}, with infinity probability density at where the event holds, and zero density elsewhere. The dirac delta function can be approximated by a Gaussian density function with zero variance. Therefore, \Eqref{eq:XT_posterior}, after taking the logarithm, can be approximated as
\begin{equation}
\small
    \begin{aligned}
        &\log p_\theta(\tilde{\bm X}_T | \mathcal{C} ) \\
        \approx & - \frac{1}{2}\Vert \tilde{\bm X}_T \Vert_2^2 -\frac{1}{2\xi_T^2} \big\Vert \bm s_0 - \bm r (\tilde{\bm X}_0) \big\Vert_2^2 + C \\
        \approx & - \frac{1}{2}\Vert \tilde{\bm X}_T \Vert_2^2 -\frac{1}{2\xi_T^2} \big\Vert \bm s_0 - \bm r (\bm g_\theta(\tilde{\bm X}_T)) \big\Vert_2^2 + C,
    \end{aligned}
    \label{eq:XT_posterior_simp}
\end{equation}
where we denote \e{\tilde{\bm X}_0 = \bm g_\theta(\tilde{\bm X}_T)} to emphasize \e{\tilde{\bm X}_0} is a function of \e{\tilde{\bm X}_T}; \e{C} is the normalizing constant; \e{\xi_T} is the standard deviation of the second Gaussian distribution. When \e{\xi_T} approaches zero, the approximation in \Eqref{eq:XT_posterior_simp} becomes exact. In practice, \e{\xi_T} can be set to a very small value. 

\Eqref{eq:XT_posterior_simp} provides a justification for solving \e{\tilde{\bm X}_T} using optimization method, because the first term can be regarded as a prior regularization and the second term as a penalty term enforcing the inpainting constraint. One can either perform gradient ascent over \e{\tilde{\bm X}_T} to find the maximum a posteriori (MAP) estimate of \e{\tilde{\bm X}_T}, or apply gradient-based sampling techniques such as Hamiltonian Markov Chain Monte Carlo (MCMC) \citep{Neal2011MCMCUH}.
to draw random samples. Note that the optimization is over the entire \e{\tilde{\bm X}_T}, not just the revealed regions, so this would ideally resolve the incoherence problem in the existing replacement methods. Since the weight on the second term is very large, we can expect to solve for an \e{\tilde{\bm X}_T} that can satisfy the inpainting constraint very well.

\subsection{One-Step Approximation}
\label{subsec:one-step}

The key limitation of the aforementioned prototype approach is that it is computationally impractical, because evaluating the final generation \e{\bm g_\theta(\tilde{\bm X}_T)} and computing its gradient involve performing forward and reverse propagation through the entire DDIM denoising process, which typically consists of tens or even hundreds of denoising steps. We thus need to derive a computationally-feasible algorithm from the prototype approach.

As discussed in Section~\ref{sec:background}, the one-step generation \e{\bm f_\theta^{(T)}(\tilde{\bm X}_T)} offers a fast approximation of the final generation, so a straightforward modification is to replace the \e{\bm g_\theta(\tilde{\bm X}_T)} in \Eqref{eq:XT_posterior_simp} with \e{\bm f_\theta^{(T)}(\tilde{\bm X}_T)}. 

Formally, we introduce a approximated conditional distribution of \e{\bm r(\tilde{\bm X}_0)} given \e{\tilde{\bm X}_T}, denoted as \e{p'_\theta(\bm r(\tilde{\bm X}_0) | \tilde{\bm X}_T)}, which is centered around the one-step generated value, \e{\bm r(\bm f_\theta^{(T)}(\tilde{\bm X}_T))}, plus a Gaussian error, \emph{i.e.},
\begin{equation}
\small
    p'_\theta(\bm r(\tilde{\bm X}_0) | \tilde{\bm X}_T) = \mathcal{N}\big(\bm r(\tilde{\bm X}_0); \bm r(\bm f_\theta^{(T)}(\tilde{\bm X}_T)), \xi^{'2}_T \bm I  \big),
    \label{eq:gauss_approx_T}
\end{equation}
where \e{\xi'_T} is the standard deviation parameter. Plugged in this approximated distribution, the approximate posterior is
\begin{equation}
\small
    \begin{aligned}
        &\log p'_\theta(\tilde{\bm X}_T | \mathcal{C} ) \\
        = & \log(p_\theta (\tilde{\bm X}_T)) + \log\big(p'_\theta \big(\bm r(\tilde{\bm X}_0) = \bm s_0 \big| \tilde{\bm X}_T \big)\big) + C'\\
        = & - \frac{1}{2}\Vert \tilde{\bm X}_T \Vert_2^2 -\frac{1}{2\xi_T^{'2}} \big\Vert \bm s_0 - \bm r (\bm f_\theta^{(T)}(\tilde{\bm X}_T)) \big\Vert_2^2 + C',
    \end{aligned}
    \label{eq:XT_posterior_approx}
\end{equation}
where \e{C'} refers to any normalizing constant, and the last line is derived from \Eqref{eq:gauss_approx_T}.

It can be easily shown that in order to minimize the approximation gap, \emph{i.e.,} the KL divergence between \e{p_\theta(\bm r(\tilde{\bm X}_0) | \tilde{\bm X}_T)} and \e{p'_\theta(\bm r(\tilde{\bm X}_0) | \tilde{\bm X}_T)}, \e{\xi^{'2}_T} should be set to
\begin{equation}
    \small
    \xi_T^{'2} = \frac{1}{N}\mathbb{E}_{p_\theta} \big[\big\Vert \bm r(\bm f_\theta^{(T)}(\tilde{\bm X}_T)) -\bm r(\tilde{\bm X}_0)\big\Vert_2^2 \big], 
    \label{eq:xi}
\end{equation}
where \e{N} is the dimension of \e{\bm s_0}. Similar to \Eqref{eq:XT_posterior_simp}, maximizing \Eqref{eq:XT_posterior_approx} over \e{\tilde{\bm X}_T} is essentially trying to satisfy the (approximated) inpainting constraint (second term) regularized by its prior (first term). However, in contrast to the exact case in \Eqref{eq:XT_posterior_simp}, where \e{\xi_T} should be as small as possible, \e{\xi'_T} should be large enough (\Eqref{eq:xi}) to capture the approximation error, which leads to a smaller weight on the approximate inpainting constraint term in \Eqref{eq:XT_posterior_approx}.

\subsection{Denoising Successive Correction}
\renewcommand{\algorithmiccomment}[1]{// #1}
\begin{algorithm}[tb]
\small
   \caption{\textsc{CoPaint-TT}}
   \label{alg:copaint+tt}
\begin{algorithmic}[1]
   \STATE {\bfseries Input:} $\vs_0$, $\{f^{(t)}_\theta(\cdot)\}^{T}_{t=1}$, time travel interval $\tau$ and frequency $K$, gradient descent number $G$ and learning rate $\{\eta_t\}^{T}_{t=1}$
   \STATE Initialize $\tilde{\bm X}_{T} \sim \mathcal{N}(0, \mathbf{I})$
   \STATE $t \gets T$, $k \gets K$
   \WHILE{$t \neq 0$}
        \STATE Optimize $\tilde{\bm X}_{t}$ to maximize Equations \ref{eq:XT_posterior_approx} \ref{eq:Xt_posterior_approx} by $G$-step gradient descent with learning rate $\eta_t$
        \STATE Generate $\tilde{\bm X}_{t-1}$ with \Eqref{eq:p_theta}
        \STATE $t \gets t - 1$
        \IF{$t \text{ mod } \tau = 0$ and $t \leq T - \tau$} 
            \IF{ $k > 0$ }
                \STATE{}\COMMENT{time trave l}
                \STATE{Generate $\tilde{\bm X}_{t+\tau}\sim q(\tilde{\bm{X}}_{t+\tau}|\tilde{\bm{X}}_{t})$ } 
                \STATE $t \gets t + \tau - 1$, $k \gets k - 1$ \\
            \ELSE
                \STATE $k \gets K$
            \ENDIF
        \ENDIF
   \ENDWHILE \\
   \STATE {\bfseries Return:} $\Tilde{\bm X}_0$
\end{algorithmic}
\end{algorithm}

Equation~\ref{eq:XT_posterior_approx} will push revealed part of the one-step approximated generation, \e{\bm r(\bm f_\theta^{(T)}(\tilde{\bm X}_T))}, towards the reference image \e{\bm s_0}. However, the actual inpainting constraint requires us to push the actual final generation, \e{\bm r(\tilde{\bm X}_0)}, to \e{\bm s_0}. As a result, optimizing Equation~\ref{eq:XT_posterior_approx} cannot exactly satisfy the inpainting constraint. To further enforce the inpainting constraint, we return to the \emph{non-deterministic} DDIM procedure, where \e{\sigma_t \neq 0}, and apply the optimization technique discussed in Sections~\ref{subsec:prototype} and \ref{subsec:one-step} to all the intermediate variables to successively correct the approximation error.

The proposed DDIM procedure samples \e{\tilde{\bm X}_{0:T}} from the approximate posterior \e{p'_\theta(\tilde{\bm X}_{0:T} | \mathcal{C})}, which is decomposed as
\begin{equation}
    \small
    p'_\theta(\tilde{\bm X}_{0:T} | \mathcal{C}) = p'_\theta(\tilde{\bm X}_T | \mathcal{C}) \prod_{t=1}^T p'_\theta(\tilde{\bm X}_{t-1} | \tilde{\bm X}_t, \mathcal{C}).
    \label{eq:markov}
\end{equation}
\e{p'_\theta(\tilde{\bm X}_T | \mathcal{C}_T)} is defined in \Eqref{eq:XT_posterior_approx}. To compute \e{p'_\theta(\tilde{\bm X}_{t-1} | \tilde{\bm X}_t, \mathcal{C})}, we introduce a set of Gaussian approximated distributions similar to \Eqref{eq:gauss_approx_T} as
\begin{equation}
\small
    p'_\theta(\bm r(\tilde{\bm X}_0) | \tilde{\bm X}_t) = \mathcal{N}\big(\bm r(\tilde{\bm X}_0); \bm r(\bm f_\theta^{(t)}(\tilde{\bm X}_t)), \xi^{'2}_t \bm I  \big),
    \label{eq:gauss_approx}
\end{equation}
where \e{\xi_{t}^2} is defined similar to \Eqref{eq:xi} (replacing \e{T} with \e{t}) to minimize the one-step approximation error. Then \e{p'_\theta(\tilde{\bm X}_{t-1} | \tilde{\bm X}_t, \mathcal{C})} can be computed as
\begin{small}
\begin{align}
        &\log p'_\theta(\tilde{\bm X}_{t-1} | \tilde{\bm X}_t, \mathcal{C}) \nonumber\\
        = &\log p_\theta(\tilde{\bm X}_{t-1} | \tilde{\bm X}_t) + \log p'_\theta\big(\bm r(\tilde{\bm X}_0) = \bm s_0| \tilde{\bm X}_{t-1}, \tilde{\bm X}_t \big) + C'\nonumber\\
        = &\log p_\theta(\tilde{\bm X}_{t-1} | \tilde{\bm X}_t) + \log p'_\theta\big(\bm r(\tilde{\bm X}_0) = \bm s_0 | \tilde{\bm X}_{t-1}\big) + C'\nonumber\\
        =& -\frac{1}{2\sigma_t^2} \Vert \tilde{\bm X}_{t-1} - \tilde{\bm \mu}_{t}\Vert_2^2 -\frac{1}{2\xi_{t-1}^{'2}} \big\Vert  \bm s_0 - \bm r \big(\bm f_\theta^{(t-1)}(\tilde{\bm X}_{t-1})\big) \big\Vert_2^2 \nonumber\\
        &+ C',
        \label{eq:Xt_posterior_approx}
\end{align} 
\end{small}

\noindent where the third line follows from the reverse Markov property of the DDIM denoising process. The first term in the last line follows from Equations~\ref{eq:q_sigma} to \ref{eq:p_theta}, with
\begin{small}
\begin{flalign}
&\resizebox{0.93\linewidth}{!}{
$\tilde{\bm \mu}_t = \sqrt{\alpha_{t-1}} f_\theta^{(t)}(\tilde{\bm X}_{t}) + \sqrt{1-\alpha_{t-1} - \sigma_t^2} \frac{\tilde{\bm X}_t - \sqrt{\alpha_t} f_\theta^{(t)}(\tilde{\bm X}_{t})}{\sqrt{1-\alpha_t}}.$}&&
\raisetag{13pt}
\end{flalign}
\end{small}

To generate the final inpainting result, we follow the following \emph{greedy} optimization procedure to find samples of \e{\tilde{\bm X}_{0:T}} that maximizes the \e{p'_\theta(\tilde{\bm X}_{0:T} | \mathcal{C})} in \Eqref{eq:markov}. \emph{First}, we sample an \e{\tilde{\bm X}_T} by optimizing \Eqref{eq:XT_posterior_approx}. \emph{Second}, given the generated value of \e{\tilde{\bm X}_t}, we sample an \e{\tilde{\bm X}_{t-1}} by optimizing \Eqref{eq:Xt_posterior_approx}.
Both steps are essentially enforcing the approximate inpainting constraints under the DDIM prior regularization. According to Figure~\ref{fig:mse_traj}, the one-step approximation error will gradually reduce as \e{t} decreases, so the algorithm would approach the inpainting constraint with increasing levels of exactness, successively correcting the approximation errors made in the previous steps. In particular, when \e{t = 1}, if we set \e{\sigma_1 = 0} and let \e{\xi_1} approach zero, we will have zero approximation error, \emph{i.e.} \e{\bm f^{(1)}_\theta(\tilde{\bm X}_1) = \tilde{\bm X}_0}, so the generated image can be made to satisfy the inpainting constraint with very small errors.

\subsection{Additional Algorithmic Designs}
\label{sec:method-additional}

Although our algorithm can eventually eliminate the one-step approximation error in the final denoising step, the error in the early denoising steps can still affect the generation quality because it affects the quality of the prior distribution for subsequent generations. We introduce additional optional designs to reduce the approximation error.

\paragraph{Multi-Step Approximation} In the early denoising steps where the approximation error is more significant, we can replace the one-step approximation with multi-step approximation, where \e{\tilde{\bm X}_0} is approximated by going through multiple deterministic denoising steps at a subset of time steps.

\paragraph{Time Travel} 
To improve the self-consistency of the intermediate examples, we can apply the time travel technique \cite{repaint,ddnm}, which periodically returns to the previous denoising steps by corrupting the intermediate images. 
Specifically, for a set of selected time steps \e{\phi} at denoising time step \e{T-\tau}, instead of progressing to \e{T-\tau-1}, we rewind to time \e{T-1} by sampling a new \e{\tilde{\bm X}_{T-1}} based on \e{q(\tilde{\bm X}_{T-1} | \tilde{\bm X}_{t-\tau})}, and repeat the denoising steps from there. After \e{K} rounds of rewinding and denoising through steps \e{T-1} to \e{T-\tau}, we then enter \e{K} rounds of rewinding and denoising loop through steps \e{T-\tau-1} to \e{T-2\tau}. This process progresses until time zero is reached.
The algorithm of \algname with time travel, abbreviated as \algnamens-TT, is shown in Algorithm~\ref{alg:copaint+tt}.

\section{Experiments}
\begin{table*}[h!]
\caption{\footnotesize Quantitative results on \texttt{CelebA-HQ}(\emph{top}) and \texttt{ImageNet} (\emph{bottom}). We report the objective metric \texttt{LPIPS} and subjective human \texttt{vote difference} score of each baseline compared with our method \textsc{CoPaint-TT}. 
Lower is better for both metrics. 
The \textit{vote difference} scores are calculated as the vote percentage of \textsc{CoPaint-TT} minus vote percentage of certain baseline.
We report the results of two human tests, \emph{i.e.}, \texttt{overall} and \texttt{coherence}, in the \texttt{Vote(\%)} column separated by $/$,  where \texttt{overall} is based on naturalness, restoration quality and coherence, while \texttt{coherence} is only based coherence of the generated image.
\texttt{vote difference} score being lower than zero indicates certain baseline is better than our method \textsc{CoPaint-TT}. \blue{Numbers marked in blue are additional results.}
}
\label{tab:main_results}
\resizebox{\textwidth}{!}{
\begin{threeparttable}
\begin{tabular}{@{}c|cccccccccccccc|cc@{}}
\toprule
\midrule
\multicolumn{17}{c}{\texttt{CelebA-HQ}}
\\ 
\midrule
\multirow{2}{*}{Method}
& \multicolumn{2}{c}{\textit{Expand}}
& \multicolumn{2}{c}{\textit{Half}}
& \multicolumn{2}{c}{\textit{Altern}}
& \multicolumn{2}{c}{\textit{S.R.}}
& \multicolumn{2}{c}{\textit{Narrow}}
& \multicolumn{2}{c}{\textit{Wide}} 
& \multicolumn{2}{c}{\textit{Text}}
& \multicolumn{2}{|c}{\bf Average}
\\
& \texttt{LPIPS}$\downarrow$ & \texttt{Vote(\%)}$\downarrow$  
& \texttt{LPIPS}$\downarrow$ & \texttt{Vote(\%)} $\downarrow$   
& \texttt{LPIPS}$\downarrow$ & \texttt{Vote(\%)} $\downarrow$  
& \texttt{LPIPS}$\downarrow$ & \texttt{Vote(\%)}  $\downarrow$  
& \texttt{LPIPS}$\downarrow$ & \texttt{Vote(\%)}  $\downarrow$  
& \texttt{LPIPS}$\downarrow$ & \texttt{Vote(\%)}  $\downarrow$  
& \texttt{LPIPS}$\downarrow$ & \texttt{Vote(\%)}  $\downarrow$  
& \texttt{LPIPS}$\downarrow$ & \texttt{Vote(\%)}  $\downarrow$  
\\ 
\midrule
\multicolumn{1}{c|}{\textsc{Blended}}   &   0.557   &   82/80    &   0.228   &   64/72    &   0.047   &   12/30    &   0.269   &   78/86    &   0.078   &   54/64    &   0.102   &   46/58    &   0.011   &   18/12 & 0.185 & 51/57
\\
\multicolumn{1}{c|}{\textsc{Ddrm}}   &   0.704   &   94/98    &   0.273   &   86/96    &   0.151   &   78/84    &   0.596   &   100/100    &   0.140   &   76/84    &   0.125   &   84/62   &   0.028   &  38/42  & 0.288 & 79/81
\\
\multicolumn{1}{c|}{\textsc{Resampling}}   &   0.536   &   60/66   &   0.231   &   68/88   &   0.050   &   24/46   &   0.261   &   64/72   &   0.077   &   50/64   &   0.102   &   40/50   &   0.013   &   \textbf{-12}/8  & 0.181 & 42/56
\\
\multicolumn{1}{c|}{\textsc{RePaint}}   &   0.496   &   24/18    &   0.199   &   2/12    &   \textbf{0.014}   &   \textbf{-32}/38    &   0.041   &   10/10    &   0.039   &   4/10    &   0.072   &   \textbf{-16/-32} &   \textbf{0.006}   &   4/\textbf{-14}  & 0.124 & \textbf{0}/6
\\
\multicolumn{1}{c|}{\textsc{Dps}}   &   \textbf{0.449}   &  \textbf{-16/-12}    &   0.261   &   28/32    &   0.166   &   58/72    &   0.182   &   60/82    &   0.160   &   72/52    &   0.181   &   30/28   &   0.152   &   58/60 & 0.222 & 41/45
\\
\multicolumn{1}{c|}{\textsc{Ddnm}}      &   0.598   &   76/94   &   0.257   &   84/72   &   0.015   &   -2/\textbf{-2}  &   0.046   &   6/\textbf{0}   &   0.071   &   14/38   &   0.111   &   28/60   &   0.014   &   \textbf{-12}/10 & 0.158 & 27/39
\\
\rowcolor{gray!20}
\multicolumn{1}{c|}{\textsc{CoPaint-Fast}}    &   \blue{0.483}   &   \blue{10/34}    &   \blue{0.203}   &   \blue{44/20}    &   \blue{0.057}   &   \blue{10/2}    &   \blue{0.084}   &   \blue{20/6}    &   \blue{0.068}   &   \blue{16/10}    &    \blue{0.096}   &   \blue{20/4}   &   \blue{0.036}   &  \blue{l4/-4} & \blue{0.147} & \blue{13/11}
\\
\rowcolor{gray!20}
\multicolumn{1}{c|}{\textsc{CoPaint}}    &   {0.472}   &   \blue{12/20}    &   {0.188}   &   \blue{40/24}    &   {0.016}   &   \blue{-6/-4}    &   {0.033}   &   \blue{22/-4}    &   {0.040}   &   \blue{20/14}    &    {0.071}   &   \blue{24/-2}   &   {0.007}   &  \blue{-12/-4} & 0.118 & \blue{15/6}
\\
\rowcolor{gray!20}
\multicolumn{1}{c|}{\textsc{CoPaint-TT}}    &   0.464   &   0/0    &   \textbf{0.180}   &   \textbf{0/0}    &   \textbf{0.014}   &   0/0    &   \textbf{0.028}   &   \textbf{0/0}    &   \textbf{0.037}   &   \textbf{0/0}    &    \textbf{0.069}   &   0/0   &   \textbf{0.006}   &   0/0 & \textbf{0.114} & \textbf{0/0}
\\
\midrule
\multicolumn{17}{c}{\texttt{ImageNet}}
\\
\midrule
\multicolumn{1}{c|}{\textsc{Blended}}   &  0.717  &  39/36  &  0.366  &  72/80  &  0.277  &  96/92  &  0.686  &  94/96  &  0.161  &  76/64  &  0.194  &  62/60  &  0.028  &  8/26 & 0.347 & 64/65
\\
\multicolumn{1}{c|}{\textsc{Ddrm}}   &  0.730  &  58/44  &  0.385  &  78/64  &  0.439  &  92/100  &  0.822 &  92/100  &  0.211  &  84/84  &  0.231  &  86/72  &  0.060  &  32/44 & 0.411 & 75/71
\\
\multicolumn{1}{c|}{\textsc{Resampling}}  &  0.704  &  38/40  &  0.353  &  58/86  &  0.259  &  72/88  &  0.624  &  94/98 &  0.151  &  66/64  &  0.183  &  76/66  &  0.028  &  22/26 & 0.329 & 61/67 
\\
\multicolumn{1}{c|}{\textsc{RePaint}}  &  0.706  &  36/36  &  0.323  &  4/24  &  0.103  &  50/22  &  0.209  &  70/66  &  \textbf{0.072}  &  32/2  &  0.156  &  24/36  &  0.014  &  22/18 & 0.226 & 34/29
\\
\multicolumn{1}{c|}{\textsc{Dps}}   &  0.673  &  38/44  &  0.512  &  82/72  &  0.474  &  100/100  & 0.511  &  96/95  &  0.447  &  94/98  &  0.468  &  96/92  &  0.438  &  92/96 & 0.503 & 87/86
\\
\multicolumn{1}{c|}{\textsc{Ddnm}}   &  0.805  &  34/76  &  0.408  &  68/64  &  0.051  &  12/12  &  0.107  &  18/36  &  0.101  &  50/70  &  0.185  &  48/60  &  \textbf{0.012}  &  \textbf{-2/-20} & 0.238 & 33/44
\\
\rowcolor{gray!20}
\multicolumn{1}{c|}{\textsc{CoPaint-Fast}}    &   \blue{0.678}   &   \blue{14/26}    &   \blue{0.335}   &   \blue{22/24}    &   \blue{0.075}   &   \blue{10/6}    &   \blue{0.128}   &   \blue{36/28}    &   \blue{0.103}   &   \blue{26/22}    &    \blue{0.167}   &   \blue{24/32}   &   \blue{0.043}   &  \blue{6/-2} & \blue{0.218} & \blue{15/19}
\\
\rowcolor{gray!20}
\multicolumn{1}{c|}{\textsc{CoPaint}}    &  0.640  &  \blue{-2/8}  &  0.307  &  \blue{6/0}  &  0.041  &  \blue{22/4}  &  \textbf{0.069}  &  \blue{20/18}  &  0.078  &  \blue{24/30}  &  0.138  &  \blue{14/16}  &  0.017  &  \blue{2/-10}  & 0.184 & \blue{12/9}
\\
\rowcolor{gray!20}
\multicolumn{1}{c|}{\textsc{CoPaint-TT}}   &  \textbf{0.636}  &  \textbf{0/0}  &  \textbf{0.294}  &  \textbf{0/0}  &  \textbf{0.039 } &  \textbf{0/0}  &  \textbf{0.069}  &  \textbf{0/0}  &  0.074  &  \textbf{0/0}  &  \textbf{0.133}  &  \textbf{0/0}  &  0.015  &  0/0 & \textbf{0.180} & \textbf{0/0}
\\
\toprule
\bottomrule
\end{tabular}
\end{threeparttable}
}
\end{table*}

\subsection{Experiment Setup} \label{exp:setup}
\paragraph{Datasets and models}
Following \citet{repaint}, we validate our method on two commonly used image datasets: \texttt{CelebA-HQ}~\citep{celeba} and \texttt{ImageNet-1K}~\citep{imagenet}. 
\texttt{CelebA-HQ} contains more than 200K celebrity images, and we use the data split provided by \citet{lama} following \citet{repaint}. 
\texttt{ImageNet-1K} is a large-scale image dataset containing 1000 categories, and the original data split is used~\citep{imagenet}. 
Since not all images in the datasets are square-shaped images that diffusion models accept, we crop all images into \e{256\times 256} size to accommodate pretrained diffusion models. 
For \texttt{CelebA-HQ} dataset, we use the diffusion model pretrained by~\citet{repaint}. For \texttt{ImageNet}, we use the model pretrained by \citet{guided}. 
We use the first five images in the validation set for hyperparameter selection. 
The first 100 images in test sets are used for evaluation following \citet{repaint}.
Following \citet{repaint,ddnm,lama}, we consider seven different degradation masks on the original images for recovering: 
\textit{Expand}, \textit{Half}, \textit{Altern}, \textit{S.R.}, \textit{Narrow}, \textit{Wide}, and \textit{Texts}.
Examples of the degraded images are in Figure~\ref{fig:qualitative_celebahq}. 

\paragraph{Metrics} \label{exp:metric}
We evaluate the quality of the inpainting results using both \emph{objective} and \emph{subjective} metrics. For the objective metric, we adopt the \texttt{LPIPS} used in ~\citet{repaint}, which computes the similarity of two images in the feature space of AlexNet~\cite{alexnet}. 
For each reference image, we generate two inpainted images and the overall average \texttt{LPIPS}  is reported.
For the subjective metrics, we conduct a human evaluation on Amazon MTurk, where each subject is presented with a masked reference image and a pair of inpainted images, one by \algnamens-TT and the other by one of the baselines. The subject is then asked to select which one is of better quality according to a set of prespecified criteria. We also introduce a third option, `cannot tell the difference', if the subject cannot find any noticeable differences between the pair.

We perform two tests where different criteria are specified. In the first test, referred to as \texttt{overall}, three criteria are introduced: 1) the inpainted image should be natural and without artifact; 2) the revealed portion should resemble the reference image; and 3) the image should be coherent. In the second test, referred to as \texttt{coherence}, only the coherence criterion is introduced. 
For both tests, we randomly sample 50 images for every mask in \texttt{CelebA-HQ} and \texttt{ImageNet} and thus result in \e{2 \times 2 \times 7 \times 50 = 1400} image pairs for comparison.
In each comparison with one baseline, we use the \texttt{vote difference (\%)}, which is the percentage of the votes for \textsc{CoPaint-TT} subtracted by that for the baseline, as the metric for the relative inpainting quality compared to the baseline.
More details about the human evaluation design could be seen in Appendix~\ref{app:user_study}.

\paragraph{Baselines and implementation details}
We focus on comparison with diffusion-model-based methods, which have been shown to achieve state-of-the-art performance over methods that do not use diffusion models~\citep{repaint}.
Specifically, the following baselines are introduced: 
\textsc{Blended}~\cite{Song2019GenerativeMB,blended},
\textsc{Ddrm}~\citep{ddrm},
\textsc{Resampling}~\citep{trippeDiffusion2022},
\textsc{RePaint}~\citep{repaint},
\textsc{Dps}~\citep{dps}, and
\textsc{Ddnm}~\citep{ddnm}.
A brief introduction about these baselines could be found in Section~\ref{sec:related_work}.

For all methods, we set the number of reverse sampling steps as 250 if not specified otherwise. 
For \textsc{RePaint}, we use their released codes\footnote{
With the released code of \textsc{RePaint} in \url{shorturl.at/AHILU} and the matching configurations, we noticed there is a slight gap between our implemented results and the reported ones in \citet{repaint}. Nevertheless, we believe our comparison with \textsc{RePaint} is fair because our methods were implemented based on the same code base, so any configuration nuances that can account for the gap are likely to affect the performance of our methods in the same direction.
}
out-of-the-shelf with exactly the same setting as reported in their paper~\citep{repaint}.
We then implement all other methods based on the \textsc{RePaint} code base and keep all hyper-parameters the same as the corresponding papers, details could be seen in Appendix~\ref{app:baseline}.
Specifically, we set gradient descent step number \e{G=2} for both \textsc{CoPaint} and \textsc{CoPaint-TT}. 
A time-efficient version of our method, \textsc{CoPaint-Fast} is further introduced with \e{G=1} and reverse sampling step number as 100.
We adopt an adaptive learning rate as \e{\eta_t=0.02\sqrt{\bar{\alpha}_t}} for all our methods.
The rationale for such a learning rate setting can be seen in Appendix~\ref{app:lr}.
For better efficiency, we simply set \e{\xi_t^{'2}=(1/1.012)^{T-t}} instead of calculating it, inspired by the empirical observation that \e{\{\xi_t\}} is increasing along \e{t} in Figure~\ref{fig:mse_traj}.
For \textsc{CoPaint-TT}, we use time travel interval \e{\tau=10} and travel frequency \e{K=1}.
The ablation studies for the hyper-parameters could be seen in Section~\ref{exp:ablation}.
Note that all methods use the same pretrained diffusion models without any modification.

\subsection{Experiment Results} \label{exp:results}
\paragraph{Quantitative results}
Table~\ref{tab:main_results} shows the quantitative results of the proposed \textsc{CoPaint-Fast}, \textsc{CoPaint} and \textsc{CoPaint-TT} together with all other baselines on both \texttt{CelebA-HQ}~(\emph{top}) and \texttt{ImageNet}~(\emph{bottom}) datasets with seven mask types. The results in the \texttt{Votes (\%)} column show the two vote difference scores, the first for \texttt{overall} test and the second for the \texttt{coherence} test.
Here are our key observations.
\underline{First}, in terms of the objective metric, \textsc{CoPaint} consistently outperforms the other baselines, and reduces the average \texttt{LPIPS} score by 5\% and 19\% beyond the best-performing baseline \textsc{RePaint} in \texttt{CelebA-HQ} and \texttt{ImageNet} dataset, respectively. 
\underline{Second}, when combined with time travel, \textsc{CoPaint-TT} can further bring down the average LPIPS score by another 3\% and 1\% in the two datasets, respectively.
Besides, \textsc{CoPaint-TT} achieves the best performance among eleven out of the fourteen inpainting tasks while achieving comparable performances with the best baseline in the rest.
\underline{Third}, in terms of subjective evaluations, \textsc{CoPaint-TT} consistently produces positive \texttt{vote difference} scores in both the \texttt{overall} and \texttt{coherence} tests in most of the comparisons, indicating that the images generated by our method are not only more coherent, but also considered superior in terms of other aspects as well, including naturalness and meeting the inpainting constraint.
Also, notice that the performance advantage of \textsc{CoPaint-TT} is generally more significant on \texttt{ImageNet}, which may be because images in \texttt{ImageNet} are more complex and thus any imperfections in the images, including incoherence, would be more conspicuous.

We further conduct an additional experiment on inpainting high-resolution images in Appendix~\ref{app:high-resolution}, where our methods still achieve the best performance compared with other baselines with competitive time efficiency. 
Besides, the proposed method could also be used in other image restoration tasks.
An additional experiment on the super-resolution task could be seen in Appendix~\ref{app:super-resolution}, where our methods show consistent superiority over other baselines.

\begin{figure}[t]
    \centering
    \includegraphics[width=0.48\linewidth]{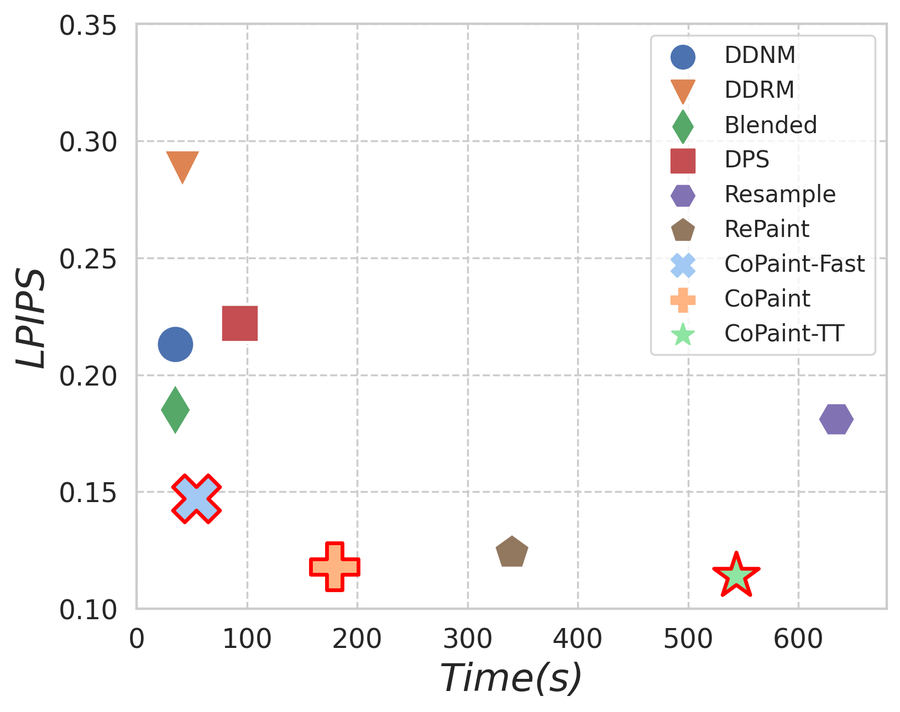}
    \includegraphics[width=0.48\linewidth]{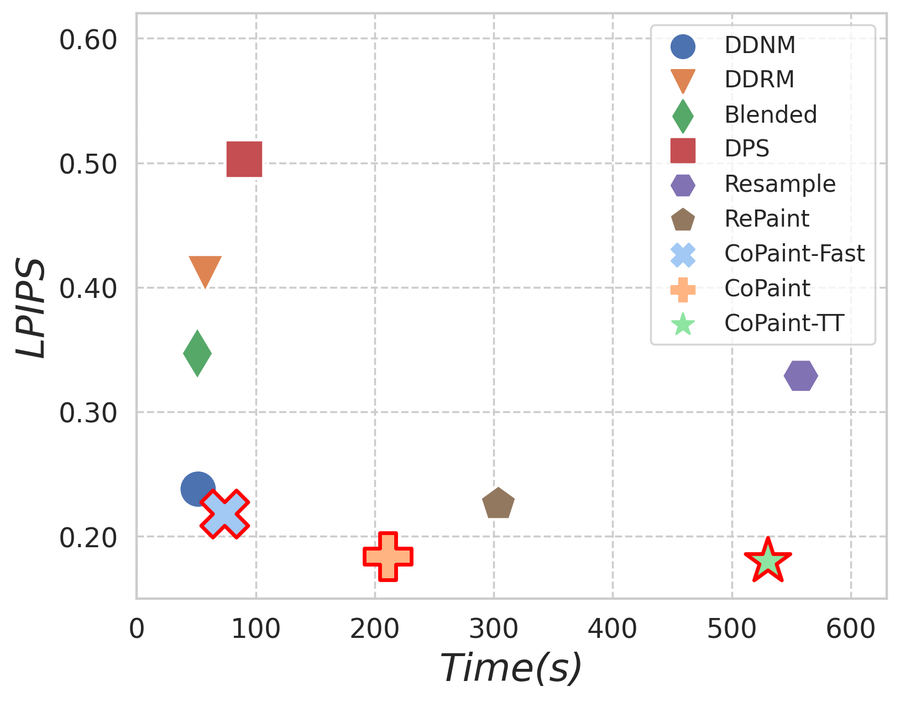}
    \caption{\footnotesize Time-performance trade-off on \texttt{CelebA-HQ} (\emph{left}) and \texttt{ImageNet} (\emph{right}). 
    The x-axis indicates the average time (\e{\downarrow}) to process one image, and the y-axis is the average \texttt{LPIPS} (\e{\downarrow}).
    }
    \label{fig:time2metric}
    \vspace{-3mm}
\end{figure}
\begin{figure}[t]
\begin{center}
\resizebox{\linewidth}{!}{
\begin{tabular}{m{0.05cm}ccccccc}
    & {\footnotesize{\textit{Expand}}} \hspace{-4mm}
    & {\footnotesize{\textit{Half}}} \hspace{-4mm}
    & {\footnotesize{\textit{A.L.}}} \hspace{-4mm}
    & {\footnotesize{\textit{S.R.}}} \hspace{-4mm}
    & {\footnotesize{\textit{Narrow}}} \hspace{-4mm}
    & {\footnotesize{\textit{Wide}}} \hspace{-4mm}
    & {\footnotesize{\textit{Text}}} \\
    \rotatebox[origin=c]{90}{\footnotesize Input } \hspace{-6mm}
    & \multicolumn{1}{m{1.9cm}}{\includegraphics[width=1.9cm,height=1.9cm]{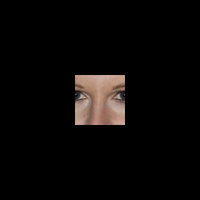}} 
    \hspace{-4mm}
    & \multicolumn{1}{m{1.9cm}}{\includegraphics[width=1.9cm,height=1.9cm]{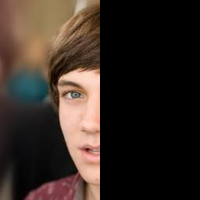}} \hspace{-4mm}
    & \multicolumn{1}{m{1.9cm}}{\includegraphics[width=1.9cm,height=1.9cm]{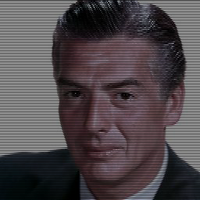}} \hspace{-4mm}
    & \multicolumn{1}{m{1.9cm}}{\includegraphics[width=1.9cm,height=1.9cm]{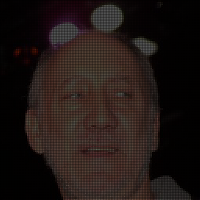}} \hspace{-4mm}
    & \multicolumn{1}{m{1.9cm}}{\includegraphics[width=1.9cm,height=1.9cm]{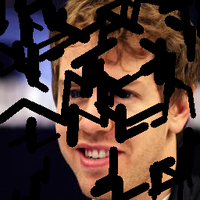}} \hspace{-4mm}
    & \multicolumn{1}{m{1.9cm}}{\includegraphics[width=1.9cm,height=1.9cm]{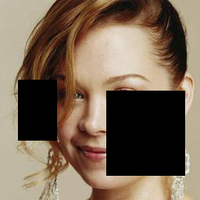}} \hspace{-4mm}
    & \multicolumn{1}{m{1.9cm}}{\includegraphics[width=1.9cm,height=1.9cm]{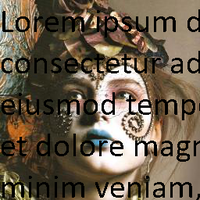}} 
    \\
    \rotatebox[origin=c]{90}{\footnotesize \textsc{Blended} } \hspace{-6mm}
    & \multicolumn{1}{m{1.9cm}}{\includegraphics[width=1.9cm,height=1.9cm]{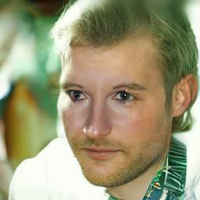}} \hspace{-4mm}
    & \multicolumn{1}{m{1.9cm}}{\includegraphics[width=1.9cm,height=1.9cm]{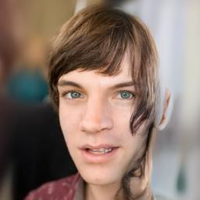}} \hspace{-4mm}
    & \multicolumn{1}{m{1.9cm}}{\includegraphics[width=1.9cm,height=1.9cm]{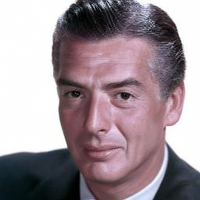}} \hspace{-4mm}
    & \multicolumn{1}{m{1.9cm}}{\includegraphics[width=1.9cm,height=1.9cm]{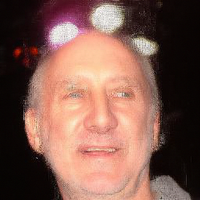}} \hspace{-4mm}
    & \multicolumn{1}{m{1.9cm}}{\includegraphics[width=1.9cm,height=1.9cm]{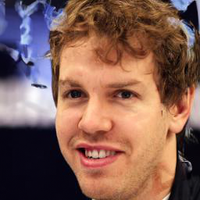}} \hspace{-4mm}
    & \multicolumn{1}{m{1.9cm}}{\includegraphics[width=1.9cm,height=1.9cm]{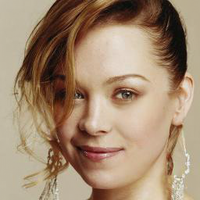}} \hspace{-4mm}
    & \multicolumn{1}{m{1.9cm}}{\includegraphics[width=1.9cm,height=1.9cm]{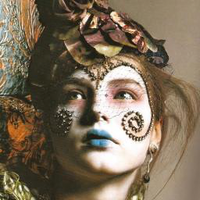}} 
    \\
    \rotatebox[origin=c]{90}{\footnotesize \textsc{Resampling} } \hspace{-6mm}
    & \multicolumn{1}{m{1.9cm}}{\includegraphics[width=1.9cm,height=1.9cm]{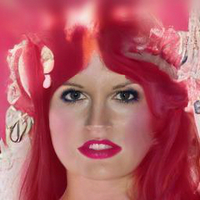}} \hspace{-4mm}
    & \multicolumn{1}{m{1.9cm}}{\includegraphics[width=1.9cm,height=1.9cm]{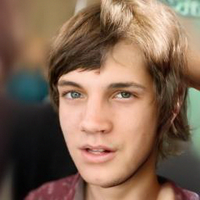}} \hspace{-4mm}
    & \multicolumn{1}{m{1.9cm}}{\includegraphics[width=1.9cm,height=1.9cm]{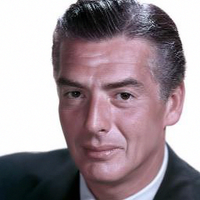}} \hspace{-4mm}
    & \multicolumn{1}{m{1.9cm}}{\includegraphics[width=1.9cm,height=1.9cm]{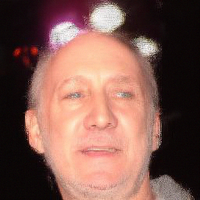}} \hspace{-4mm}
    & \multicolumn{1}{m{1.9cm}}{\includegraphics[width=1.9cm,height=1.9cm]{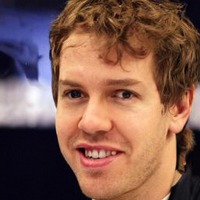}} \hspace{-4mm}
    & \multicolumn{1}{m{1.9cm}}{\includegraphics[width=1.9cm,height=1.9cm]{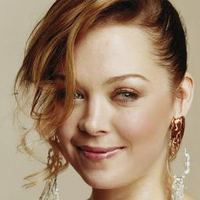}} \hspace{-4mm}
    & \multicolumn{1}{m{1.9cm}}{\includegraphics[width=1.9cm,height=1.9cm]{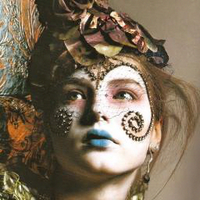}} 
    \\
    \rotatebox[origin=c]{90}{\footnotesize \textsc{Ddrm} } \hspace{-6mm}
    & \multicolumn{1}{m{1.9cm}}{\includegraphics[width=1.9cm,height=1.9cm]{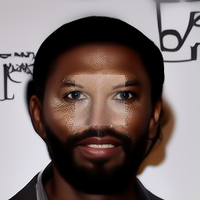}} \hspace{-4mm}
    & \multicolumn{1}{m{1.9cm}}{\includegraphics[width=1.9cm,height=1.9cm]{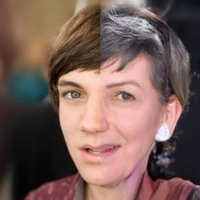}} \hspace{-4mm}
    & \multicolumn{1}{m{1.9cm}}{\includegraphics[width=1.9cm,height=1.9cm]{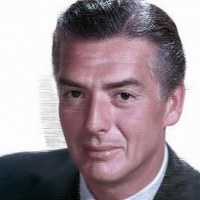}} \hspace{-4mm}
    & \multicolumn{1}{m{1.9cm}}{\includegraphics[width=1.9cm,height=1.9cm]{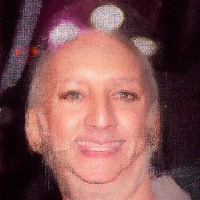}} \hspace{-4mm}
    & \multicolumn{1}{m{1.9cm}}{\includegraphics[width=1.9cm,height=1.9cm]{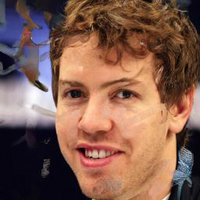}} \hspace{-4mm}
    & \multicolumn{1}{m{1.9cm}}{\includegraphics[width=1.9cm,height=1.9cm]{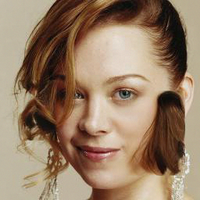}} \hspace{-4mm}
    & \multicolumn{1}{m{1.9cm}}{\includegraphics[width=1.9cm,height=1.9cm]{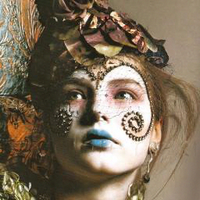}} 
    \\
    \rotatebox[origin=c]{90}{\footnotesize \textsc{RePaint} } \hspace{-6mm}
    & \multicolumn{1}{m{1.9cm}}{\includegraphics[width=1.9cm,height=1.9cm]{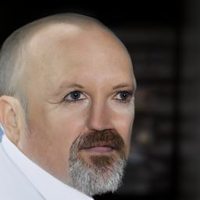}} \hspace{-4mm}
    & \multicolumn{1}{m{1.9cm}}{\includegraphics[width=1.9cm,height=1.9cm]{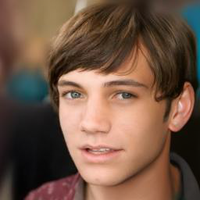}} \hspace{-4mm}
    & \multicolumn{1}{m{1.9cm}}{\includegraphics[width=1.9cm,height=1.9cm]{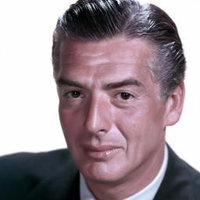}} \hspace{-4mm}
    & \multicolumn{1}{m{1.9cm}}{\includegraphics[width=1.9cm,height=1.9cm]{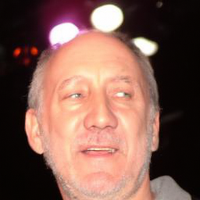}} \hspace{-4mm}
    & \multicolumn{1}{m{1.9cm}}{\includegraphics[width=1.9cm,height=1.9cm]{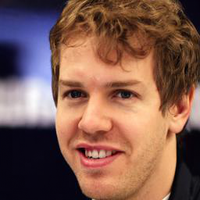}} \hspace{-4mm}
    & \multicolumn{1}{m{1.9cm}}{\includegraphics[width=1.9cm,height=1.9cm]{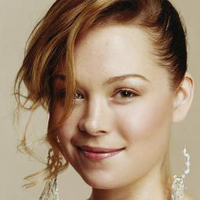}} \hspace{-4mm}
    & \multicolumn{1}{m{1.9cm}}{\includegraphics[width=1.9cm,height=1.9cm]{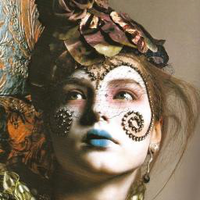}} 
    \\
    \rotatebox[origin=c]{90}{\footnotesize \textsc{Ddnm} } \hspace{-6mm}
    & \multicolumn{1}{m{1.9cm}}{\includegraphics[width=1.9cm,height=1.9cm]{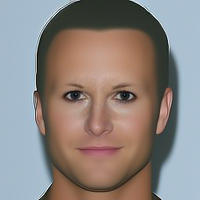}} \hspace{-4mm}
    & \multicolumn{1}{m{1.9cm}}{\includegraphics[width=1.9cm,height=1.9cm]{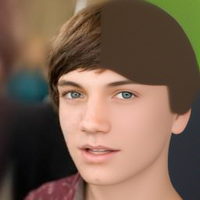}} \hspace{-4mm}
    & \multicolumn{1}{m{1.9cm}}{\includegraphics[width=1.9cm,height=1.9cm]{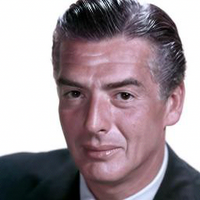}} \hspace{-4mm}
    & \multicolumn{1}{m{1.9cm}}{\includegraphics[width=1.9cm,height=1.9cm]{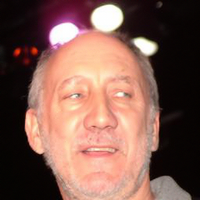}} \hspace{-4mm}
    & \multicolumn{1}{m{1.9cm}}{\includegraphics[width=1.9cm,height=1.9cm]{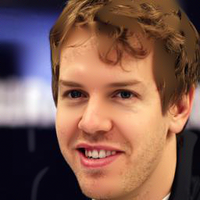}} \hspace{-4mm}
    & \multicolumn{1}{m{1.9cm}}{\includegraphics[width=1.9cm,height=1.9cm]{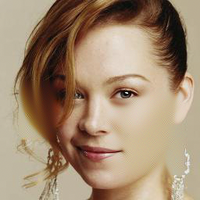}} \hspace{-4mm}
    & \multicolumn{1}{m{1.9cm}}{\includegraphics[width=1.9cm,height=1.9cm]{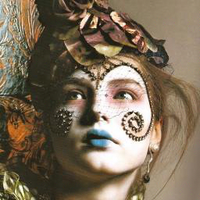}} 
    \\
    \rotatebox[origin=c]{90}{\footnotesize \textsc{Dps} } \hspace{-6mm}
    & \multicolumn{1}{m{1.9cm}}{\includegraphics[width=1.9cm,height=1.9cm]{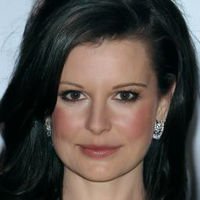}} \hspace{-4mm}
    & \multicolumn{1}{m{1.9cm}}{\includegraphics[width=1.9cm,height=1.9cm]{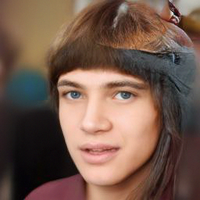}} \hspace{-4mm}
    & \multicolumn{1}{m{1.9cm}}{\includegraphics[width=1.9cm,height=1.9cm]{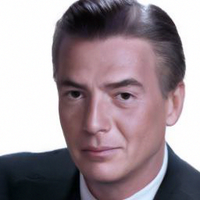}} \hspace{-4mm}
    & \multicolumn{1}{m{1.9cm}}{\includegraphics[width=1.9cm,height=1.9cm]{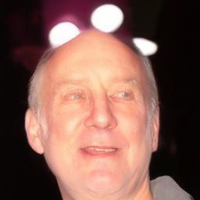}} \hspace{-4mm}
    & \multicolumn{1}{m{1.9cm}}{\includegraphics[width=1.9cm,height=1.9cm]{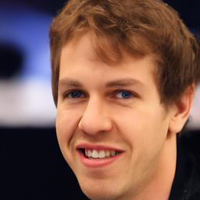}} \hspace{-4mm}
    & \multicolumn{1}{m{1.9cm}}{\includegraphics[width=1.9cm,height=1.9cm]{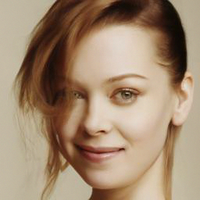}} \hspace{-4mm}
    & \multicolumn{1}{m{1.9cm}}{\includegraphics[width=1.9cm,height=1.9cm]{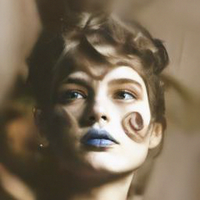}} 
    \\
    \rotatebox[origin=c]{90}{\footnotesize \bf \textsc{CoPaint-Fast} } \hspace{-6mm}
    & \multicolumn{1}{m{1.9cm}}{\includegraphics[width=1.9cm,height=1.9cm]{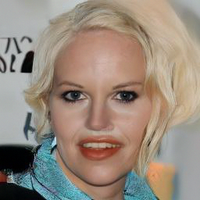}} \hspace{-4mm}
    & \multicolumn{1}{m{1.9cm}}{\includegraphics[width=1.9cm,height=1.9cm]{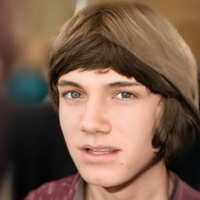}} \hspace{-4mm}
    & \multicolumn{1}{m{1.9cm}}{\includegraphics[width=1.9cm,height=1.9cm]{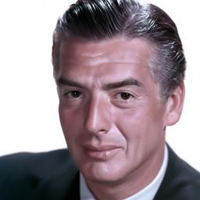}} \hspace{-4mm}
    & \multicolumn{1}{m{1.9cm}}{\includegraphics[width=1.9cm,height=1.9cm]{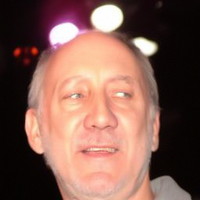}} \hspace{-4mm}
    & \multicolumn{1}{m{1.9cm}}{\includegraphics[width=1.9cm,height=1.9cm]{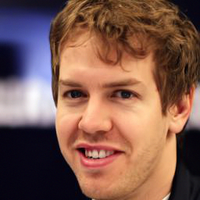}} \hspace{-4mm}
    & \multicolumn{1}{m{1.9cm}}{\includegraphics[width=1.9cm,height=1.9cm]{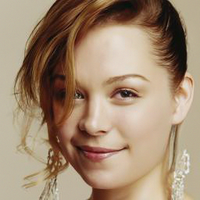}} \hspace{-4mm}
    & \multicolumn{1}{m{1.9cm}}{\includegraphics[width=1.9cm,height=1.9cm]{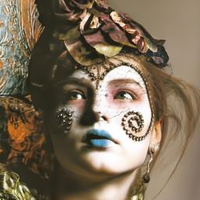}} 
    \\
    \rotatebox[origin=c]{90}{\footnotesize \bf \textsc{CoPaint} } \hspace{-6mm}
    & \multicolumn{1}{m{1.9cm}}{\includegraphics[width=1.9cm,height=1.9cm]{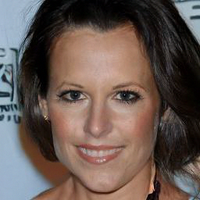}} \hspace{-4mm}
    & \multicolumn{1}{m{1.9cm}}{\includegraphics[width=1.9cm,height=1.9cm]{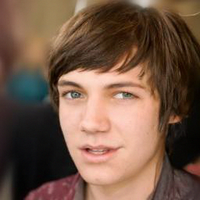}} \hspace{-4mm}
    & \multicolumn{1}{m{1.9cm}}{\includegraphics[width=1.9cm,height=1.9cm]{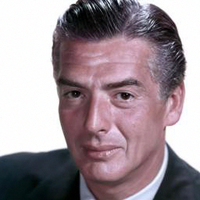}} \hspace{-4mm}
    & \multicolumn{1}{m{1.9cm}}{\includegraphics[width=1.9cm,height=1.9cm]{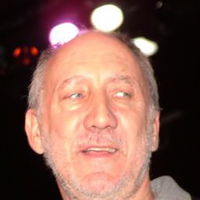}} \hspace{-4mm}
    & \multicolumn{1}{m{1.9cm}}{\includegraphics[width=1.9cm,height=1.9cm]{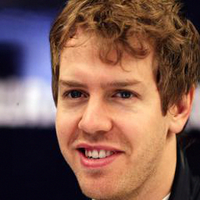}} \hspace{-4mm}
    & \multicolumn{1}{m{1.9cm}}{\includegraphics[width=1.9cm,height=1.9cm]{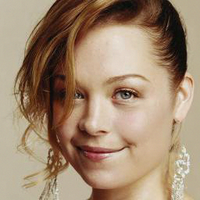}} \hspace{-4mm}
    & \multicolumn{1}{m{1.9cm}}{\includegraphics[width=1.9cm,height=1.9cm]{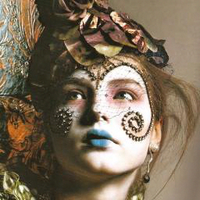}} 
    \\
    \rotatebox[origin=c]{90}{\footnotesize \bf \textsc{CoPaint-TT} } \hspace{-6mm}
    & \multicolumn{1}{m{1.9cm}}{\includegraphics[width=1.9cm,height=1.9cm]{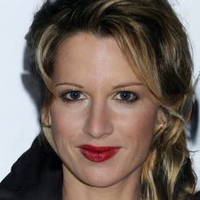}} \hspace{-4mm}
    & \multicolumn{1}{m{1.9cm}}{\includegraphics[width=1.9cm,height=1.9cm]{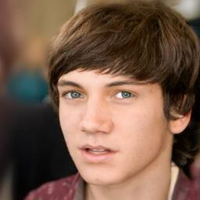}} \hspace{-4mm}
    & \multicolumn{1}{m{1.9cm}}{\includegraphics[width=1.9cm,height=1.9cm]{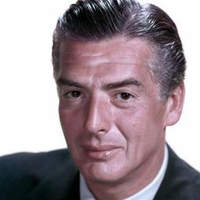}} \hspace{-4mm}
    & \multicolumn{1}{m{1.9cm}}{\includegraphics[width=1.9cm,height=1.9cm]{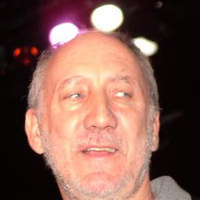}} \hspace{-4mm}
    & \multicolumn{1}{m{1.9cm}}{\includegraphics[width=1.9cm,height=1.9cm]{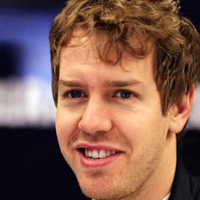}} \hspace{-4mm}
    & \multicolumn{1}{m{1.9cm}}{\includegraphics[width=1.9cm,height=1.9cm]{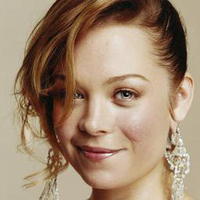}} \hspace{-4mm}
    & \multicolumn{1}{m{1.9cm}}{\includegraphics[width=1.9cm,height=1.9cm]{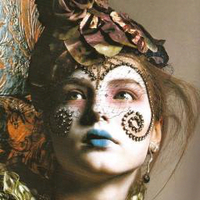}} 
    \\
\end{tabular}}
\end{center}
\caption{\footnotesize Qualitative results of baselines and ours (\textsc{CoPaint}, \textsc{CoPaint-TT}) on \texttt{CelebA-HQ} with seven degradation masks.}
\label{fig:qualitative_celebahq}
\end{figure}

\begin{figure}[t]
\begin{center}
\resizebox{\linewidth}{!}{
\begin{tabular}{m{0.07cm}ccccccc}
    & {\footnotesize{\textit{Expand}}} \hspace{-4mm}
    & {\footnotesize{\textit{Half}}} \hspace{-4mm}
    & {\footnotesize{\textit{Altern}}} \hspace{-4mm}
    & {\footnotesize{\textit{S.R.}}} \hspace{-4mm}
    & {\footnotesize{\textit{Narrow}}} \hspace{-4mm}
    & {\footnotesize{\textit{Wide}}} \hspace{-4mm}
    & {\footnotesize{\textit{Text}}} \\
    \rotatebox[origin=c]{90}{\footnotesize Input } \hspace{-4mm}
    & \multicolumn{1}{m{1.9cm}}{\includegraphics[width=1.9cm,height=1.9cm]{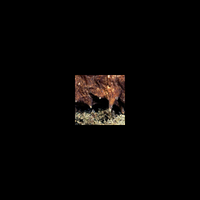}} \hspace{-4mm}
    & \multicolumn{1}{m{1.9cm}}{\includegraphics[width=1.9cm,height=1.9cm]{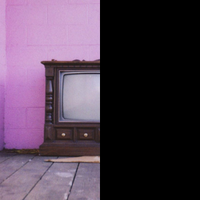}} \hspace{-4mm}
    & \multicolumn{1}{m{1.9cm}}{\includegraphics[width=1.9cm,height=1.9cm]{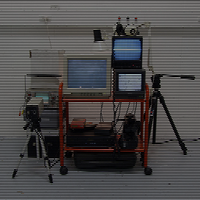}} \hspace{-4mm}
    & \multicolumn{1}{m{1.9cm}}{\includegraphics[width=1.9cm,height=1.9cm]{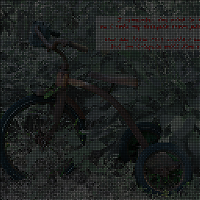}} \hspace{-4mm}
    & \multicolumn{1}{m{1.9cm}}{\includegraphics[width=1.9cm,height=1.9cm]{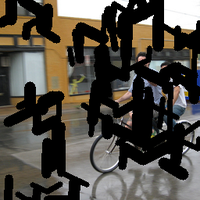}} \hspace{-4mm}
    & \multicolumn{1}{m{1.9cm}}{\includegraphics[width=1.9cm,height=1.9cm]{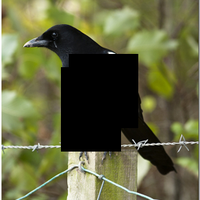}} \hspace{-4mm}
    & \multicolumn{1}{m{1.9cm}}{\includegraphics[width=1.9cm,height=1.9cm]{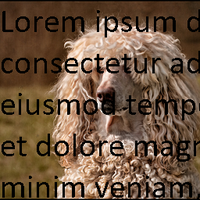}} 
    \\
    \rotatebox[origin=c]{90}{\footnotesize \textsc{Blended} } \hspace{-4mm}
    & \multicolumn{1}{m{1.9cm}}{\includegraphics[width=1.9cm,height=1.9cm]{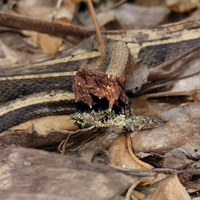}} \hspace{-4mm}
    & \multicolumn{1}{m{1.9cm}}{\includegraphics[width=1.9cm,height=1.9cm]{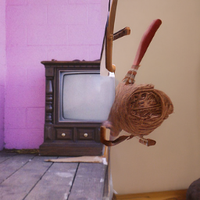}} \hspace{-4mm}
    & \multicolumn{1}{m{1.9cm}}{\includegraphics[width=1.9cm,height=1.9cm]{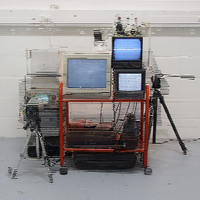}} \hspace{-4mm}
    & \multicolumn{1}{m{1.9cm}}{\includegraphics[width=1.9cm,height=1.9cm]{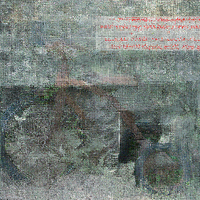}} \hspace{-4mm}
    & \multicolumn{1}{m{1.9cm}}{\includegraphics[width=1.9cm,height=1.9cm]{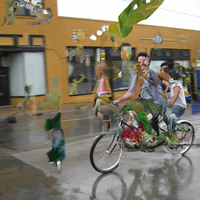}} \hspace{-4mm}
    & \multicolumn{1}{m{1.9cm}}{\includegraphics[width=1.9cm,height=1.9cm]{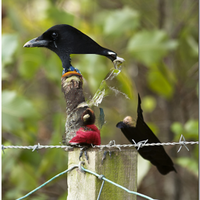}} \hspace{-4mm}
    & \multicolumn{1}{m{1.9cm}}{\includegraphics[width=1.9cm,height=1.9cm]{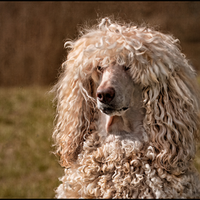}} 
    \\
    \rotatebox[origin=c]{90}{\footnotesize \textsc{Resampling} } \hspace{-4mm}
    & \multicolumn{1}{m{1.9cm}}{\includegraphics[width=1.9cm,height=1.9cm]{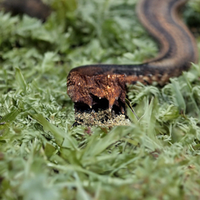}} \hspace{-4mm}
    & \multicolumn{1}{m{1.9cm}}{\includegraphics[width=1.9cm,height=1.9cm]{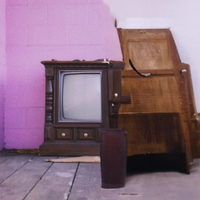}} \hspace{-4mm}
    & \multicolumn{1}{m{1.9cm}}{\includegraphics[width=1.9cm,height=1.9cm]{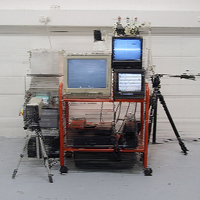}} \hspace{-4mm}
    & \multicolumn{1}{m{1.9cm}}{\includegraphics[width=1.9cm,height=1.9cm]{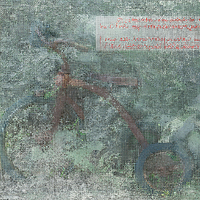}} \hspace{-4mm}
    & \multicolumn{1}{m{1.9cm}}{\includegraphics[width=1.9cm,height=1.9cm]{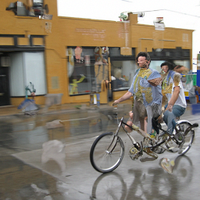}} \hspace{-4mm}
    & \multicolumn{1}{m{1.9cm}}{\includegraphics[width=1.9cm,height=1.9cm]{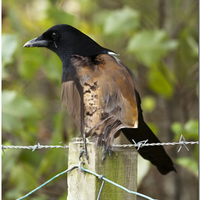}} \hspace{-4mm}
    & \multicolumn{1}{m{1.9cm}}{\includegraphics[width=1.9cm,height=1.9cm]{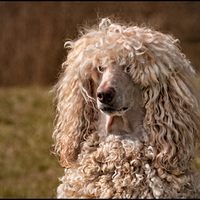}} 
    \\
    \rotatebox[origin=c]{90}{\footnotesize \textsc{Ddrm} } \hspace{-4mm}
    & \multicolumn{1}{m{1.9cm}}{\includegraphics[width=1.9cm,height=1.9cm]{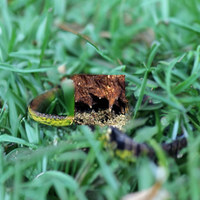}} \hspace{-4mm}
    & \multicolumn{1}{m{1.9cm}}{\includegraphics[width=1.9cm,height=1.9cm]{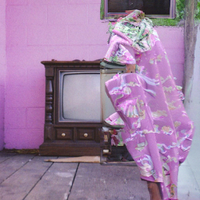}} \hspace{-4mm}
    & \multicolumn{1}{m{1.9cm}}{\includegraphics[width=1.9cm,height=1.9cm]{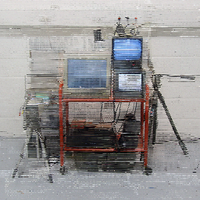}} \hspace{-4mm}
    & \multicolumn{1}{m{1.9cm}}{\includegraphics[width=1.9cm,height=1.9cm]{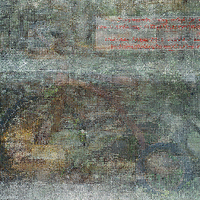}} \hspace{-4mm}
    & \multicolumn{1}{m{1.9cm}}{\includegraphics[width=1.9cm,height=1.9cm]{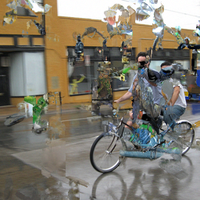}} \hspace{-4mm}
    & \multicolumn{1}{m{1.9cm}}{\includegraphics[width=1.9cm,height=1.9cm]{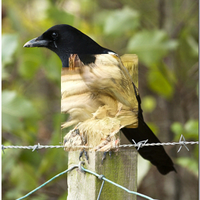}} \hspace{-4mm}
    & \multicolumn{1}{m{1.9cm}}{\includegraphics[width=1.9cm,height=1.9cm]{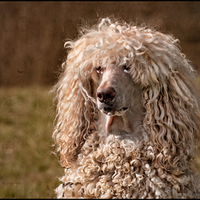}} 
    \\
    \rotatebox[origin=c]{90}{\footnotesize \textsc{RePaint} } \hspace{-4mm}
    & \multicolumn{1}{m{1.9cm}}{\includegraphics[width=1.9cm,height=1.9cm]{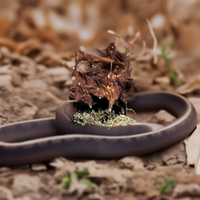}} \hspace{-4mm}
    & \multicolumn{1}{m{1.9cm}}{\includegraphics[width=1.9cm,height=1.9cm]{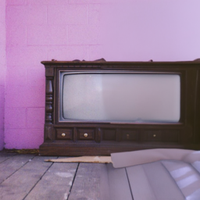}} \hspace{-4mm}
    & \multicolumn{1}{m{1.9cm}}{\includegraphics[width=1.9cm,height=1.9cm]{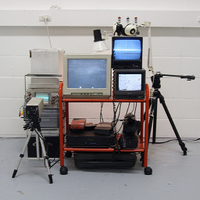}} \hspace{-4mm}
    & \multicolumn{1}{m{1.9cm}}{\includegraphics[width=1.9cm,height=1.9cm]{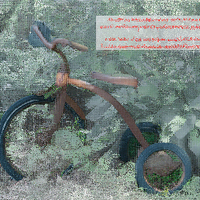}} \hspace{-4mm}
    & \multicolumn{1}{m{1.9cm}}{\includegraphics[width=1.9cm,height=1.9cm]{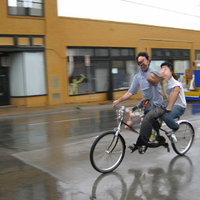}} \hspace{-4mm}
    & \multicolumn{1}{m{1.9cm}}{\includegraphics[width=1.9cm,height=1.9cm]{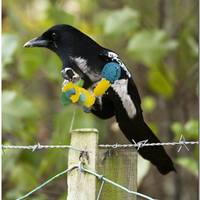}} \hspace{-4mm}
    & \multicolumn{1}{m{1.9cm}}{\includegraphics[width=1.9cm,height=1.9cm]{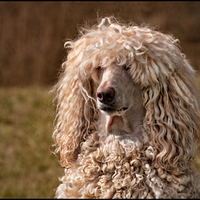}} 
    \\
    \rotatebox[origin=c]{90}{\footnotesize \textsc{Ddnm} } \hspace{-4mm}
    & \multicolumn{1}{m{1.9cm}}{\includegraphics[width=1.9cm,height=1.9cm]{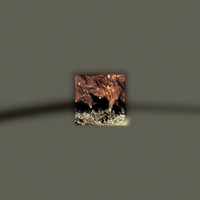}} \hspace{-4mm}
    & \multicolumn{1}{m{1.9cm}}{\includegraphics[width=1.9cm,height=1.9cm]{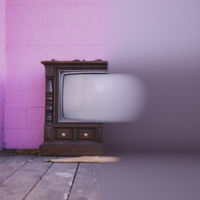}} \hspace{-4mm}
    & \multicolumn{1}{m{1.9cm}}{\includegraphics[width=1.9cm,height=1.9cm]{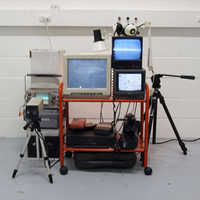}} \hspace{-4mm}
    & \multicolumn{1}{m{1.9cm}}{\includegraphics[width=1.9cm,height=1.9cm]{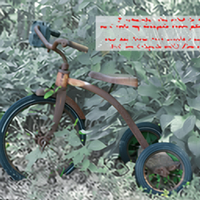}} \hspace{-4mm}
    & \multicolumn{1}{m{1.9cm}}{\includegraphics[width=1.9cm,height=1.9cm]{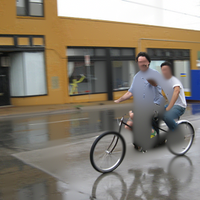}} \hspace{-4mm}
    & \multicolumn{1}{m{1.9cm}}{\includegraphics[width=1.9cm,height=1.9cm]{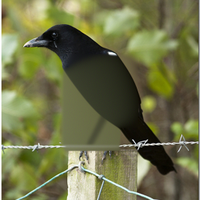}} \hspace{-4mm}
    & \multicolumn{1}{m{1.9cm}}{\includegraphics[width=1.9cm,height=1.9cm]{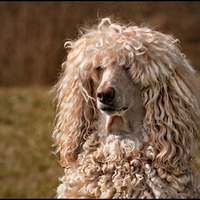}} 
    \\
    \rotatebox[origin=c]{90}{\footnotesize \textsc{Dps} } \hspace{-4mm}
    & \multicolumn{1}{m{1.9cm}}{\includegraphics[width=1.9cm,height=1.9cm]{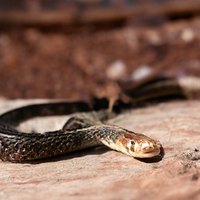}} \hspace{-4mm}
    & \multicolumn{1}{m{1.9cm}}{\includegraphics[width=1.9cm,height=1.9cm]{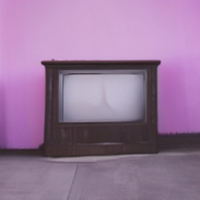}} \hspace{-4mm}
    & \multicolumn{1}{m{1.9cm}}{\includegraphics[width=1.9cm,height=1.9cm]{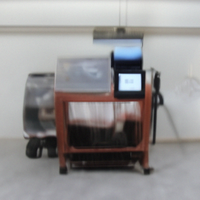}} \hspace{-4mm}
    & \multicolumn{1}{m{1.9cm}}{\includegraphics[width=1.9cm,height=1.9cm]{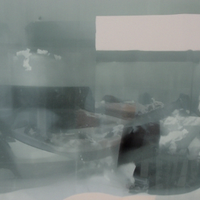}} \hspace{-4mm}
    & \multicolumn{1}{m{1.9cm}}{\includegraphics[width=1.9cm,height=1.9cm]{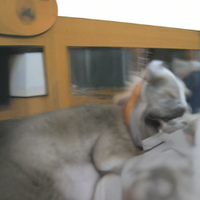}} \hspace{-4mm}
    & \multicolumn{1}{m{1.9cm}}{\includegraphics[width=1.9cm,height=1.9cm]{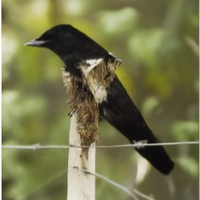}} \hspace{-4mm}
    & \multicolumn{1}{m{1.9cm}}{\includegraphics[width=1.9cm,height=1.9cm]{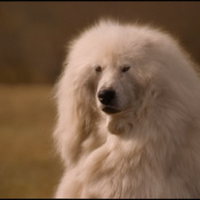}} 
    \\
    \rotatebox[origin=c]{90}{\footnotesize \bf \textsc{CoPaint-Fast} } \hspace{-4mm}
    & \multicolumn{1}{m{1.9cm}}{\includegraphics[width=1.9cm,height=1.9cm]{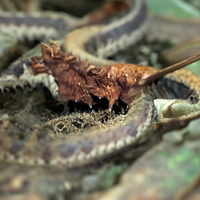}} \hspace{-4mm}
    & \multicolumn{1}{m{1.9cm}}{\includegraphics[width=1.9cm,height=1.9cm]{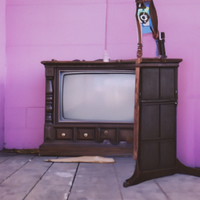}} \hspace{-4mm}
    & \multicolumn{1}{m{1.9cm}}{\includegraphics[width=1.9cm,height=1.9cm]{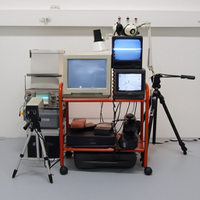}} \hspace{-4mm}
    & \multicolumn{1}{m{1.9cm}}{\includegraphics[width=1.9cm,height=1.9cm]{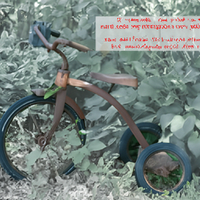}} \hspace{-4mm}
    & \multicolumn{1}{m{1.9cm}}{\includegraphics[width=1.9cm,height=1.9cm]{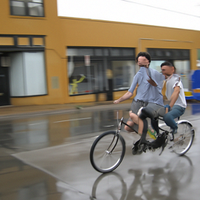}} \hspace{-4mm}
    & \multicolumn{1}{m{1.9cm}}{\includegraphics[width=1.9cm,height=1.9cm]{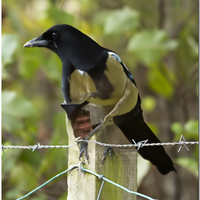}} \hspace{-4mm}
    & \multicolumn{1}{m{1.9cm}}{\includegraphics[width=1.9cm,height=1.9cm]{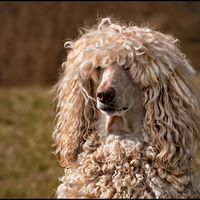}} 
    \\
    \rotatebox[origin=c]{90}{\footnotesize \bf \textsc{CoPaint} } \hspace{-4mm}
    & \multicolumn{1}{m{1.9cm}}{\includegraphics[width=1.9cm,height=1.9cm]{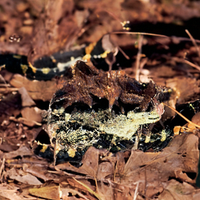}} \hspace{-4mm}
    & \multicolumn{1}{m{1.9cm}}{\includegraphics[width=1.9cm,height=1.9cm]{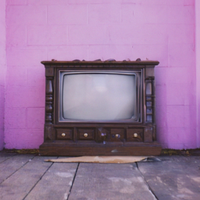}} \hspace{-4mm}
    & \multicolumn{1}{m{1.9cm}}{\includegraphics[width=1.9cm,height=1.9cm]{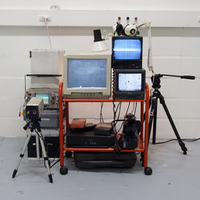}} \hspace{-4mm}
    & \multicolumn{1}{m{1.9cm}}{\includegraphics[width=1.9cm,height=1.9cm]{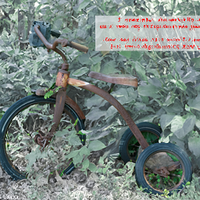}} \hspace{-4mm}
    & \multicolumn{1}{m{1.9cm}}{\includegraphics[width=1.9cm,height=1.9cm]{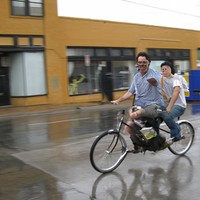}} \hspace{-4mm}
    & \multicolumn{1}{m{1.9cm}}{\includegraphics[width=1.9cm,height=1.9cm]{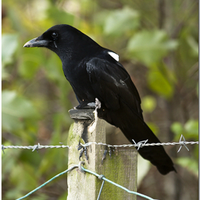}} \hspace{-4mm}
    & \multicolumn{1}{m{1.9cm}}{\includegraphics[width=1.9cm,height=1.9cm]{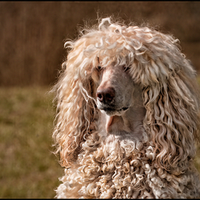}} 
    \\
    \rotatebox[origin=c]{90}{\footnotesize \bf \textsc{CoPaint-TT} } \hspace{-4mm}
    & \multicolumn{1}{m{1.9cm}}{\includegraphics[width=1.9cm,height=1.9cm]{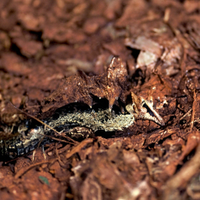}} \hspace{-4mm}
    & \multicolumn{1}{m{1.9cm}}{\includegraphics[width=1.9cm,height=1.9cm]{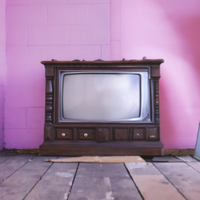}} \hspace{-4mm}
    & \multicolumn{1}{m{1.9cm}}{\includegraphics[width=1.9cm,height=1.9cm]{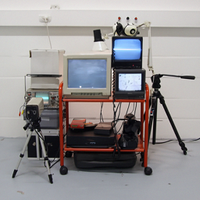}} \hspace{-4mm}
    & \multicolumn{1}{m{1.9cm}}{\includegraphics[width=1.9cm,height=1.9cm]{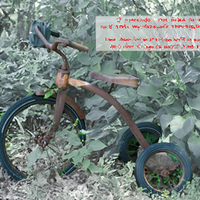}} \hspace{-4mm}
    & \multicolumn{1}{m{1.9cm}}{\includegraphics[width=1.9cm,height=1.9cm]{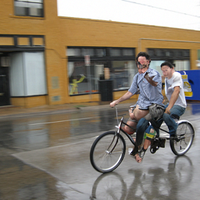}} \hspace{-4mm}
    & \multicolumn{1}{m{1.9cm}}{\includegraphics[width=1.9cm,height=1.9cm]{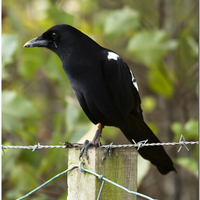}} \hspace{-4mm}
    & \multicolumn{1}{m{1.9cm}}{\includegraphics[width=1.9cm,height=1.9cm]{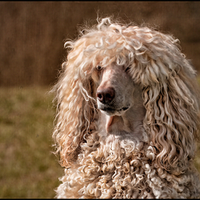}} 
    \\
\end{tabular}}
\end{center}
\caption{\footnotesize Qualitative results of baselines and ours (\textsc{CoPaint}, \textsc{CoPaint-TT}) on \texttt{ImageNet} with seven degradation masks.}
\label{fig:qualitative_imagenet}
\end{figure}

\begin{figure}[h!]
\vspace{4.2mm}
    \centering
    \resizebox{\linewidth}{!}{
    \begin{tabular}{cccccccc}
        \hspace{-5mm} \rotatebox[origin=c]{90}{\footnotesize \textsc{Blended} } \hspace{-5mm}
        & \multicolumn{1}{m{2cm}}{\includegraphics[width=2cm,height=2cm]{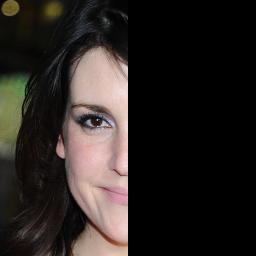}} \hspace{-4mm}
        & \multicolumn{1}{m{2cm}}{\includegraphics[width=2cm,height=2cm]{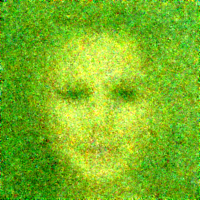}} \hspace{-4mm}
        & \multicolumn{1}{m{2cm}}{\includegraphics[width=2cm,height=2cm]{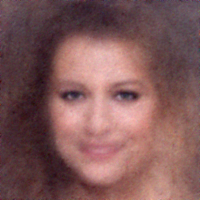}} \hspace{-4mm}
        & \multicolumn{1}{m{2cm}}{\includegraphics[width=2cm,height=2cm]{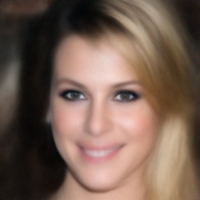}} \hspace{-4mm}
        & \multicolumn{1}{m{2cm}}{\includegraphics[width=2cm,height=2cm]{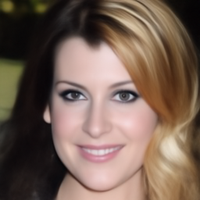}} \hspace{-4mm}
        & \multicolumn{1}{m{2cm}}{\includegraphics[width=2cm,height=2cm]{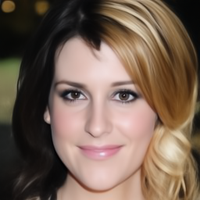}} \hspace{-4mm}
        & \multicolumn{1}{m{2cm}}{\includegraphics[width=2cm,height=2cm]{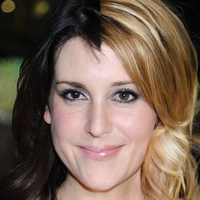}} 
        \\
        \hspace{-5mm} \rotatebox[origin=c]{90}{\footnotesize \bf \textsc{CoPaint} } \hspace{-5mm}
        & \multicolumn{1}{m{2cm}}{\includegraphics[width=2cm,height=2cm]{figures/coherence/10252.png}}\hspace{-4mm}
        & \multicolumn{1}{m{2cm}}{\includegraphics[width=2cm,height=2cm]{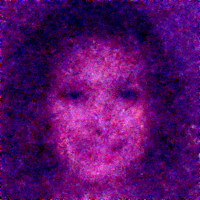}} \hspace{-4mm}
        & \multicolumn{1}{m{2cm}}{\includegraphics[width=2cm,height=2cm]{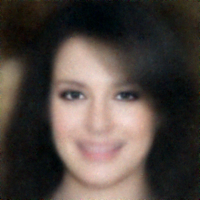}} \hspace{-4mm}
        & \multicolumn{1}{m{2cm}}{\includegraphics[width=2cm,height=2cm]{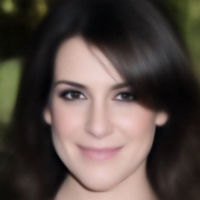}} \hspace{-4mm}
        & \multicolumn{1}{m{2cm}}{\includegraphics[width=2cm,height=2cm]{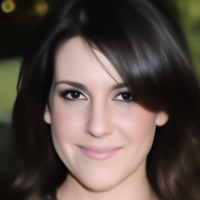}} \hspace{-4mm}
        & \multicolumn{1}{m{2cm}}{\includegraphics[width=2cm,height=2cm]{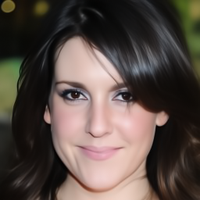}} \hspace{-4mm}
        & \multicolumn{1}{m{2cm}}{\includegraphics[width=2cm,height=2cm]{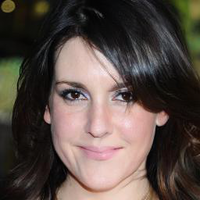}} 
        \\
        & {\large{Input}} \hspace{-4mm}
        & {\large{{$t=T$}}} \hspace{-4mm}
        & {\large{{$t=0.8T$}}} \hspace{-4mm}
        & {\large{{$t=0.6T$}}} \hspace{-4mm}
        & {\large{{$t=0.4T$}}} \hspace{-4mm}
        & {\large{{$t=0.2T$}}} \hspace{-4mm}
        & {\large{{$t=0$}}} 
    \end{tabular}}
    \caption{\footnotesize Coherence study  of baseline \textsc{Blended} and our methods \textsc{CoPaint} on \texttt{CelebA-HQ} dataset with \textit{Half} mask.}
    \label{fig:coherence_short}
    \vspace{-3mm}
\end{figure}

\paragraph{Time-performance trade-off}
Figure~\ref{fig:time2metric} shows the running time of the proposed methods with other baselines on both \texttt{CelebA-HQ} (\emph{left}) and \texttt{ImageNet} (\emph{right}).
In each subfigure, the \e{x}-axis denotes the average running time of each method for processing one image, while the \e{y}-axis represents the average \texttt{LPIPS} score over seven mask types.
The position closer to the left-bottom corner of the figure indicates better performance and time efficiency.
\textsc{CoPaint-TT} achieves the best performance, although it has a larger computational cost than most baselines.
On the other hand, with almost comparable performance, \textsc{CoPaint} reduces the time cost by nearly 60\% in both datasets.
Compared with the best-performing baseline \textsc{RePaint}, \textsc{CoPaint} lies to the left-bottom of \textsc{RePaint} in both datasets, demonstrating its advantage in the time-efficiency tradeoff.
Moreover, we show that \textsc{CoPaint-Fast} is four times faster than \textsc{CoPaint} and is comparable to other baseline methods in terms of running time.
\textsc{CoPaint-Fast} also achieves competitive performances in both two datasets.
Specifically, \textsc{CoPaint-Fast} outperforms other baselines except for \textsc{RePaint} in \texttt{CelebA-HQ} and beats all baselines in \texttt{ImageNet}.

\paragraph{Qualitative results}
We show some example generated images \texttt{CelebA-HQ} and \texttt{ImageNet} in Figures~\ref{fig:qualitative_celebahq} and \ref{fig:qualitative_imagenet}, respectively.
More qualitative results with large size could be seen in Appendix~\ref{app:qualitative_results}.
There are two key observations.
\underline{First}, our method achieves better coherence compared with other baselines, which is  particularly significant with larger masks, such as \textit{Expand} and \textit{Half}.
For example, in the second column in Figure~\ref{fig:qualitative_imagenet} with the \textit{Half} mask, the revealed part of the input is half of a television, as shown in the first row. 
In contrast to the failed completions generated by most baselines, both \textsc{CoPaint} and \textsc{CoPaint-TT} successfully generate a television with matching size and style.
\textsc{CoPaint-Fast} shows slight performance degradation due to the limited number of reverse sampling and gradient descent.
\underline{Second}, although some baselines, such as \textsc{Dps}, also generate relatively coherent images, our methods produce more realistic images.
For example, the televisions generated by our methods have more decorations and grains, while the television generated by \textsc{Dps} appears smooth and lacks details.

\subsection{Coherence Study}
To show how \algname ensures coherence along the denoising process, we present a coherence study, where we plot one-step generations over time steps \e{t=} \e{\{T,} \e{0.8T,} \e{0.6T, 0.4T, 0.2T, 1\}} for the baseline \textsc{Blended} and our method \textsc{CoPaint} in Figure~\ref{fig:coherence_short}.
As can be observed, although the revealed part is a woman with black hair, \textsc{Blended} keeps generating blond hair for the woman.
This is consistent with the known bias in \texttt{CelebA-HQ} dataset, that women are more correlated with blond hair~\citep{Liu2021JustTT}.
The problem is that directly replacing the revealed portion of the image along the denoising process does not require the unrevealed portion to be consistent with the context of the revealed region.
By contrast, our method could effectively generate a coherent image with black hair.

\begin{table}[t]
\caption{\footnotesize Ablation study of the gradient descent number $G$, the time travel frequency $K$, the time travel interval $\tau$, and the step number for approximating $\tilde{\bm X}_0$ in each time step $H$. The results are based on the testing set of \texttt{CelebA-HQ} dataset with \textit{Half} mask.}
\label{tab:ablation_G_lr}

\centering

\resizebox{0.33\textwidth}{!}{
\centering
\footnotesize
\begin{tabular}{cccc}
\toprule[1pt]
\midrule
    Method & $G$ & \texttt{LPIPS}$\downarrow$ & \texttt{Time (s)}$\downarrow$ \\
    \midrule
    \multirow{2}{*}{\textsc{CoPaint-TT}} & 1 & 0.187 & 326 \\
    & 2 & 0.180 & 562 \\
    & 5 & 0.192 & 1365 \\
\midrule
\midrule
    Method & $K$ & \texttt{LPIPS}$\downarrow$ & \texttt{Time (s)}$\downarrow$ \\
    \midrule
    \multirow{2}{*}{\textsc{CoPaint-TT}} 
    & 1 & 0.180 & 562 \\
    & 2 & 0.179 & 721 \\
    & 5 & 0.181 & 1428 \\
\midrule
\midrule
    Method & $\tau$ & \texttt{LPIPS}$\downarrow$ & \texttt{Time (s)}$\downarrow$ \\
    \midrule
    \multirow{2}{*}{\textsc{CoPaint-TT}} 
    & 2 & 0.186 & 567 \\
    & 5 & 0.178 & 569 \\
    & 10 & 0.180 & 562 \\
    & 20 & 0.181 & 564 \\
\midrule
\midrule
    Method & $H$ & \texttt{LPIPS}$\downarrow$ & \texttt{Time (s)}$\downarrow$ \\
    \midrule
    \multirow{2}{*}{\textsc{CoPaint-TT}} 
    & 1 & 0.180 & 562 \\
    & 2 & 0.176 & 1491 \\
    & 5 & 0.177 & 3346 \\
\midrule
\bottomrule[1pt]
\end{tabular}
}
\end{table}
\subsection{Ablation Study} \label{exp:ablation}
We investigate the design choices of three hyperparameters, gradient descent step number \e{G}, time travel frequency \e{K}, time travel interval \e{\tau}, and the effects of multi-step approximation as mentioned in Section~\ref{sec:method}.
Specifically, we conduct our experiments on the \texttt{CelebA-HQ} with \textit{Half} mask.
The results could be seen in Table~\ref{tab:ablation_G_lr}.

As shown in Algorithm~\ref{alg:copaint+tt}, a \e{G}-step gradient descent method is adopted for optimizing \e{\tilde{\bm X}_t} at each time step.
In Table~\ref{tab:ablation_G_lr} (\emph{first}), we see that a larger \e{G} would not always introduce better performances \textsc{CoPaint-TT}.
As we optimize \e{\tilde{\bm X}_t} to minimize the mean square error (corresponding to the second term in \Eqref{eq:Xt_posterior_approx}) only in the revealed region, a larger gradient descent number may introduce an overfitting problem and thus lead the poor performances.

Table~\ref{tab:ablation_G_lr} (\emph{second} and \emph{third}) shows the effects of time travel frequency \e{K} and interval \e{\tau}.
Different from \textsc{RePaint}~\citep{repaint} where \e{K=9} is used, we see that \e{K=1} is sufficient for our method, demonstrating that our proposed method is better at imposing the inpainting constraints than the simple replacement operations adopted by \textsc{RePaint}.
Besides, we show that the value of time travel interval \e{\tau} does not have a significant impact on the performance with \e{\tau \geq 5}. 

As we have mentioned in Section~\ref{sec:method-additional}, the one-step approximation for \e{\tilde{\bm X}_0} could be replaced with multi-step approximation by going through multiple deterministic denoising steps at a subset of time steps.
We denote the approximation step number as \e{H}, and its effects could be seen in   Table~\ref{tab:ablation_G_lr} (\emph{fourth}).
We see that with a minor decrease in \texttt{LPIPS}, the time cost dramatically increases.
With \e{H=5}, it takes about six times longer than \e{H=1} for processing one image.
We leave it for our future work to explore how to improve computational efficiency for multi-step approximation.

\section{Conclusion}
In this paper, we proposed a diffusion-based image inpainting method, \textsc{CoPaint}, which introduces a Bayesian framework to jointly modify both revealed and unrevealed parts of intermediate variables in each time step along the denoising process, leading to better coherence in the inpainted image. 
\algnamens's approximation error of the posterior distribution is designed to gradually drop to zero, thus strongly enforcing the inpainting constraint. 
Results from extensive experiments showed that \textsc{CoPaint} outperforms existing diffusion-based methods in both objective and subjective metrics in terms of coherence and overall quality. 
However, there are still some imperfections in \algnamens, due to the suboptimal greedy optimization and one-step approximation error.
See the failure case study in Appendix~\ref{app:fail} and the discussion about potential societal impacts in Appendix~\ref{app:societal}. 
In the next step, we plan to replace our greedy optimization with more plausible sampling methods and investigate ways to further reduce approximation error.

\bibliography{ref}

\begin{thebibliography}{57}
\providecommand{\natexlab}[1]{#1}
\providecommand{\url}[1]{\texttt{#1}}
\expandafter\ifx\csname urlstyle\endcsname\relax
  \providecommand{\doi}[1]{doi: #1}\else
  \providecommand{\doi}{doi: \begingroup \urlstyle{rm}\Url}\fi

\bibitem[Avrahami et~al.(2021)Avrahami, Lischinski, and Fried]{blended}
Avrahami, O., Lischinski, D., and Fried, O.
\newblock Blended diffusion for text-driven editing of natural images.
\newblock \emph{2022 IEEE/CVF Conference on Computer Vision and Pattern
  Recognition (CVPR)}, pp.\  18187--18197, 2021.

\bibitem[Bansal et~al.(2022)Bansal, Borgnia, Chu, Li, Kazemi, Huang, Goldblum,
  Geiping, and Goldstein]{Bansal2022ColdDI}
Bansal, A., Borgnia, E., Chu, H.-M., Li, J., Kazemi, H., Huang, F., Goldblum,
  M., Geiping, J., and Goldstein, T.
\newblock Cold diffusion: Inverting arbitrary image transforms without noise.
\newblock \emph{ArXiv}, abs/2208.09392, 2022.

\bibitem[Batzolis et~al.(2022)Batzolis, Stanczuk, Schonlieb, and
  Etmann]{Batzolis2022NonUniformDM}
Batzolis, G., Stanczuk, J., Schonlieb, C.-B., and Etmann, C.
\newblock Non-uniform diffusion models.
\newblock \emph{ArXiv}, abs/2207.09786, 2022.

\bibitem[Benton et~al.(2022)Benton, Shi, Bortoli, Deligiannidis, and
  Doucet]{Benton2022FromDD}
Benton, J., Shi, Y., Bortoli, V.~D., Deligiannidis, G., and Doucet, A.
\newblock From denoising diffusions to denoising markov models.
\newblock \emph{ArXiv}, abs/2211.03595, 2022.

\bibitem[Bond-Taylor et~al.(2021)Bond-Taylor, Hessey, Sasaki, Breckon, and
  Willcocks]{BondTaylor2021UnleashingTP}
Bond-Taylor, S., Hessey, P., Sasaki, H., Breckon, T., and Willcocks, C.~G.
\newblock Unleashing transformers: Parallel token prediction with discrete
  absorbing diffusion for fast high-resolution image generation from
  vector-quantized codes.
\newblock In \emph{European Conference on Computer Vision}, 2021.

\bibitem[Chung et~al.(2021)Chung, Sim, and
  Ye]{chungComeCloserDiffuseFaster2022}
Chung, H., Sim, B., and Ye, J.-C.
\newblock Come-closer-diffuse-faster: Accelerating conditional diffusion models
  for inverse problems through stochastic contraction.
\newblock \emph{2022 IEEE/CVF Conference on Computer Vision and Pattern
  Recognition (CVPR)}, pp.\  12403--12412, 2021.

\bibitem[Chung et~al.(2022{\natexlab{a}})Chung, Kim, Mccann, Klasky, and
  Ye]{dps}
Chung, H., Kim, J., Mccann, M.~T., Klasky, M.~L., and Ye, J.~C.
\newblock Diffusion posterior sampling for general noisy inverse problems.
\newblock \emph{arXiv preprint arXiv:2209.14687}, 2022{\natexlab{a}}.

\bibitem[Chung et~al.(2022{\natexlab{b}})Chung, Sim, Ryu, and
  Ye]{Chung2022ImprovingDM}
Chung, H., Sim, B., Ryu, D., and Ye, J.~C.
\newblock Improving diffusion models for inverse problems using manifold
  constraints.
\newblock \emph{ArXiv}, abs/2206.00941, 2022{\natexlab{b}}.

\bibitem[Dhariwal \& Nichol(2021{\natexlab{a}})Dhariwal and
  Nichol]{dhariwalDiffusion2021}
Dhariwal, P. and Nichol, A.
\newblock Diffusion models beat gans on image synthesis.
\newblock \emph{ArXiv}, abs/2105.05233, 2021{\natexlab{a}}.

\bibitem[Dhariwal \& Nichol(2021{\natexlab{b}})Dhariwal and Nichol]{guided}
Dhariwal, P. and Nichol, A.
\newblock Diffusion models beat gans on image synthesis.
\newblock \emph{Advances in Neural Information Processing Systems},
  34:\penalty0 8780--8794, 2021{\natexlab{b}}.

\bibitem[Guo et~al.(2019)Guo, Chen, Yu, Chen, and Liu]{Guo2019ProgressiveII}
Guo, Z., Chen, Z., Yu, T., Chen, J., and Liu, S.
\newblock Progressive image inpainting with full-resolution residual network.
\newblock \emph{Proceedings of the 27th ACM International Conference on
  Multimedia}, 2019.

\bibitem[Ho et~al.(2020)Ho, Jain, and Abbeel]{hoDenoising}
Ho, J., Jain, A., and Abbeel, P.
\newblock Denoising diffusion probabilistic models.
\newblock \emph{ArXiv}, abs/2006.11239, 2020.

\bibitem[Hong et~al.(2019)Hong, Xiong, Ji, and Fan]{Hong2019DeepFN}
Hong, X., Xiong, P., Ji, R., and Fan, H.
\newblock Deep fusion network for image completion.
\newblock \emph{Proceedings of the 27th ACM International Conference on
  Multimedia}, 2019.

\bibitem[Horita et~al.(2022)Horita, Yang, Chen, Koyama, and
  Aizawa]{Horita2022ASD}
Horita, D., Yang, J., Chen, D., Koyama, Y., and Aizawa, K.
\newblock A structure-guided diffusion model for large-hole diverse image
  completion.
\newblock \emph{ArXiv}, abs/2211.10437, 2022.

\bibitem[Horwitz \& Hoshen(2022)Horwitz and Hoshen]{Horwitz2022ConffusionCI}
Horwitz, E. and Hoshen, Y.
\newblock Conffusion: Confidence intervals for diffusion models.
\newblock \emph{ArXiv}, abs/2211.09795, 2022.

\bibitem[Iizuka et~al.(2017)Iizuka, Simo-Serra, and
  Ishikawa]{Iizuka2017GloballyAL}
Iizuka, S., Simo-Serra, E., and Ishikawa, H.
\newblock Globally and locally consistent image completion.
\newblock \emph{ACM Transactions on Graphics (TOG)}, 36:\penalty0 1 -- 14,
  2017.

\bibitem[Kawar et~al.(2022)Kawar, Elad, Ermon, and Song]{ddrm}
Kawar, B., Elad, M., Ermon, S., and Song, J.
\newblock Denoising diffusion restoration models.
\newblock \emph{arXiv preprint arXiv:2201.11793}, 2022.

\bibitem[Krizhevsky(2014)]{alexnet}
Krizhevsky, A.
\newblock One weird trick for parallelizing convolutional neural networks.
\newblock \emph{ArXiv}, abs/1404.5997, 2014.

\bibitem[Ku et~al.(2022)Ku, Siu, Cheng, and Chan]{Ku2022IntelligentPP}
Ku, W.-F., Siu, W.~C., Cheng, X., and Chan, H.~A.
\newblock Intelligent painter: Picture composition with resampling diffusion
  model.
\newblock \emph{ArXiv}, abs/2210.17106, 2022.

\bibitem[Li et~al.(2022)Li, Yu, Zhou, Song, Lin, and Jia]{Li2022SDMSD}
Li, W., Yu, X., Zhou, K., Song, Y., Lin, Z., and Jia, J.
\newblock Sdm: Spatial diffusion model for large hole image inpainting.
\newblock \emph{ArXiv}, abs/2212.02963, 2022.

\bibitem[Liu et~al.(2021)Liu, Haghgoo, Chen, Raghunathan, Koh, Sagawa, Liang,
  and Finn]{Liu2021JustTT}
Liu, E.~Z., Haghgoo, B., Chen, A.~S., Raghunathan, A., Koh, P.~W., Sagawa, S.,
  Liang, P., and Finn, C.
\newblock Just train twice: Improving group robustness without training group
  information.
\newblock In \emph{International Conference on Machine Learning}, pp.\
  6781--6792. PMLR, 2021.

\bibitem[Liu et~al.(2018)Liu, Reda, Shih, Wang, Tao, and
  Catanzaro]{liu2018image}
Liu, G., Reda, F.~A., Shih, K.~J., Wang, T.-C., Tao, A., and Catanzaro, B.
\newblock Image inpainting for irregular holes using partial convolutions.
\newblock In \emph{European Conference on Computer Vision}, 2018.

\bibitem[Liu et~al.(2020)Liu, Jiang, Song, Huang, and
  Yang]{Liu2020CorrectionTR}
Liu, H., Jiang, B., Song, Y., Huang, W., and Yang, C.
\newblock Correction to: Rethinking image inpainting via a mutual
  encoder-decoder with feature equalizations.
\newblock \emph{Computer Vision – ECCV 2020}, 12347:\penalty0 C1 -- C1, 2020.

\bibitem[Liu et~al.(2022)Liu, Wang, Wang, and Rui]{Liu2022DelvingGI}
Liu, H., Wang, Y., Wang, M., and Rui, Y.
\newblock Delving globally into texture and structure for image inpainting.
\newblock \emph{Proceedings of the 30th ACM International Conference on
  Multimedia}, 2022.

\bibitem[Liu et~al.(2014)Liu, Luo, Wang, and Tang]{celeba}
Liu, Z., Luo, P., Wang, X., and Tang, X.
\newblock Deep learning face attributes in the wild.
\newblock \emph{2015 IEEE International Conference on Computer Vision (ICCV)},
  pp.\  3730--3738, 2014.

\bibitem[Lugmayr et~al.(2022)Lugmayr, Danelljan, Romero, Yu, Timofte, and
  Gool]{repaint}
Lugmayr, A., Danelljan, M., Romero, A., Yu, F., Timofte, R., and Gool, L.~V.
\newblock Repaint: Inpainting using denoising diffusion probabilistic models.
\newblock \emph{2022 IEEE/CVF Conference on Computer Vision and Pattern
  Recognition (CVPR)}, pp.\  11451--11461, 2022.

\bibitem[Nazeri et~al.(2019)Nazeri, Ng, Joseph, Qureshi, and
  Ebrahimi]{Nazeri2019EdgeConnectSG}
Nazeri, K., Ng, E., Joseph, T., Qureshi, F.~Z., and Ebrahimi, M.
\newblock Edgeconnect: Structure guided image inpainting using edge prediction.
\newblock \emph{2019 IEEE/CVF International Conference on Computer Vision
  Workshop (ICCVW)}, pp.\  3265--3274, 2019.

\bibitem[Neal(2011)]{Neal2011MCMCUH}
Neal, R.~M.
\newblock Mcmc using hamiltonian dynamics.
\newblock \emph{arXiv: Computation}, pp.\  139--188, 2011.

\bibitem[Nichol et~al.(2021)Nichol, Dhariwal, Ramesh, Shyam, Mishkin, McGrew,
  Sutskever, and Chen]{nicholGLIDE2022}
Nichol, A., Dhariwal, P., Ramesh, A., Shyam, P., Mishkin, P., McGrew, B.,
  Sutskever, I., and Chen, M.
\newblock Glide: Towards photorealistic image generation and editing with
  text-guided diffusion models.
\newblock In \emph{International Conference on Machine Learning}, 2021.

\bibitem[Pathak et~al.(2016)Pathak, Krahenbuhl, Donahue, Darrell, and
  Efros]{pathak2016context}
Pathak, D., Krahenbuhl, P., Donahue, J., Darrell, T., and Efros, A.~A.
\newblock Context encoders: Feature learning by inpainting.
\newblock In \emph{Proceedings of the IEEE conference on computer vision and
  pattern recognition}, pp.\  2536--2544, 2016.

\bibitem[Peng et~al.(2021)Peng, Liu, Xu, and Li]{Peng2021GeneratingDS}
Peng, J., Liu, D., Xu, S., and Li, H.
\newblock Generating diverse structure for image inpainting with hierarchical
  vq-vae.
\newblock \emph{2021 IEEE/CVF Conference on Computer Vision and Pattern
  Recognition (CVPR)}, pp.\  10770--10779, 2021.

\bibitem[Pokle et~al.(2022)Pokle, Geng, and Kolter]{pokleDeep2022}
Pokle, A., Geng, Z., and Kolter, Z.
\newblock Deep equilibrium approaches to diffusion models.
\newblock \emph{ArXiv}, abs/2210.12867, 2022.

\bibitem[Reddy et~al.(2022)Reddy, Priya, Vinuthna, Reddy, and
  Reddy]{Reddy2022ExplorationOI}
Reddy, V.~R., Priya, B.~L., Vinuthna, P., Reddy, K.~P., and Reddy, D.~S.
\newblock Exploration of image inpainting approaches and challenges: A survey.
\newblock \emph{International Journal of Computer Engineering in Research
  Trends}, 2022.

\bibitem[Rombach et~al.(2021)Rombach, Blattmann, Lorenz, Esser, and
  Ommer]{rombachHighResolution2022}
Rombach, R., Blattmann, A., Lorenz, D., Esser, P., and Ommer, B.
\newblock High-resolution image synthesis with latent diffusion models.
\newblock \emph{2022 IEEE/CVF Conference on Computer Vision and Pattern
  Recognition (CVPR)}, pp.\  10674--10685, 2021.

\bibitem[Russakovsky et~al.(2015)Russakovsky, Deng, Su, Krause, Satheesh, Ma,
  Huang, Karpathy, Khosla, Bernstein, et~al.]{imagenet}
Russakovsky, O., Deng, J., Su, H., Krause, J., Satheesh, S., Ma, S., Huang, Z.,
  Karpathy, A., Khosla, A., Bernstein, M., et~al.
\newblock Imagenet large scale visual recognition challenge.
\newblock \emph{International journal of computer vision}, 115\penalty0
  (3):\penalty0 211--252, 2015.

\bibitem[Saharia et~al.(2021{\natexlab{a}})Saharia, Chan, Chang, Lee, Ho,
  Salimans, Fleet, and Norouzi]{sahariaPalette2022}
Saharia, C., Chan, W., Chang, H., Lee, C.~A., Ho, J., Salimans, T., Fleet,
  D.~J., and Norouzi, M.
\newblock Palette: Image-to-image diffusion models.
\newblock \emph{ACM SIGGRAPH 2022 Conference Proceedings}, 2021{\natexlab{a}}.

\bibitem[Saharia et~al.(2021{\natexlab{b}})Saharia, Ho, Chan, Salimans, Fleet,
  and Norouzi]{Saharia2021ImageSV}
Saharia, C., Ho, J., Chan, W., Salimans, T., Fleet, D.~J., and Norouzi, M.
\newblock Image super-resolution via iterative refinement.
\newblock \emph{IEEE transactions on pattern analysis and machine
  intelligence}, PP, 2021{\natexlab{b}}.

\bibitem[Shah et~al.(2022)Shah, Gautam, and Singh]{Shah2022OverviewOI}
Shah, R., Gautam, A., and Singh, S.~K.
\newblock Overview of image inpainting techniques: A survey.
\newblock \emph{2022 IEEE Region 10 Symposium (TENSYMP)}, pp.\  1--6, 2022.

\bibitem[Sohl-Dickstein et~al.(2015)Sohl-Dickstein, Weiss, Maheswaranathan, and
  Ganguli]{sohl-dicksteinDeep2015}
Sohl-Dickstein, J.~N., Weiss, E.~A., Maheswaranathan, N., and Ganguli, S.
\newblock Deep unsupervised learning using nonequilibrium thermodynamics.
\newblock \emph{ArXiv}, abs/1503.03585, 2015.

\bibitem[Song et~al.(2020)Song, Meng, and Ermon]{songDenoising2022}
Song, J., Meng, C., and Ermon, S.
\newblock Denoising diffusion implicit models.
\newblock \emph{ArXiv}, abs/2010.02502, 2020.

\bibitem[Song \& Ermon(2019{\natexlab{a}})Song and Ermon]{Song2019GenerativeMB}
Song, Y. and Ermon, S.
\newblock Generative modeling by estimating gradients of the data distribution.
\newblock \emph{ArXiv}, abs/1907.05600, 2019{\natexlab{a}}.

\bibitem[Song \& Ermon(2019{\natexlab{b}})Song and Ermon]{songGenerative2020}
Song, Y. and Ermon, S.
\newblock Generative modeling by estimating gradients of the data distribution.
\newblock \emph{ArXiv}, abs/1907.05600, 2019{\natexlab{b}}.

\bibitem[Song et~al.(2018)Song, Yang, Shen, Wang, Huang, and
  Kuo]{Song2018SPGNetSP}
Song, Y., Yang, C., Shen, Y., Wang, P., Huang, Q., and Kuo, C.-C.~J.
\newblock Spg-net: Segmentation prediction and guidance network for image
  inpainting.
\newblock \emph{ArXiv}, abs/1805.03356, 2018.

\bibitem[Suvorov et~al.(2022)Suvorov, Logacheva, Mashikhin, Remizova, Ashukha,
  Silvestrov, Kong, Goka, Park, and Lempitsky]{lama}
Suvorov, R., Logacheva, E., Mashikhin, A., Remizova, A., Ashukha, A.,
  Silvestrov, A., Kong, N., Goka, H., Park, K., and Lempitsky, V.
\newblock Resolution-robust large mask inpainting with fourier convolutions.
\newblock In \emph{Proceedings of the IEEE/CVF Winter Conference on
  Applications of Computer Vision}, pp.\  2149--2159, 2022.

\bibitem[Trippe et~al.(2022)Trippe, Yim, Tischer, Broderick, Baker, Barzilay,
  and Jaakkola]{trippeDiffusion2022}
Trippe, B.~L., Yim, J., Tischer, D.~K., Broderick, T., Baker, D., Barzilay, R.,
  and Jaakkola, T.
\newblock Diffusion probabilistic modeling of protein backbones in 3d for the
  motif-scaffolding problem.
\newblock \emph{ArXiv}, abs/2206.04119, 2022.

\bibitem[Vo et~al.(2018)Vo, Duong, and P{\'e}rez]{Vo2018StructuralI}
Vo, H.~V., Duong, N. Q.~K., and P{\'e}rez, P.
\newblock Structural inpainting.
\newblock \emph{Proceedings of the 26th ACM international conference on
  Multimedia}, 2018.

\bibitem[Wan et~al.(2021)Wan, Zhang, Chen, and Liao]{Wan2021HighFidelityPI}
Wan, Z., Zhang, J., Chen, D., and Liao, J.
\newblock High-fidelity pluralistic image completion with transformers.
\newblock \emph{2021 IEEE/CVF International Conference on Computer Vision
  (ICCV)}, pp.\  4672--4681, 2021.

\bibitem[Wang et~al.(2022)Wang, Yu, and Zhang]{ddnm}
Wang, Y., Yu, J., and Zhang, J.
\newblock Zero-shot image restoration using denoising diffusion null-space
  model.
\newblock \emph{arXiv preprint arXiv:2212.00490}, 2022.

\bibitem[Weng et~al.(2022)Weng, Ding, and Zhou]{Weng2022ASO}
Weng, Y., Ding, S., and Zhou, T.
\newblock A survey on improved gan based image inpainting.
\newblock \emph{2022 2nd International Conference on Consumer Electronics and
  Computer Engineering (ICCECE)}, pp.\  319--322, 2022.

\bibitem[Whang et~al.(2021)Whang, Delbracio, Talebi, Saharia, Dimakis, and
  Milanfar]{whangDeblurring2021}
Whang, J., Delbracio, M., Talebi, H., Saharia, C., Dimakis, A.~G., and
  Milanfar, P.
\newblock Deblurring via stochastic refinement.
\newblock \emph{2022 IEEE/CVF Conference on Computer Vision and Pattern
  Recognition (CVPR)}, pp.\  16272--16282, 2021.

\bibitem[Xiang et~al.(2022)Xiang, Zou, Nawaz, Huang, Zhang, and
  Yu]{Xiang2022DeepLF}
Xiang, H., Zou, Q., Nawaz, M.~A., Huang, X., Zhang, F., and Yu, H.
\newblock Deep learning for image inpainting: A survey.
\newblock \emph{Pattern Recognit.}, 134:\penalty0 109046, 2022.

\bibitem[Xiao et~al.(2018)Xiao, Li, and Chen]{Xiao2018DeepIG}
Xiao, Q., Li, G., and Chen, Q.
\newblock Deep inception generative network for cognitive image inpainting.
\newblock \emph{ArXiv}, abs/1812.01458, 2018.

\bibitem[Yang et~al.(2022)Yang, Zhang, Hong, Xu, Zhao, Shao, Zhang, Yang, and
  Cui]{yangDiffusion2022}
Yang, L., Zhang, Z., Hong, S., Xu, R., Zhao, Y., Shao, Y., Zhang, W., Yang,
  M.-H., and Cui, B.
\newblock Diffusion models: A comprehensive survey of methods and applications.
\newblock \emph{ArXiv}, abs/2209.00796, 2022.

\bibitem[Yu et~al.(2021)Yu, Zhan, Wu, Pan, Cui, Lu, Ma, Xie, and
  Miao]{Yu2021DiverseII}
Yu, Y., Zhan, F., Wu, R., Pan, J., Cui, K., Lu, S., Ma, F., Xie, X., and Miao,
  C.
\newblock Diverse image inpainting with bidirectional and autoregressive
  transformers.
\newblock \emph{Proceedings of the 29th ACM International Conference on
  Multimedia}, 2021.

\bibitem[Zhao et~al.(2020)Zhao, Mo, Lin, Wang, Zuo, Chen, Xing, and
  Lu]{Zhao2020UCTGANDI}
Zhao, L., Mo, Q., Lin, S., Wang, Z., Zuo, Z., Chen, H., Xing, W., and Lu, D.
\newblock Uctgan: Diverse image inpainting based on unsupervised cross-space
  translation.
\newblock \emph{2020 IEEE/CVF Conference on Computer Vision and Pattern
  Recognition (CVPR)}, pp.\  5740--5749, 2020.

\bibitem[Zhao et~al.(2021)Zhao, Cui, Sheng, Dong, Liang, Chang, and
  Xu]{Zhao2021LargeSI}
Zhao, S., Cui, J., Sheng, Y., Dong, Y., Liang, X., Chang, E. I.-C., and Xu, Y.
\newblock Large scale image completion via co-modulated generative adversarial
  networks.
\newblock \emph{ArXiv}, abs/2103.10428, 2021.

\bibitem[Zheng et~al.(2019)Zheng, Cham, and Cai]{Zheng2019PluralisticIC}
Zheng, C., Cham, T.-J., and Cai, J.
\newblock Pluralistic image completion.
\newblock \emph{2019 IEEE/CVF Conference on Computer Vision and Pattern
  Recognition (CVPR)}, pp.\  1438--1447, 2019.

\end{thebibliography}
\bibliographystyle{arxiv2023}

\newpage
\appendix
\onecolumn
\section{Experiment Setup}
\subsection{Human Evaluation} \label{app:user_study}
\begin{figure}[t]
    \centering
    \includegraphics[width=0.65\linewidth]{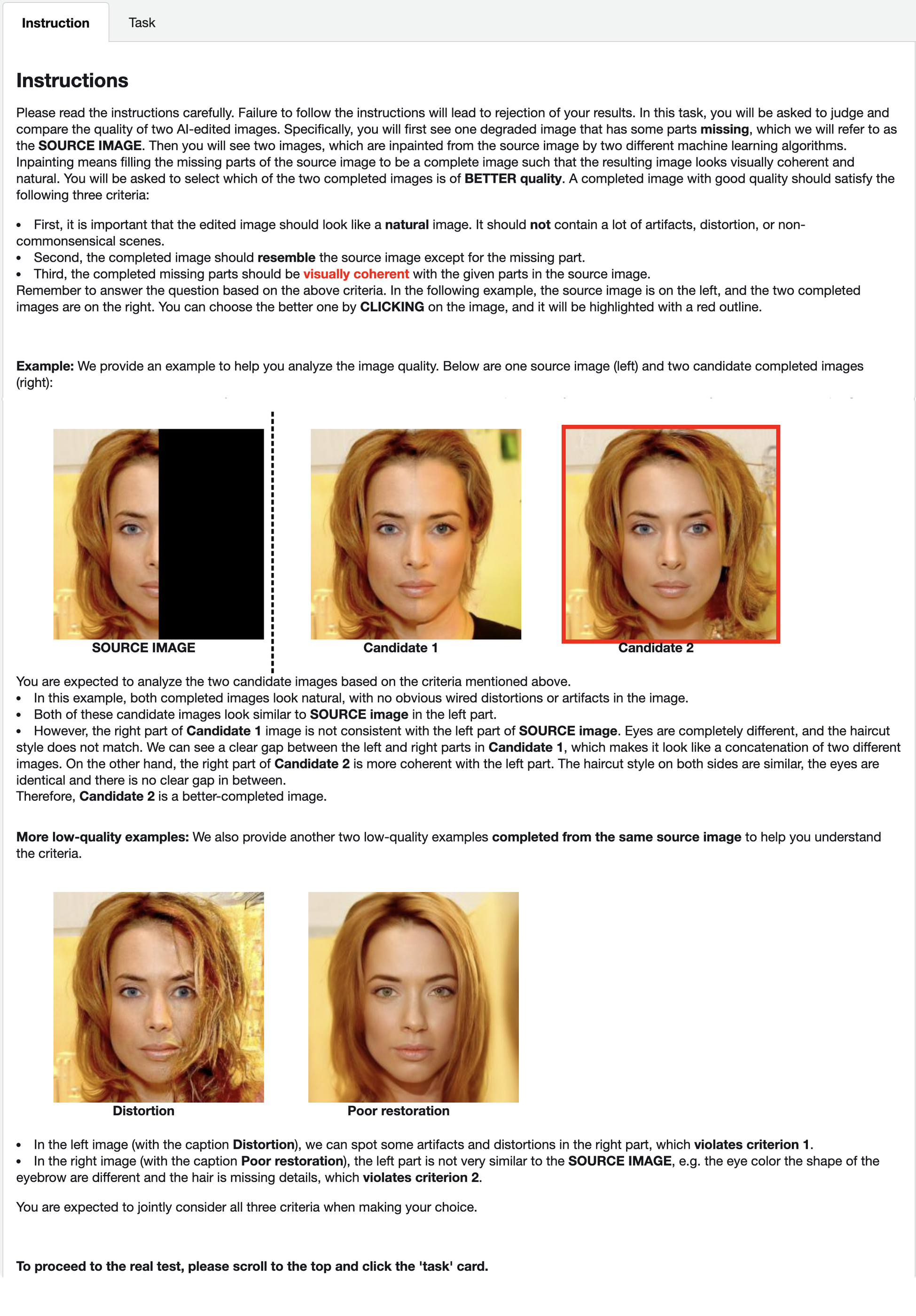}
    \includegraphics[width=0.64\linewidth]{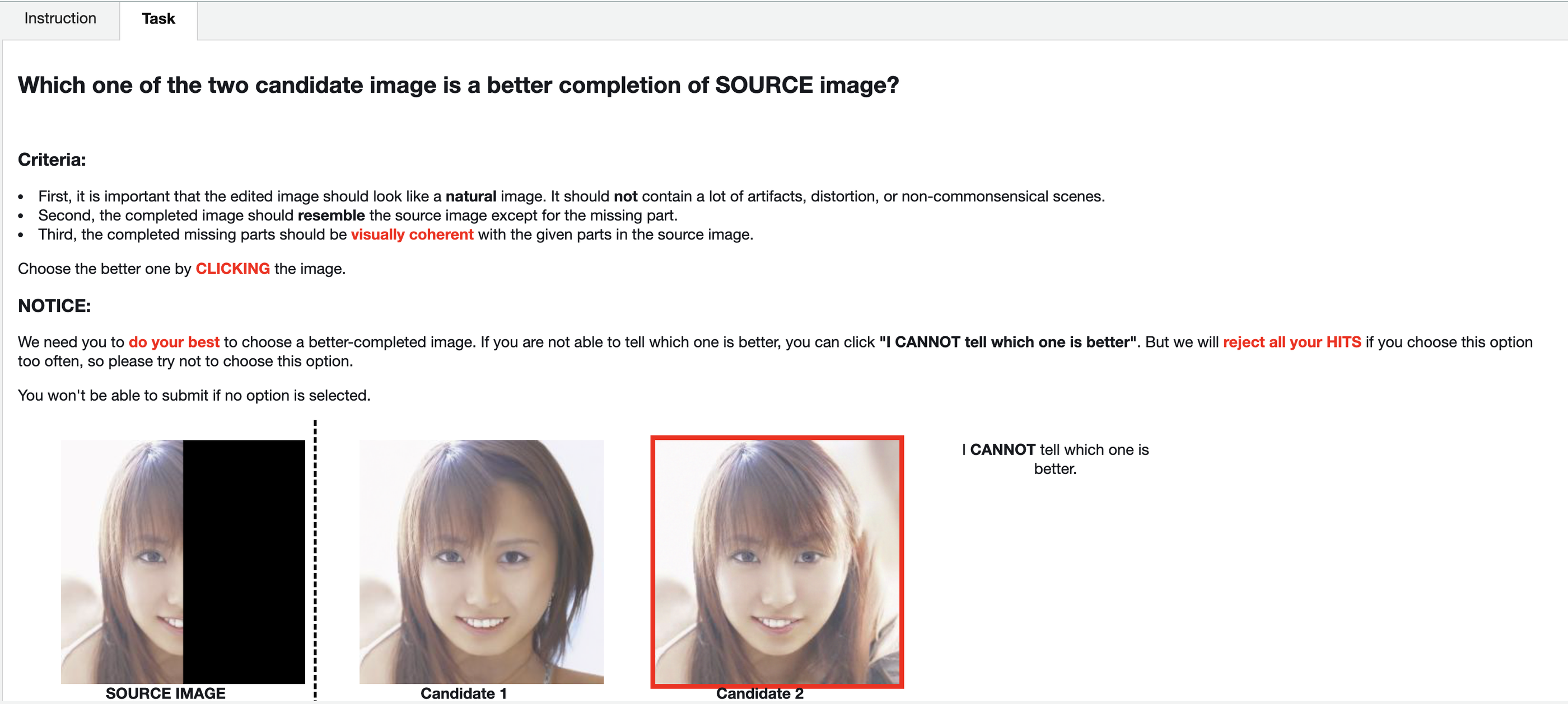}
    \caption{Human evaluation interface for \texttt{overall} test. Evaluators are asked to choose an image of better quality from two \textbf{Candidate} images following the criteria listed in the instructions.}
    \label{fig:app-mturkoverall}
\end{figure}

\begin{figure}[t]
    \centering
    \includegraphics[width=0.65\linewidth]{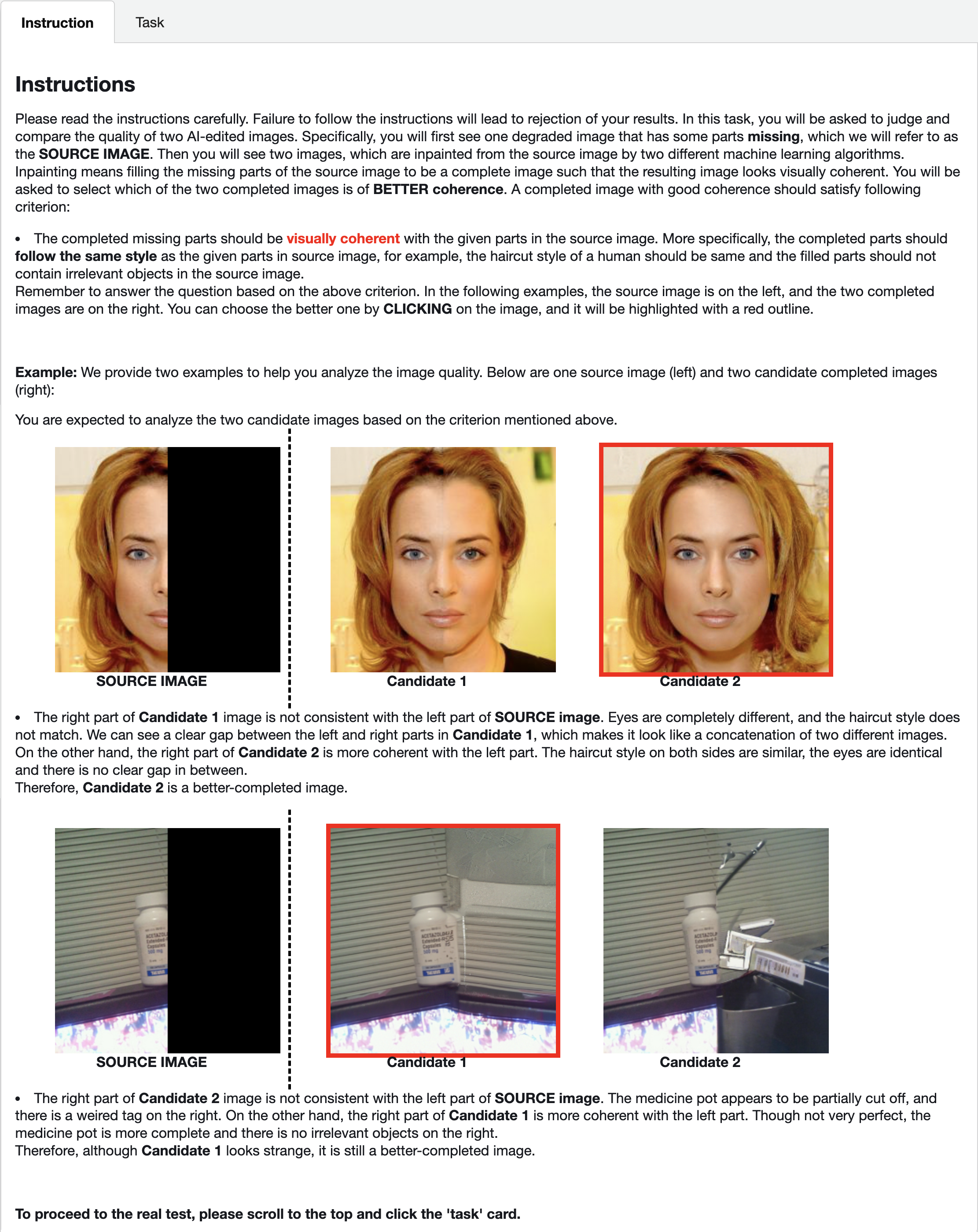}
    \includegraphics[width=0.64\linewidth]{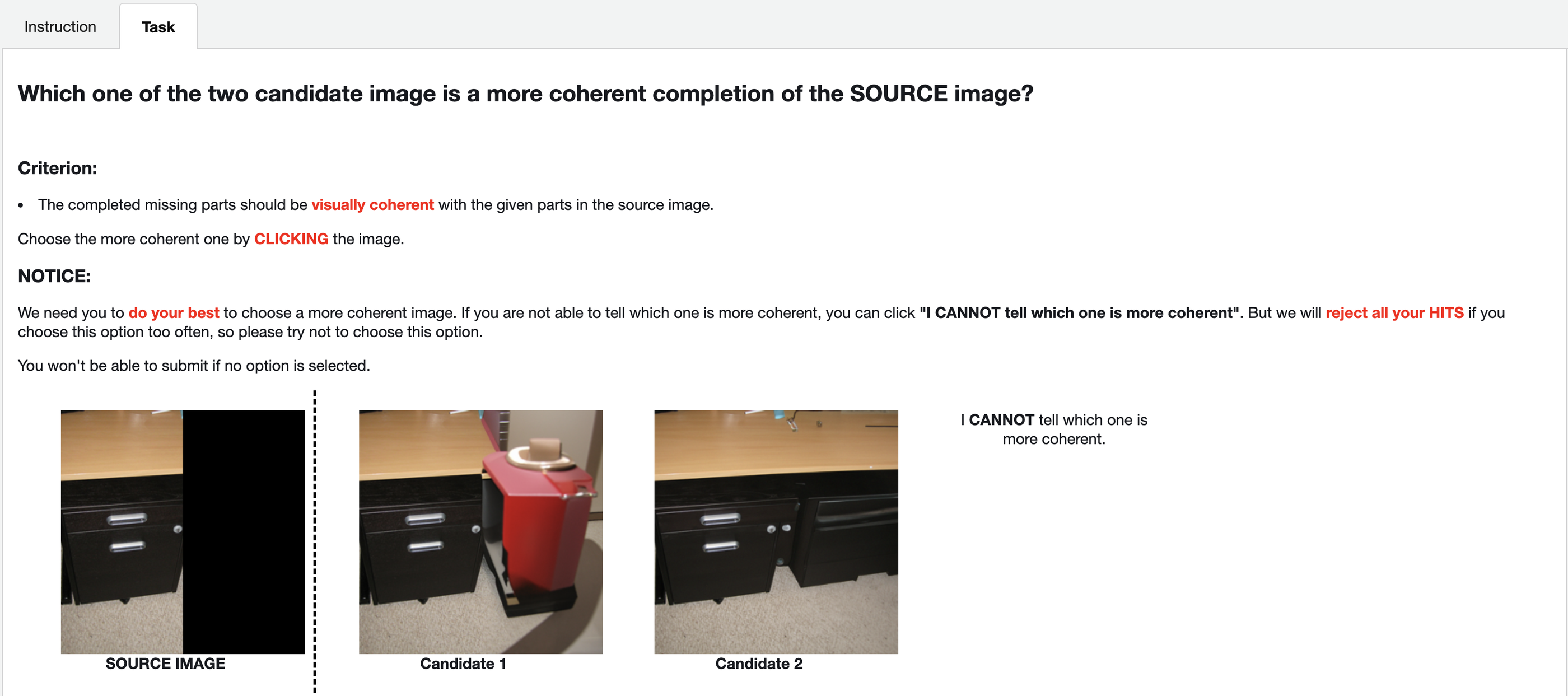}
    \caption{Human evaluation interface for \texttt{coherence} test.  Evaluators are asked to choose an image of better quality from two \textbf{Candidate} images following the criteria listed in the instructions.}
    \label{fig:app-mturkcoherence}
\end{figure}

As described in Section~\ref{exp:metric}, we conduct two human evaluations on Amazon Mturk\footnote{\url{https://www.mturk.com}} to evaluate the quality of inpainted images. Figures~\ref{fig:app-mturkoverall} and~\ref{fig:app-mturkcoherence} show the user interface for the human evaluations, where evaluators are asked to select an image of better quality from two candidate images inpainted by different algorithms according to the criteria listed in instructions. To avoid bias, we put the candidate images in random order. As mentioned in Section~\ref{exp:metric}, we perform two user studies, with one of them focusing on \texttt{overall} quality and the other focusing on \texttt{coherence}. 

Detailed criteria of \texttt{overall} test are:
\begin{itemize}[noitemsep]
    \item It is important that the edited image should look like a \textbf{natural image}. It should not contain a lot of artifacts, distortion or non-commonsensical scenes.
    \item The completed image should \textbf{resemble the source image} except in the missing part.
    \item The completed missing parts should be \textbf{visually coherent} with the given parts in the source image.
\end{itemize}
Detailed criteria of \texttt{coherence} test are:
\begin{itemize}[noitemsep]
    \item The completed missing parts should be visually coherent with the given parts in the source image. More specifically, the completed parts should follow the same style as the given parts in source image, for example, the haircut style of a human should be same and the filled parts should not contain irrelevant objects in the source image. 
\end{itemize}

\clearpage
\subsection{Implementation Details of Baselines}
\label{app:baseline}
We implement all methods based on the code\footnote{\url{shorturl.at/AHILU}} released by~\citet{repaint} and generate images with the same pretrained diffusion model. For \texttt{CelebA-HQ} dataset, we use the model pretrained by~\citet{repaint}. For \texttt{ImageNet}, we use the model pretrained by \citet{guided}. 
For all experiments, we set the number of reverse sampling steps as 250 if not specified otherwise. 
All experiments are done on an Nvidia-V100-SXM2-32GB GPU.
The key hyper-parameters for each baseline method are listed below:

\textbf{\textsc{Blended}}, we use \textsc{DDPM}~\cite{songGenerative2020} sampler with 250 sampling steps.

\textbf{\textsc{Ddrm}}, we perform all experiments with the default setting \e{\eta_{B}=1.0, \eta=0.85}.

\textbf{\textsc{Resampling}}, we generate and resample twenty images\footnote{It is the maximum affordable number for a 32G GPU.} in each time step, and select the two with the highest posterior probability when \e{t=1}.

\textbf{\textsc{RePaint}}, we perform all experiments with the default setting, where jump length \e{j=10} and resampling number \e{n=10}.

\textbf{\textsc{DPS}}, we perform all experiments following the setting of Gaussian noise measurement in the original paper, where the measurement noise is set to 0 and the step size \e{\xi_i=1/ \left\lVert \mathbf{y}-\mathbf{A}(\mathbf{\hat{x}_0}(\mathbf{x_i}))\right\rVert}.

\textbf{\textsc{Ddnm}}, we perform all experiments with the default setting, where linear degradation operator \e{\mathbf{A}= r} and its pseudo-inverse \e{\mathbf{A^{\dag}}=r}.

\subsection{Adaptive Learning Rate for Our Method}
\label{app:lr}
In our algorithm~\ref{alg:copaint+tt}, \e{\tilde{\bm X}_t} is optimized to maximize the posterior in each time step.  
It is equivalent to optimize \e{\tilde{\bm X}_t} by minimizing the following loss,
\begin{equation}
\begin{small}
\begin{aligned}
\mathcal{L}_t = 
        - \log p'_\theta(\tilde{\bm X}_{t} | \tilde{\bm X}_{t+1}, \mathcal{C}) 
        = \frac{1}{2\sigma_t^2} \Vert \tilde{\bm X}_{t} - \tilde{\bm \mu}_{t}\Vert_2^2 +\frac{1}{2\xi_{t}^{'2}} \big\Vert  \bm s_0 - \bm r \big(\bm f_\theta^{(t)}(\tilde{\bm X}_{t})\big) \big\Vert_2^2, 
\end{aligned} 
\end{small}
\end{equation}
and its gradient on \e{\tilde{\bm X}_t} could be calculated as follows,
\begin{equation}
\begin{small}
\begin{aligned}
\nabla_{\tilde{\bm X}_t}  \mathcal{L} 
&= \frac{1}{\sigma_t^2} \big( \tilde{\bm X}_{t} - \tilde{\bm \mu}_{t} \big) +\frac{1}{\xi_{t}^{'2}}
\big[ \nabla_{\tilde{\bm X}_t} \bm r \big(\bm f_\theta^{(t)}(\tilde{\bm X}_{t})  \big]
\big[  \bm s_0 - \bm r \big(\bm f_\theta^{(t)}(\tilde{\bm X}_{t})\big) \big]  \\
&= \frac{1}{\sigma_t^2} \big( \tilde{\bm X}_{t} - \tilde{\bm \mu}_{t} \big) +\frac{1}{\xi_{t}^{'2}}
\frac{\partial \bm f_\theta^{(t)}(\tilde{\bm X}_{t}) }{\partial \tilde{\bm X}_t}
\frac{\partial \bm r \big(\bm f_\theta^{(t)}(\tilde{\bm X}_{t}) \big)}{\partial \bm f_\theta^{(t)}(\tilde{\bm X}_{t}) }
\big[  \bm s_0 - \bm r \big(\bm f_\theta^{(t)}(\tilde{\bm X}_{t})\big) \big].
\end{aligned}
\end{small}
\end{equation}

where we note that \e{\frac{\partial \bm r \big(\bm f_\theta^{(t)}(\tilde{\bm X}_{t}) \big)}{\partial f_\theta^{(t)}(\tilde{\bm X}_{t}) }} is a diagonal matrix with either one or zero. 
Following the one-step approximation function \e{f^{(t)}_\theta(\tilde{\bm X}_t)} in DDIM~\citep{songDenoising2022}, we have
\begin{equation}
\begin{small}
\begin{aligned}
    \frac{\partial f_\theta^{(t)}(\tilde{\bm X}_{t}) }{\partial \tilde{\bm X}_t} &=
    \frac{
    \partial
    \bigg(
    (\tilde{\bm X}_t - \sqrt{1-\bar{\alpha}_t}\bm \epsilon^{(t)}_{\theta}(\tilde{\bm X}_t)) / \sqrt{\bar{\alpha}_t}
    \bigg)
    }{
    \partial
    \tilde{\bm X}_t
    } \\
    &= \frac{
    \bm 1 - \sqrt{1-\bar{\alpha}_t} \nabla_{\tilde{\bm X}_t}\bm \epsilon^{(t)}_{\theta}(\tilde{\bm X}_t)
    }{
    \sqrt{\bar{\alpha}_t}
    }
\end{aligned}
\end{small}
\end{equation}

Given the fact that \e{\{\bar{\alpha}_t\}} is strictly decreasing, \e{1/\sqrt{\bar{\alpha}_t}} could be very large when \e{t} is large and thus lead to large gradient magnitudes for updating \e{\tilde{\bm X}_t}.
In practice, we find that it would easily result in NaN if optimizing \e{\tilde{\bm X}_t} directly with the gradient.
To alleviate the problem, we multiply the learning rate with an offset term \e{\sqrt{\bar{\alpha}_t}}.
With a base learning rate \e{0.02}, we finally use \e{0.02\sqrt{\bar{\alpha}_t}} as our learning rate.

\clearpage
\section{Qualitative Results}
\label{app:qualitative_results}
\begin{figure*}[t]
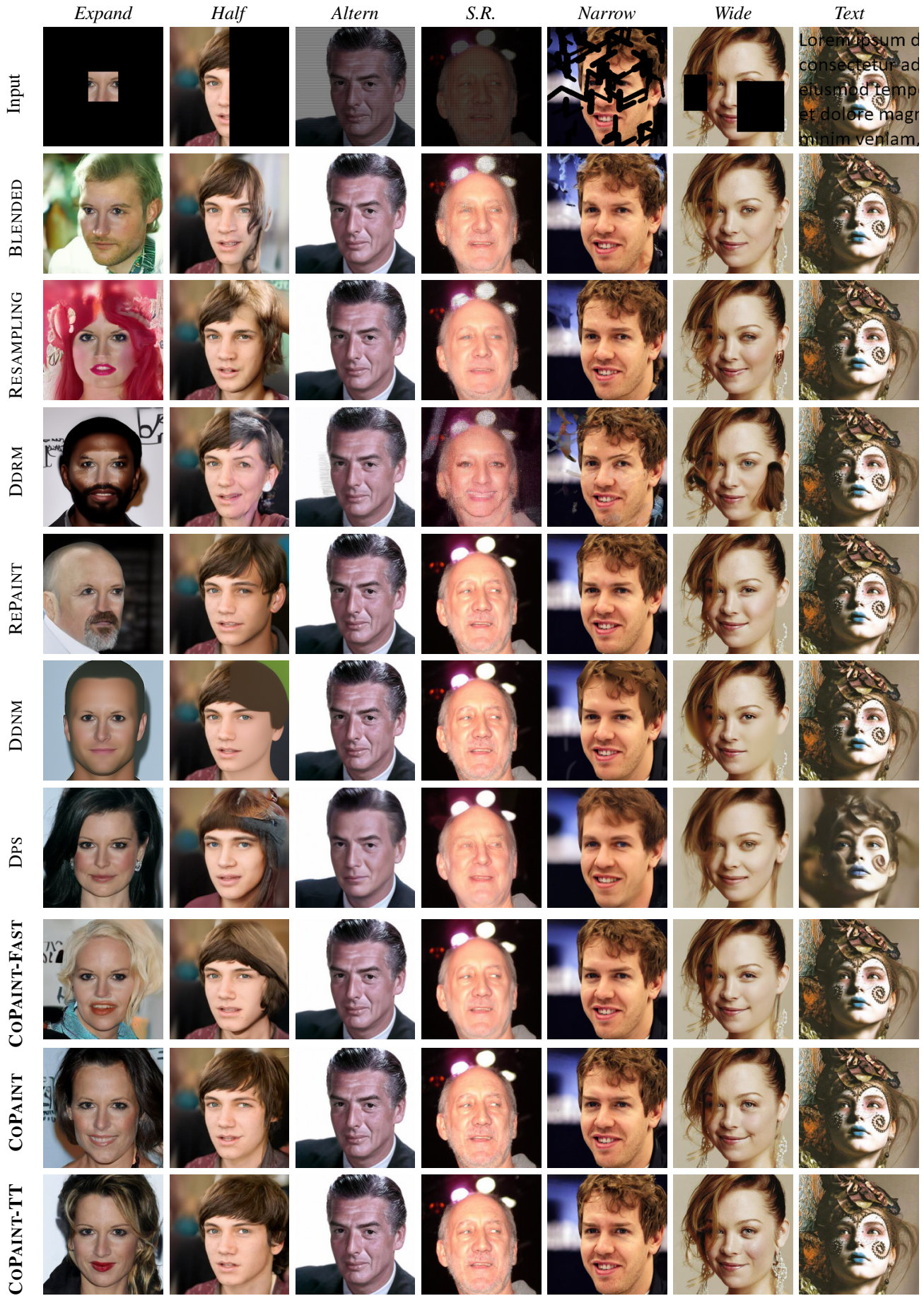

\begin{center}
\resizebox{\textwidth}{!}{
\begin{tabular}{cccccccc}
    & {\footnotesize{\textit{Expand}}} \hspace{-4mm}
    & {\footnotesize{\textit{Half}}} \hspace{-4mm}
    & {\footnotesize{\textit{Altern}}} \hspace{-4mm}
    & {\footnotesize{\textit{S.R.}}} \hspace{-4mm}
    & {\footnotesize{\textit{Narrow}}} \hspace{-4mm}
    & {\footnotesize{\textit{Wide}}} \hspace{-4mm}
    & {\footnotesize{\textit{Text}}} \\
    \hspace{-5mm} \rotatebox[origin=c]{90}{\footnotesize Input } \hspace{-4mm}
    & \multicolumn{1}{m{2cm}}{\includegraphics[width=2cm,height=2cm]{figures/celeba_qualitative/expand/masked.png}} \hspace{-4mm}
    & \multicolumn{1}{m{2cm}}{\includegraphics[width=2cm,height=2cm]{figures/celeba_qualitative/half/masked.png}} \hspace{-4mm}
    & \multicolumn{1}{m{2cm}}{\includegraphics[width=2cm,height=2cm]{figures/celeba_qualitative/altern/masked.png}} \hspace{-4mm}
    & \multicolumn{1}{m{2cm}}{\includegraphics[width=2cm,height=2cm]{figures/celeba_qualitative/superr/masked.png}} \hspace{-4mm}
    & \multicolumn{1}{m{2cm}}{\includegraphics[width=2cm,height=2cm]{figures/celeba_qualitative/narrow/masked.png}} \hspace{-4mm}
    & \multicolumn{1}{m{2cm}}{\includegraphics[width=2cm,height=2cm]{figures/celeba_qualitative/wide/masked.png}} \hspace{-4mm}
    & \multicolumn{1}{m{2cm}}{\includegraphics[width=2cm,height=2cm]{figures/celeba_qualitative/text/masked.png}} \hspace{-4mm}
    \\
\end{tabular}}
\end{center}
\vskip -0.1in
\caption{\footnotesize Qualitative results of baseline \textsc{Blended} and our methods (\textsc{CoPaint}, \textsc{CoPaint-TT}) on \texttt{CelebA-HQ} with seven masks.}
\label{fig:app-qualitative-celeba0}
\vskip -0.2in
\end{figure*}

\begin{figure*}[t]
\begin{center}
\resizebox{\textwidth}{!}{
\begin{tabular}{cccccccc}
    & {\footnotesize{\textit{Expand}}} \hspace{-4mm}
    & {\footnotesize{\textit{Half}}} \hspace{-4mm}
    & {\footnotesize{\textit{Altern}}} \hspace{-4mm}
    & {\footnotesize{\textit{S.R.}}} \hspace{-4mm}
    & {\footnotesize{\textit{Narrow}}} \hspace{-4mm}
    & {\footnotesize{\textit{Wide}}} \hspace{-4mm}
    & {\footnotesize{\textit{Text}}} \\
    \hspace{-5mm} \rotatebox[origin=c]{90}{\footnotesize Input } \hspace{-4mm}
    & \multicolumn{1}{m{2cm}}{\includegraphics[width=2cm,height=2cm]{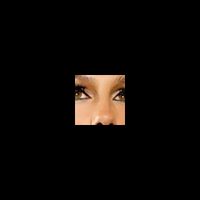}} \hspace{-4mm}
    & \multicolumn{1}{m{2cm}}{\includegraphics[width=2cm,height=2cm]{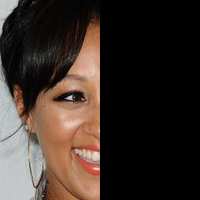}} \hspace{-4mm}
    & \multicolumn{1}{m{2cm}}{\includegraphics[width=2cm,height=2cm]{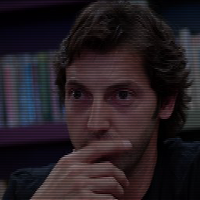}} \hspace{-4mm}
    & \multicolumn{1}{m{2cm}}{\includegraphics[width=2cm,height=2cm]{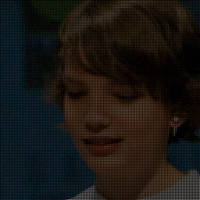}} \hspace{-4mm}
    & \multicolumn{1}{m{2cm}}{\includegraphics[width=2cm,height=2cm]{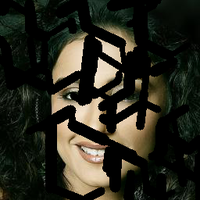}} \hspace{-4mm}
    & \multicolumn{1}{m{2cm}}{\includegraphics[width=2cm,height=2cm]{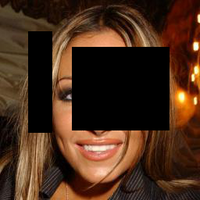}} \hspace{-4mm}
    & \multicolumn{1}{m{2cm}}{\includegraphics[width=2cm,height=2cm]{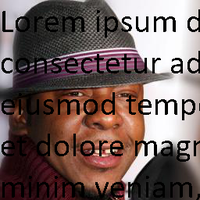}} \hspace{-4mm}
    \\
\end{tabular}}
\end{center}
\vskip -0.1in
\caption{\footnotesize Qualitative results of baseline methods and our methods (\textsc{CoPaint}, \textsc{CoPaint-TT}) on \texttt{CelebA-HQ} with seven masks.}
\label{fig:app-qualitative-celeba1}
\vskip -0.2in
\end{figure*}

\begin{figure*}[t]
\vskip 0.1in
\begin{center}
\resizebox{\textwidth}{!}{
\begin{tabular}{cccccccc}
    & {\footnotesize{\textit{Expand}}} \hspace{-4mm}
    & {\footnotesize{\textit{Half}}} \hspace{-4mm}
    & {\footnotesize{\textit{Altern}}} \hspace{-4mm}
    & {\footnotesize{\textit{S.R.}}} \hspace{-4mm}
    & {\footnotesize{\textit{Narrow}}} \hspace{-4mm}
    & {\footnotesize{\textit{Wide}}} \hspace{-4mm}
    & {\footnotesize{\textit{Text}}} \\
    \hspace{-5mm} \rotatebox[origin=c]{90}{\footnotesize Input } \hspace{-4mm}
    & \multicolumn{1}{m{2cm}}{\includegraphics[width=2cm,height=2cm]{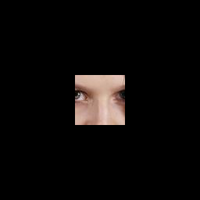}} \hspace{-4mm}
    & \multicolumn{1}{m{2cm}}{\includegraphics[width=2cm,height=2cm]{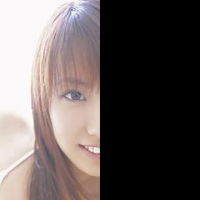}} \hspace{-4mm}
    & \multicolumn{1}{m{2cm}}{\includegraphics[width=2cm,height=2cm]{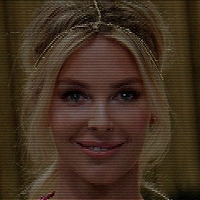}} \hspace{-4mm}
    & \multicolumn{1}{m{2cm}}{\includegraphics[width=2cm,height=2cm]{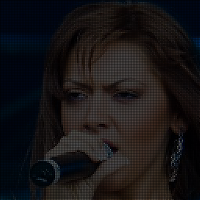}} \hspace{-4mm}
    & \multicolumn{1}{m{2cm}}{\includegraphics[width=2cm,height=2cm]{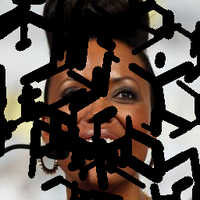}} \hspace{-4mm}
    & \multicolumn{1}{m{2cm}}{\includegraphics[width=2cm,height=2cm]{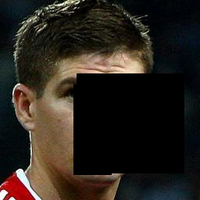}} \hspace{-4mm}
    & \multicolumn{1}{m{2cm}}{\includegraphics[width=2cm,height=2cm]{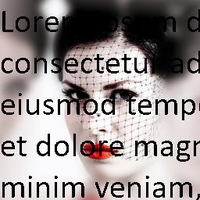}} \hspace{-4mm}
    \\
\end{tabular}}
\end{center}
\vskip -0.1in
\caption{\footnotesize Qualitative results of baseline methods and our methods (\textsc{CoPaint}, \textsc{CoPaint-TT}) on \texttt{CelebA-HQ} with seven masks.}
\label{fig:app-qualitative-celeba2}
\vskip -0.2in
\end{figure*}

\begin{figure*}[t]
\begin{center}
\resizebox{\textwidth}{!}{
\begin{tabular}{cccccccc}
    & {\footnotesize{\textit{Expand}}} \hspace{-4mm}
    & {\footnotesize{\textit{Half}}} \hspace{-4mm}
    & {\footnotesize{\textit{Altern}}} \hspace{-4mm}
    & {\footnotesize{\textit{S.R.}}} \hspace{-4mm}
    & {\footnotesize{\textit{Narrow}}} \hspace{-4mm}
    & {\footnotesize{\textit{Wide}}} \hspace{-4mm}
    & {\footnotesize{\textit{Text}}} \\
    \hspace{-5mm} \rotatebox[origin=c]{90}{\footnotesize Input } \hspace{-4mm}
    & \multicolumn{1}{m{2cm}}{\includegraphics[width=2cm,height=2cm]{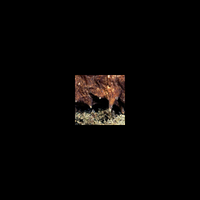}} \hspace{-4mm}
    & \multicolumn{1}{m{2cm}}{\includegraphics[width=2cm,height=2cm]{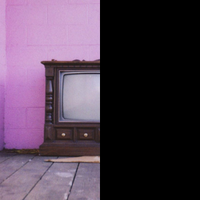}} \hspace{-4mm}
    & \multicolumn{1}{m{2cm}}{\includegraphics[width=2cm,height=2cm]{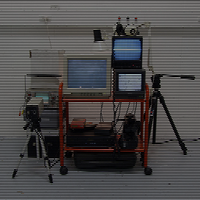}} \hspace{-4mm}
    & \multicolumn{1}{m{2cm}}{\includegraphics[width=2cm,height=2cm]{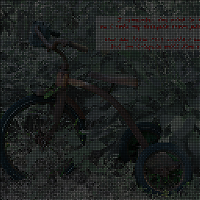}} \hspace{-4mm}
    & \multicolumn{1}{m{2cm}}{\includegraphics[width=2cm,height=2cm]{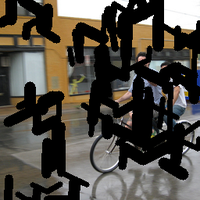}} \hspace{-4mm}
    & \multicolumn{1}{m{2cm}}{\includegraphics[width=2cm,height=2cm]{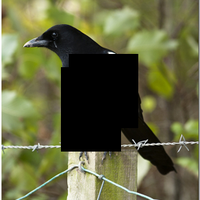}} \hspace{-4mm}
    & \multicolumn{1}{m{2cm}}{\includegraphics[width=2cm,height=2cm]{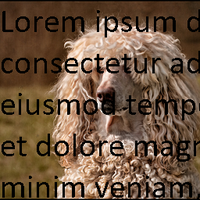}} \hspace{-4mm}
    \\
\end{tabular}}
\end{center}
\vskip -0.1in
\caption{\footnotesize Qualitative results of baseline \textsc{Blended} and our methods (\textsc{CoPaint}, \textsc{CoPaint-TT}) on \texttt{ImageNet} with seven masks.}
\label{fig:app-qualitative-imagenet0}
\vskip -0.2in
\end{figure*}

\begin{figure*}[t]
\begin{center}
\resizebox{\textwidth}{!}{
\begin{tabular}{cccccccc}
    & {\footnotesize{\textit{Expand}}} \hspace{-4mm}
    & {\footnotesize{\textit{Half}}} \hspace{-4mm}
    & {\footnotesize{\textit{Altern}}} \hspace{-4mm}
    & {\footnotesize{\textit{S.R.}}} \hspace{-4mm}
    & {\footnotesize{\textit{Narrow}}} \hspace{-4mm}
    & {\footnotesize{\textit{Wide}}} \hspace{-4mm}
    & {\footnotesize{\textit{Text}}} \\
    \hspace{-5mm} \rotatebox[origin=c]{90}{\footnotesize Input } \hspace{-4mm}
    & \multicolumn{1}{m{2cm}}{\includegraphics[width=2cm,height=2cm]{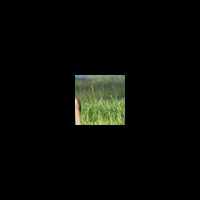}} \hspace{-4mm}
    & \multicolumn{1}{m{2cm}}{\includegraphics[width=2cm,height=2cm]{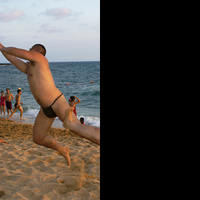}} \hspace{-4mm}
    & \multicolumn{1}{m{2cm}}{\includegraphics[width=2cm,height=2cm]{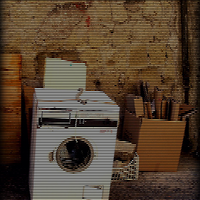}} \hspace{-4mm}
    & \multicolumn{1}{m{2cm}}{\includegraphics[width=2cm,height=2cm]{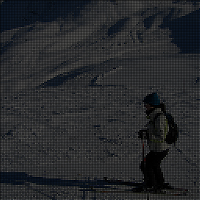}} \hspace{-4mm}
    & \multicolumn{1}{m{2cm}}{\includegraphics[width=2cm,height=2cm]{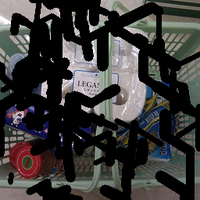}} \hspace{-4mm}
    & \multicolumn{1}{m{2cm}}{\includegraphics[width=2cm,height=2cm]{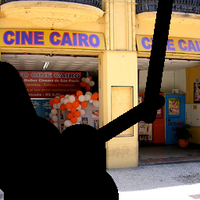}} \hspace{-4mm}
    & \multicolumn{1}{m{2cm}}{\includegraphics[width=2cm,height=2cm]{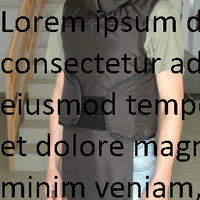}} \hspace{-4mm}
    \\
\end{tabular}}
\end{center}
\vskip -0.1in
\caption{\footnotesize Qualitative results of baseline methods and our methods (\textsc{CoPaint}, \textsc{CoPaint-TT}) on \texttt{ImageNet} with seven masks.}
\label{fig:app-qualitative-imagenet1}
\vskip -0.2in
\end{figure*}

\begin{figure*}[t]
\begin{center}
\resizebox{\textwidth}{!}{
\begin{tabular}{cccccccc}
    & {\footnotesize{\textit{Expand}}} \hspace{-4mm}
    & {\footnotesize{\textit{Half}}} \hspace{-4mm}
    & {\footnotesize{\textit{Altern}}} \hspace{-4mm}
    & {\footnotesize{\textit{S.R.}}} \hspace{-4mm}
    & {\footnotesize{\textit{Narrow}}} \hspace{-4mm}
    & {\footnotesize{\textit{Wide}}} \hspace{-4mm}
    & {\footnotesize{\textit{Text}}} \\
    \hspace{-5mm} \rotatebox[origin=c]{90}{\footnotesize Input } \hspace{-4mm}
    & \multicolumn{1}{m{2cm}}{\includegraphics[width=2cm,height=2cm]{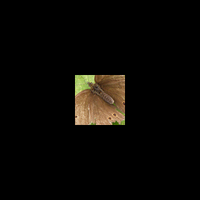}} \hspace{-4mm}
    & \multicolumn{1}{m{2cm}}{\includegraphics[width=2cm,height=2cm]{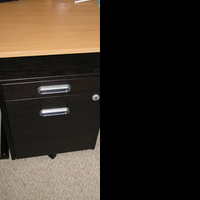}} \hspace{-4mm}
    & \multicolumn{1}{m{2cm}}{\includegraphics[width=2cm,height=2cm]{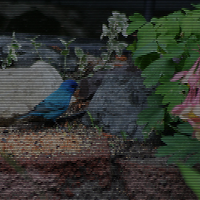}} \hspace{-4mm}
    & \multicolumn{1}{m{2cm}}{\includegraphics[width=2cm,height=2cm]{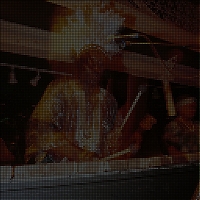}} \hspace{-4mm}
    & \multicolumn{1}{m{2cm}}{\includegraphics[width=2cm,height=2cm]{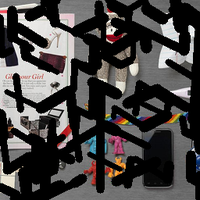}} \hspace{-4mm}
    & \multicolumn{1}{m{2cm}}{\includegraphics[width=2cm,height=2cm]{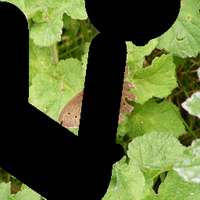}} \hspace{-4mm}
    & \multicolumn{1}{m{2cm}}{\includegraphics[width=2cm,height=2cm]{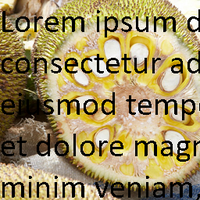}} \hspace{-4mm}
    \\
\end{tabular}}
\end{center}
\vskip -0.1in
\caption{\footnotesize Qualitative results of baseline methods and our methods (\textsc{CoPaint}, \textsc{CoPaint-TT}) on \texttt{ImageNet} with seven masks.}
\label{fig:app-qualitative-imagenet2}
\vskip -0.2in
\end{figure*}

We provide the larger size version of Figures~\ref{fig:qualitative_celebahq} and \ref{fig:qualitative_imagenet} in Figures~\ref{fig:app-qualitative-celeba0} and \ref{fig:app-qualitative-imagenet0}.
More qualitative results are further provided on \texttt{CelebA-HQ} in Figure~\ref{fig:app-qualitative-celeba1}, Figure~\ref{fig:app-qualitative-celeba2} and more qualitative results on \texttt{ImageNet} in Figure~\ref{fig:app-qualitative-imagenet1}, Figure~\ref{fig:app-qualitative-imagenet2} in this section. 



\clearpage

\section{Additional High-resolution Inpainting Experiments}
\label{app:high-resolution}
\begin{figure}[h]
    \centering
    \includegraphics[width=0.48\linewidth]{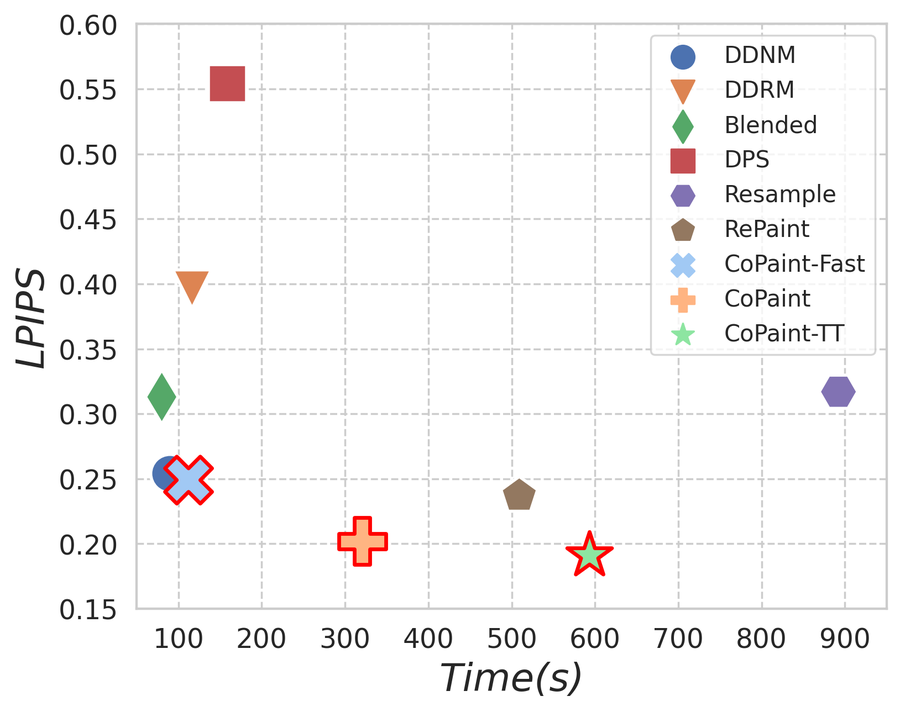}
    \caption{\footnotesize Time-performance trade-off on \texttt{ImageNet} for \e{512\times 512} inpainting. 
    The x-axis indicates the average time (\e{\downarrow}) to process one image, and the y-axis is the average \texttt{LPIPS} (\e{\downarrow}).
    }
    \label{fig:512time2metric}
    \vspace{-3mm}
\end{figure}

We conduct an additional experiment on inpainting images, where we use the released \e{512\times512} diffusion model\footnote{\url{https://github.com/openai/guided-diffusion}} pre-trained on \texttt{ImageNet} dataset as the backbone. 
The quantitative results could be found in Table~\ref{tab:high-resolution}. 
As can be observed, our methods still achieve the best \textit{LPIPS} compared with other baselines. 
For example, \textsc{CoPaint-TT} reduces \textit{LPIPS} by 19.4\% compared with the most competing baseline \textsc{RePaint}. 
In Figure~\ref{fig:512time2metric} with the time-performance tradeoff, we show that our method is able to outperform other baselines except for \textsc{RePaint} with a comparable computational time budget (\textsc{CoPaint-Fast}), and outperforms all baseline methods given more computational budget (\textsc{CoPaint} and \textsc{CoPaint-TT}). 
\begin{table*}[h!]
\caption{\footnotesize 
Quantitative results on \texttt{ImageNet} for $512\times512$ resolution inpainting. 
Lower is better for \texttt{LPIPS}. 
}
\vspace*{0.1in}
\label{tab:high-resolution}
\centering
\resizebox{0.8\textwidth}{!}{
\begin{threeparttable}
\begin{tabular}{@{}c|ccccccc|c@{}}
\toprule
\midrule
\multicolumn{9}{c}{\texttt{ImageNet-512}}
\\ 
\midrule
\multirow{2}{*}{Method}
& \multicolumn{1}{c}{\textit{Expand}}
& \multicolumn{1}{c}{\textit{Half}}
& \multicolumn{1}{c}{\textit{Altern}}
& \multicolumn{1}{c}{\textit{S.R.}}
& \multicolumn{1}{c}{\textit{Narrow}}
& \multicolumn{1}{c}{\textit{Wide}} 
& \multicolumn{1}{c}{\textit{Text}}
& \multicolumn{1}{|c}{\bf Average}
\\
& \texttt{LPIPS}$\downarrow$  
& \texttt{LPIPS}$\downarrow$
& \texttt{LPIPS}$\downarrow$
& \texttt{LPIPS}$\downarrow$
& \texttt{LPIPS}$\downarrow$
& \texttt{LPIPS}$\downarrow$
& \texttt{LPIPS}$\downarrow$
& \texttt{LPIPS}$\downarrow$
\\ 
\midrule
\multicolumn{1}{c|}{\textsc{Blended}}  & 0.739          & 0.377          & 0.210          & 0.495          & 0.157          & 0.179          & 0.038          & 0.313     
\\
\multicolumn{1}{c|}{\textsc{Ddrm}} & 0.859          & 0.391          & 0.339          & 0.712          & 0.204          & 0.197          & 0.073          & 0.396   
\\
\multicolumn{1}{c|}{\textsc{Resampling}}   & 0.799 & 0.366 & 0.205 & 0.482 & 0.157 & 0.173 & 0.039 & 0.317  
\\
\multicolumn{1}{c|}{\textsc{RePaint}} & 0.835          & 0.351          & 0.066          & 0.158          & \textbf{0.083} & 0.146          & \textbf{0.019} & 0.237
\\
\multicolumn{1}{c|}{\textsc{Dps}} &  0.750          & 0.575          & 0.513          & 0.543          & 0.496          & 0.519          & 0.480          & 0.554 
\\
\multicolumn{1}{c|}{\textsc{Ddnm}}    & 0.850          & 0.406          & 0.033          & 0.079          & 0.173          & 0.193          & 0.044          & 0.254    
\\
\rowcolor{gray!20}
\multicolumn{1}{c|}{\textsc{CoPaint-Fast}}   & 0.678 & 0.335 & 0.075 & 0.128 & 0.103 & 0.167 & 0.043 & 0.218 
\\
\rowcolor{gray!20}
\multicolumn{1}{c|}{\textsc{CoPaint}}   & 0.732          & 0.310          & 0.033          & 0.067          & 0.100          & 0.146          & 0.026          & 0.202   
\\
\rowcolor{gray!20}
\multicolumn{1}{c|}{\textsc{CoPaint-TT}}    & \textbf{0.726} & \textbf{0.292} & \textbf{0.022} & \textbf{0.043} & 0.093          & \textbf{0.136} & 0.025          & \textbf{0.191}   
\\
\toprule
\bottomrule
\end{tabular}
\end{threeparttable}
}
\end{table*}

\clearpage 

\section{Additional Super-resolution Experiments}
\label{app:super-resolution}
We conduct an additional experiment with our method on the super-resolution task.
Specifically, we apply average pooling to downsample a \e{256\times256} image to a lower resolution at different scales following \textsc{Ddnm}~\cite{ddnm} and then use different methods to reconstruct the original \e{256\times256} image.
We compare our method with \textsc{Dps}~\cite{dps}, \textsc{Ddrm}~\cite{ddrm}, and \textsc{ddnm}~\citep{ddnm} as they are suitable for the super-resolution task. 

\begin{figure*}[h!]
\centering
\resizebox{0.8\textwidth}{!}{
\begin{tabular}{ccccccc}
    & {\footnotesize{\textsc{Input}}} \hspace{-4mm}
    & {\footnotesize{\textsc{Dps}}} \hspace{-4mm}
    & {\footnotesize{\textsc{Ddrm}}} \hspace{-4mm}
    & {\footnotesize{\textsc{Ddnm}}} \hspace{-4mm}
    & {\footnotesize{\textsc{CoPaint}}} \hspace{-4mm}
    & {\footnotesize{\textsc{CoPaint-TT}}} \hspace{-4mm} \\
    \multirow{3}{*}{\rotatebox[origin=c]{90}{\textsc{$2\times$SR}} }
    & \multicolumn{1}{m{2cm}}{\includegraphics[width=2cm,height=2cm]{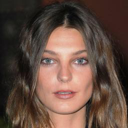}} \hspace{-4mm}
    & \multicolumn{1}{m{2cm}}{\includegraphics[width=2cm,height=2cm]{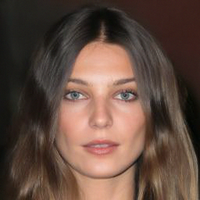}} \hspace{-4mm}
    & \multicolumn{1}{m{2cm}}{\includegraphics[width=2cm,height=2cm]{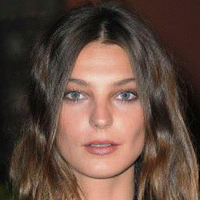}} \hspace{-4mm}
    & \multicolumn{1}{m{2cm}}{\includegraphics[width=2cm,height=2cm]{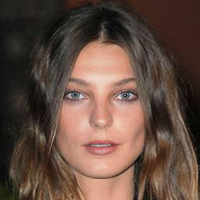}} \hspace{-4mm}
    & \multicolumn{1}{m{2cm}}{\includegraphics[width=2cm,height=2cm]{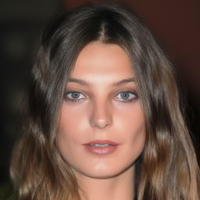}} \hspace{-4mm}
    & \multicolumn{1}{m{2cm}}{\includegraphics[width=2cm,height=2cm]{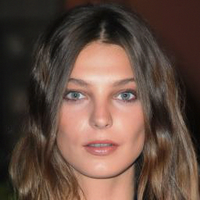}} \hspace{-4mm}
    \\
    & \multicolumn{1}{m{2cm}}{\includegraphics[width=2cm,height=2cm]{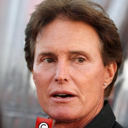}} \hspace{-4mm}
    & \multicolumn{1}{m{2cm}}{\includegraphics[width=2cm,height=2cm]{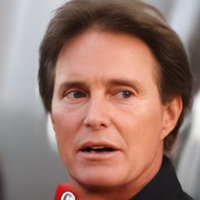}} \hspace{-4mm}
    & \multicolumn{1}{m{2cm}}{\includegraphics[width=2cm,height=2cm]{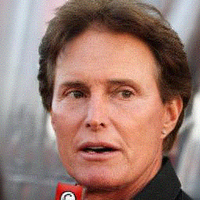}} \hspace{-4mm}
    & \multicolumn{1}{m{2cm}}{\includegraphics[width=2cm,height=2cm]{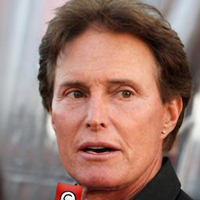}} \hspace{-4mm}
    & \multicolumn{1}{m{2cm}}{\includegraphics[width=2cm,height=2cm]{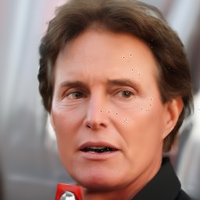}} \hspace{-4mm}
    & \multicolumn{1}{m{2cm}}{\includegraphics[width=2cm,height=2cm]{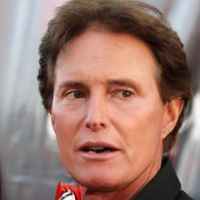}} \hspace{-4mm}
    \\
    & \multicolumn{1}{m{2cm}}{\includegraphics[width=2cm,height=2cm]{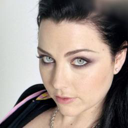}} \hspace{-4mm}
    & \multicolumn{1}{m{2cm}}{\includegraphics[width=2cm,height=2cm]{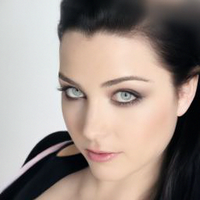}} \hspace{-4mm}
    & \multicolumn{1}{m{2cm}}{\includegraphics[width=2cm,height=2cm]{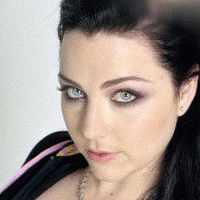}} \hspace{-4mm}
    & \multicolumn{1}{m{2cm}}{\includegraphics[width=2cm,height=2cm]{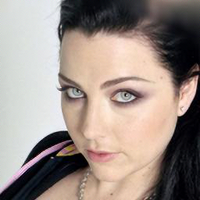}} \hspace{-4mm}
    & \multicolumn{1}{m{2cm}}{\includegraphics[width=2cm,height=2cm]{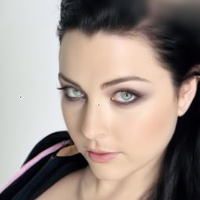}} \hspace{-4mm}
    & \multicolumn{1}{m{2cm}}{\includegraphics[width=2cm,height=2cm]{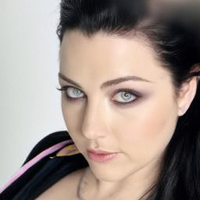}} \hspace{-4mm}
    \\
    \multirow{3}{*}{\rotatebox[origin=c]{90}{\textsc{$4\times$SR}}}
    & \multicolumn{1}{m{2cm}}{\includegraphics[width=2cm,height=2cm]{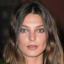}} \hspace{-4mm}
    & \multicolumn{1}{m{2cm}}{\includegraphics[width=2cm,height=2cm]{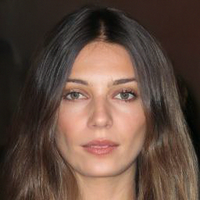}} \hspace{-4mm}
    & \multicolumn{1}{m{2cm}}{\includegraphics[width=2cm,height=2cm]{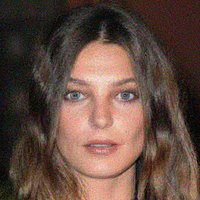}} \hspace{-4mm}
    & \multicolumn{1}{m{2cm}}{\includegraphics[width=2cm,height=2cm]{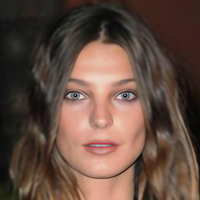}} \hspace{-4mm}
    & \multicolumn{1}{m{2cm}}{\includegraphics[width=2cm,height=2cm]{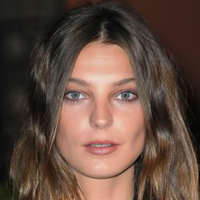}} \hspace{-4mm}
    & \multicolumn{1}{m{2cm}}{\includegraphics[width=2cm,height=2cm]{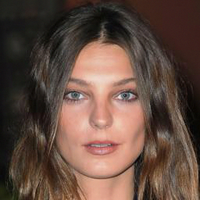}} \hspace{-4mm}
    \\
    & \multicolumn{1}{m{2cm}}{\includegraphics[width=2cm,height=2cm]{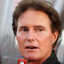}} \hspace{-4mm}
    & \multicolumn{1}{m{2cm}}{\includegraphics[width=2cm,height=2cm]{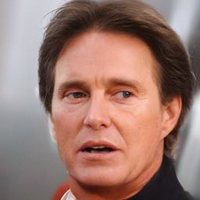}} \hspace{-4mm}
    & \multicolumn{1}{m{2cm}}{\includegraphics[width=2cm,height=2cm]{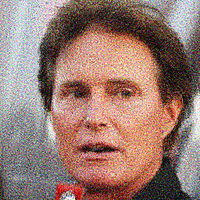}} \hspace{-4mm}
    & \multicolumn{1}{m{2cm}}{\includegraphics[width=2cm,height=2cm]{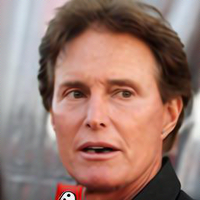}} \hspace{-4mm}
    & \multicolumn{1}{m{2cm}}{\includegraphics[width=2cm,height=2cm]{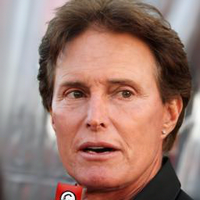}} \hspace{-4mm}
    & \multicolumn{1}{m{2cm}}{\includegraphics[width=2cm,height=2cm]{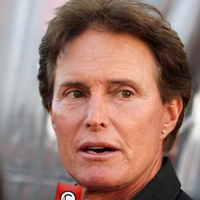}} \hspace{-4mm}
    \\
    & \multicolumn{1}{m{2cm}}{\includegraphics[width=2cm,height=2cm]{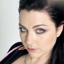}} \hspace{-4mm}
    & \multicolumn{1}{m{2cm}}{\includegraphics[width=2cm,height=2cm]{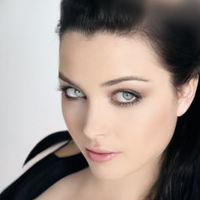}} \hspace{-4mm}
    & \multicolumn{1}{m{2cm}}{\includegraphics[width=2cm,height=2cm]{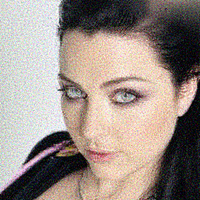}} \hspace{-4mm}
    & \multicolumn{1}{m{2cm}}{\includegraphics[width=2cm,height=2cm]{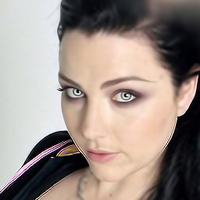}} \hspace{-4mm}
    & \multicolumn{1}{m{2cm}}{\includegraphics[width=2cm,height=2cm]{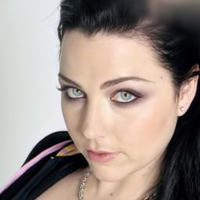}} \hspace{-4mm}
    & \multicolumn{1}{m{2cm}}{\includegraphics[width=2cm,height=2cm]{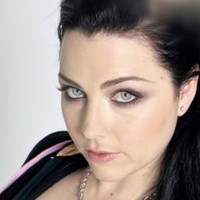}} \hspace{-4mm}
    \\
    \multirow{3}{*}{\rotatebox[origin=c]{90}{\textsc{$8\times$SR}}}
    & \multicolumn{1}{m{2cm}}{\includegraphics[width=2cm,height=2cm]{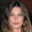}} \hspace{-4mm}
    & \multicolumn{1}{m{2cm}}{\includegraphics[width=2cm,height=2cm]{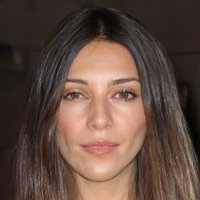}} \hspace{-4mm}
    & \multicolumn{1}{m{2cm}}{\includegraphics[width=2cm,height=2cm]{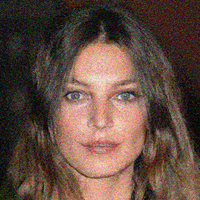}} \hspace{-4mm}
    & \multicolumn{1}{m{2cm}}{\includegraphics[width=2cm,height=2cm]{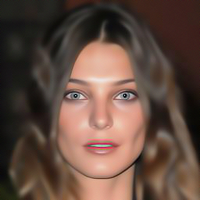}} \hspace{-4mm}
    & \multicolumn{1}{m{2cm}}{\includegraphics[width=2cm,height=2cm]{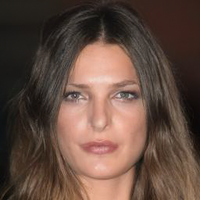}} \hspace{-4mm}
    & \multicolumn{1}{m{2cm}}{\includegraphics[width=2cm,height=2cm]{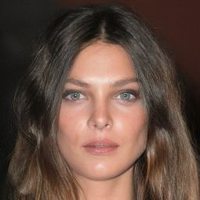}} \hspace{-4mm}
    \\
    & \multicolumn{1}{m{2cm}}{\includegraphics[width=2cm,height=2cm]{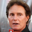}} \hspace{-4mm}
    & \multicolumn{1}{m{2cm}}{\includegraphics[width=2cm,height=2cm]{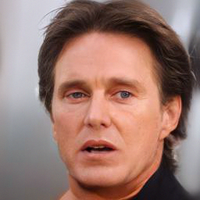}} \hspace{-4mm}
    & \multicolumn{1}{m{2cm}}{\includegraphics[width=2cm,height=2cm]{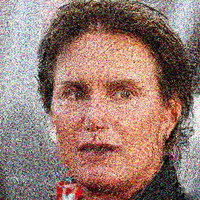}} \hspace{-4mm}
    & \multicolumn{1}{m{2cm}}{\includegraphics[width=2cm,height=2cm]{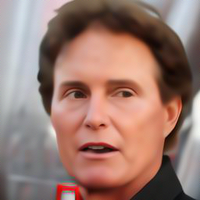}} \hspace{-4mm}
    & \multicolumn{1}{m{2cm}}{\includegraphics[width=2cm,height=2cm]{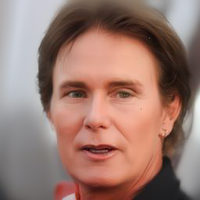}} \hspace{-4mm}
    & \multicolumn{1}{m{2cm}}{\includegraphics[width=2cm,height=2cm]{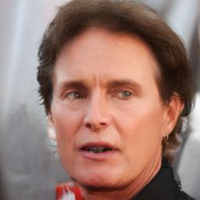}} \hspace{-4mm}
    \\
    & \multicolumn{1}{m{2cm}}{\includegraphics[width=2cm,height=2cm]{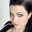}} \hspace{-4mm}
    & \multicolumn{1}{m{2cm}}{\includegraphics[width=2cm,height=2cm]{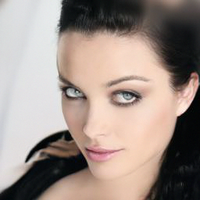}} \hspace{-4mm}
    & \multicolumn{1}{m{2cm}}{\includegraphics[width=2cm,height=2cm]{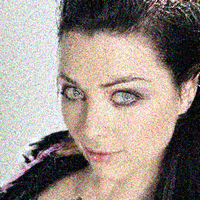}} \hspace{-4mm}
    & \multicolumn{1}{m{2cm}}{\includegraphics[width=2cm,height=2cm]{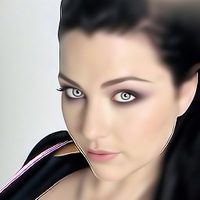}} \hspace{-4mm}
    & \multicolumn{1}{m{2cm}}{\includegraphics[width=2cm,height=2cm]{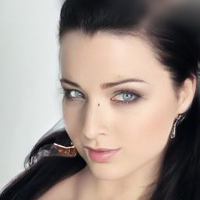}} \hspace{-4mm}
    & \multicolumn{1}{m{2cm}}{\includegraphics[width=2cm,height=2cm]{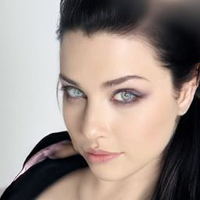}} \hspace{-4mm}
\end{tabular}
}
\caption{\footnotesize Qualitative results of applying different methods to super-resolution task on \texttt{CelebA-HQ} dataset.}
\label{fig:app-super-resolution-celeba}
\vskip -0.5in
\end{figure*}

\begin{figure*}[t!]
\centering
\resizebox{0.8\textwidth}{!}{
\begin{tabular}{ccccccc}
    & {\footnotesize{\textsc{Input}}} \hspace{-4mm}
    & {\footnotesize{\textsc{Dps}}} \hspace{-4mm}
    & {\footnotesize{\textsc{Ddrm}}} \hspace{-4mm}
    & {\footnotesize{\textsc{Ddnm}}} \hspace{-4mm}
    & {\footnotesize{\textsc{CoPaint}}} \hspace{-4mm}
    & {\footnotesize{\textsc{CoPaint-TT}}} \hspace{-4mm} \\
    \multirow{3}{*}{\rotatebox[origin=c]{90}{\textsc{$2\times$SR}}}
    & \multicolumn{1}{m{2cm}}{\includegraphics[width=2cm,height=2cm]{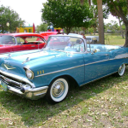}} \hspace{-4mm}
    & \multicolumn{1}{m{2cm}}{\includegraphics[width=2cm,height=2cm]{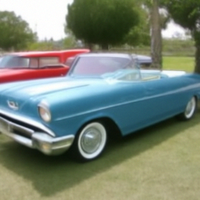}} \hspace{-4mm}
    & \multicolumn{1}{m{2cm}}{\includegraphics[width=2cm,height=2cm]{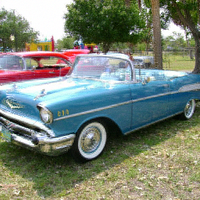}} \hspace{-4mm}
    & \multicolumn{1}{m{2cm}}{\includegraphics[width=2cm,height=2cm]{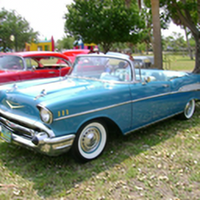}} \hspace{-4mm}
    & \multicolumn{1}{m{2cm}}{\includegraphics[width=2cm,height=2cm]{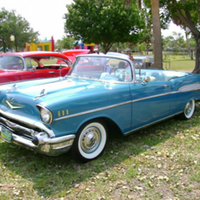}} \hspace{-4mm}
    & \multicolumn{1}{m{2cm}}{\includegraphics[width=2cm,height=2cm]{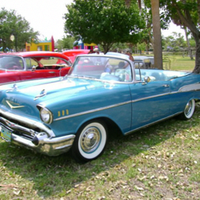}} \hspace{-4mm}
    \\
    & \multicolumn{1}{m{2cm}}{\includegraphics[width=2cm,height=2cm]{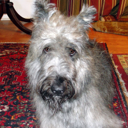}} \hspace{-4mm}
    & \multicolumn{1}{m{2cm}}{\includegraphics[width=2cm,height=2cm]{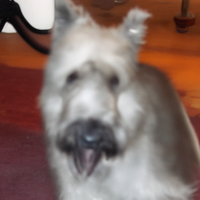}} \hspace{-4mm}
    & \multicolumn{1}{m{2cm}}{\includegraphics[width=2cm,height=2cm]{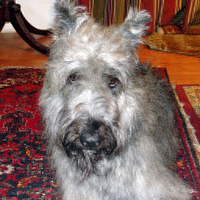}} \hspace{-4mm}
    & \multicolumn{1}{m{2cm}}{\includegraphics[width=2cm,height=2cm]{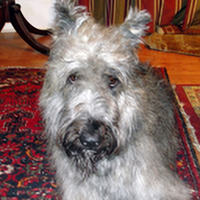}} \hspace{-4mm}
    & \multicolumn{1}{m{2cm}}{\includegraphics[width=2cm,height=2cm]{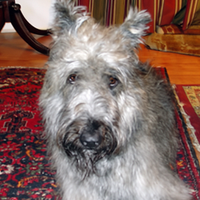}} \hspace{-4mm}
    & \multicolumn{1}{m{2cm}}{\includegraphics[width=2cm,height=2cm]{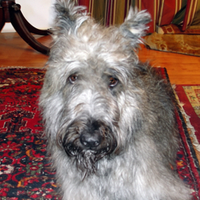}} \hspace{-4mm}
    \\
    & \multicolumn{1}{m{2cm}}{\includegraphics[width=2cm,height=2cm]{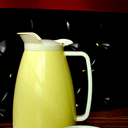}} \hspace{-4mm}
    & \multicolumn{1}{m{2cm}}{\includegraphics[width=2cm,height=2cm]{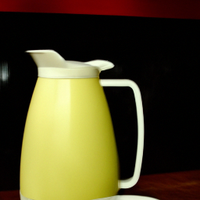}} \hspace{-4mm}
    & \multicolumn{1}{m{2cm}}{\includegraphics[width=2cm,height=2cm]{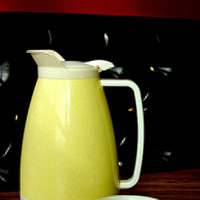}} \hspace{-4mm}
    & \multicolumn{1}{m{2cm}}{\includegraphics[width=2cm,height=2cm]{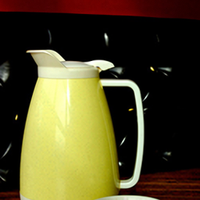}} \hspace{-4mm}
    & \multicolumn{1}{m{2cm}}{\includegraphics[width=2cm,height=2cm]{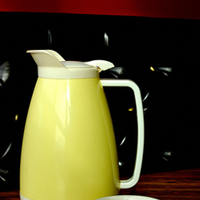}} \hspace{-4mm}
    & \multicolumn{1}{m{2cm}}{\includegraphics[width=2cm,height=2cm]{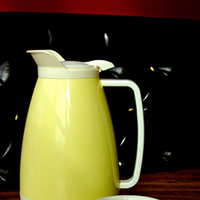}} \hspace{-4mm}
    \\
    \multirow{3}{*}{\rotatebox[origin=c]{90}{\textsc{$4\times$SR}}}
    & \multicolumn{1}{m{2cm}}{\includegraphics[width=2cm,height=2cm]{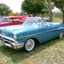}} \hspace{-4mm}
    & \multicolumn{1}{m{2cm}}{\includegraphics[width=2cm,height=2cm]{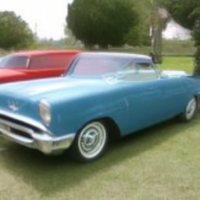}} \hspace{-4mm}
    & \multicolumn{1}{m{2cm}}{\includegraphics[width=2cm,height=2cm]{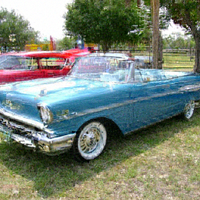}} \hspace{-4mm}
    & \multicolumn{1}{m{2cm}}{\includegraphics[width=2cm,height=2cm]{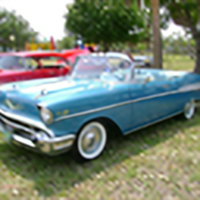}} \hspace{-4mm}
    & \multicolumn{1}{m{2cm}}{\includegraphics[width=2cm,height=2cm]{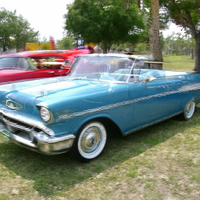}} \hspace{-4mm}
    & \multicolumn{1}{m{2cm}}{\includegraphics[width=2cm,height=2cm]{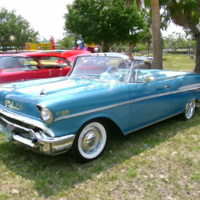}} \hspace{-4mm}
    \\
    & \multicolumn{1}{m{2cm}}{\includegraphics[width=2cm,height=2cm]{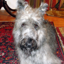}} \hspace{-4mm}
    & \multicolumn{1}{m{2cm}}{\includegraphics[width=2cm,height=2cm]{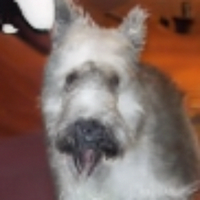}} \hspace{-4mm}
    & \multicolumn{1}{m{2cm}}{\includegraphics[width=2cm,height=2cm]{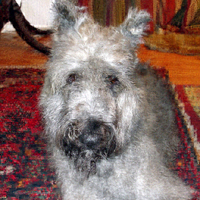}} \hspace{-4mm}
    & \multicolumn{1}{m{2cm}}{\includegraphics[width=2cm,height=2cm]{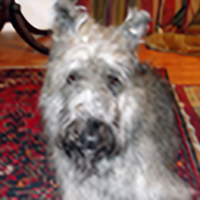}} \hspace{-4mm}
    & \multicolumn{1}{m{2cm}}{\includegraphics[width=2cm,height=2cm]{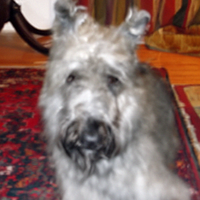}} \hspace{-4mm}
    & \multicolumn{1}{m{2cm}}{\includegraphics[width=2cm,height=2cm]{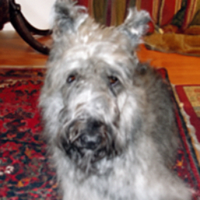}} \hspace{-4mm}
    \\
    & \multicolumn{1}{m{2cm}}{\includegraphics[width=2cm,height=2cm]{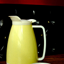}} \hspace{-4mm}
    & \multicolumn{1}{m{2cm}}{\includegraphics[width=2cm,height=2cm]{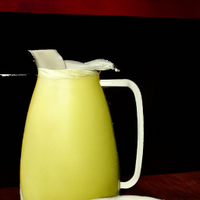}} \hspace{-4mm}
    & \multicolumn{1}{m{2cm}}{\includegraphics[width=2cm,height=2cm]{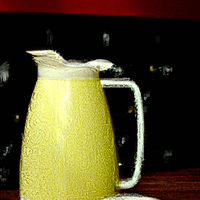}} \hspace{-4mm}
    & \multicolumn{1}{m{2cm}}{\includegraphics[width=2cm,height=2cm]{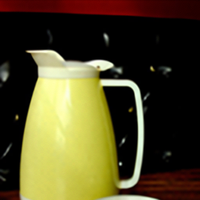}} \hspace{-4mm}
    & \multicolumn{1}{m{2cm}}{\includegraphics[width=2cm,height=2cm]{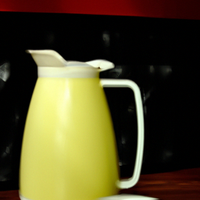}} \hspace{-4mm}
    & \multicolumn{1}{m{2cm}}{\includegraphics[width=2cm,height=2cm]{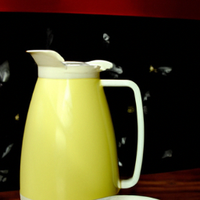}} \hspace{-4mm}
    \\
    \multirow{3}{*}{\rotatebox[origin=c]{90}{\textsc{$8\times$SR}}}
    & \multicolumn{1}{m{2cm}}{\includegraphics[width=2cm,height=2cm]{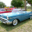}} \hspace{-4mm}
    & \multicolumn{1}{m{2cm}}{\includegraphics[width=2cm,height=2cm]{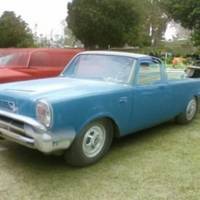}} \hspace{-4mm}
    & \multicolumn{1}{m{2cm}}{\includegraphics[width=2cm,height=2cm]{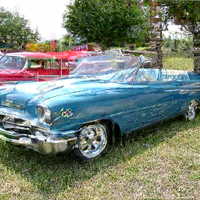}} \hspace{-4mm}
    & \multicolumn{1}{m{2cm}}{\includegraphics[width=2cm,height=2cm]{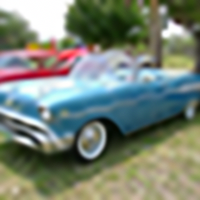}} \hspace{-4mm}
    & \multicolumn{1}{m{2cm}}{\includegraphics[width=2cm,height=2cm]{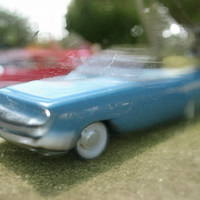}} \hspace{-4mm}
    & \multicolumn{1}{m{2cm}}{\includegraphics[width=2cm,height=2cm]{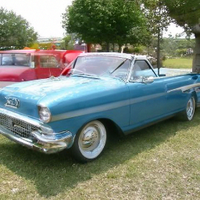}} \hspace{-4mm}
    \\
    & \multicolumn{1}{m{2cm}}{\includegraphics[width=2cm,height=2cm]{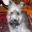}} \hspace{-4mm}
    & \multicolumn{1}{m{2cm}}{\includegraphics[width=2cm,height=2cm]{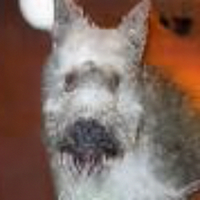}} \hspace{-4mm}
    & \multicolumn{1}{m{2cm}}{\includegraphics[width=2cm,height=2cm]{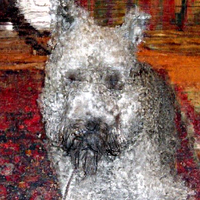}} \hspace{-4mm}
    & \multicolumn{1}{m{2cm}}{\includegraphics[width=2cm,height=2cm]{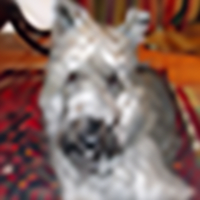}} \hspace{-4mm}
    & \multicolumn{1}{m{2cm}}{\includegraphics[width=2cm,height=2cm]{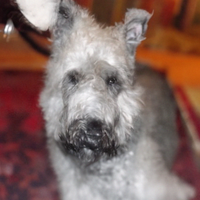}} \hspace{-4mm}
    & \multicolumn{1}{m{2cm}}{\includegraphics[width=2cm,height=2cm]{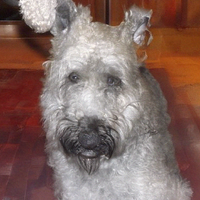}} \hspace{-4mm}
    \\
    & \multicolumn{1}{m{2cm}}{\includegraphics[width=2cm,height=2cm]{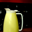}} \hspace{-4mm}
    & \multicolumn{1}{m{2cm}}{\includegraphics[width=2cm,height=2cm]{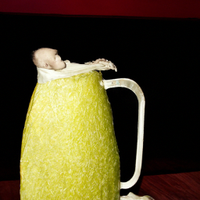}} \hspace{-4mm}
    & \multicolumn{1}{m{2cm}}{\includegraphics[width=2cm,height=2cm]{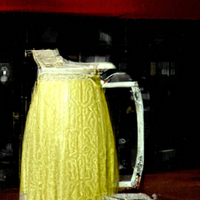}} \hspace{-4mm}
    & \multicolumn{1}{m{2cm}}{\includegraphics[width=2cm,height=2cm]{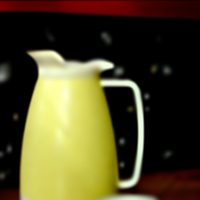}} \hspace{-4mm}
    & \multicolumn{1}{m{2cm}}{\includegraphics[width=2cm,height=2cm]{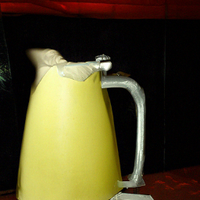}} \hspace{-4mm}
    & \multicolumn{1}{m{2cm}}{\includegraphics[width=2cm,height=2cm]{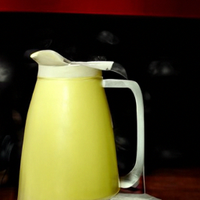}} \hspace{-4mm}
    \\
\end{tabular}
}
\caption{\footnotesize Qualitative results of applying different methods to super-resolution task on \texttt{ImageNet} dataset.}
\label{fig:app-super-resolution-imagenet}
\end{figure*}

The quantitative results in Table~\ref{tab:super-resolution} demonstrate the consistent superiority of our method compared with other baselines. 
The qualitative results are shown in Figures~\ref{fig:app-super-resolution-celeba} and \ref{fig:app-super-resolution-imagenet}.
Although the most competing baseline \textsc{Ddnm} performs well in \e{2\times} and \e{4\times} super-resolution, their generated images in \e{8\times} super-resolution are more blurry and lack finer details such as hair, as demonstrated in the first \texttt{CelebA-HQ} example, and fur, as demonstrated in the second \texttt{ImageNet} example. 
In contrast, our method produces more natural-looking images with better details. 
\begin{table*}[h!]
\centering
\caption{\footnotesize Quantitative results of super-resolution task on \texttt{CelebA-HQ}(top) and \texttt{ImageNet} (\emph{bottom}) datasets. Following~\cite{ddnm}, we apply average-pooling to a $256\times256$ image to obtain the low-resolution input and then reconstruct the original image using different methods. We perform experiments for three different scales, \textit{i.e.}, \textsc{$2\times$}, \textsc{$4\times$} and \textsc{$8\times$},  with the image being downsampled at the corresponding scale.
We report the objective metric \texttt{LPIPS} of each baseline. Lower is better for \texttt{LPIPS}. 
}
\vspace*{0.1in}
\label{tab:super-resolution}
\resizebox{0.6\textwidth}{!}{
\begin{threeparttable}
\begin{tabular}{@{}c|ccccc@{}}
\toprule
\midrule
\multicolumn{6}{c}{\texttt{ImageNet}}
\\ 
\midrule
\multirow{2}{*}{Scale Factor}
& \multicolumn{1}{c}{\textsc{Dps}}
& \multicolumn{1}{c}{\textsc{Ddrm}}
& \multicolumn{1}{c}{\textsc{Ddnm}}
& \multicolumn{1}{c}{\textsc{CoPaint}} 
& \multicolumn{1}{c}{\textsc{CoPaint-TT}}
\\
& {\footnotesize \texttt{LPIPS}$\downarrow$ }  
& {\footnotesize \texttt{LPIPS}$\downarrow$ }
& {\footnotesize \texttt{LPIPS}$\downarrow$ }
& {\footnotesize \texttt{LPIPS}$\downarrow$ }
& {\footnotesize \texttt{LPIPS}$\downarrow$ }
\\ 
\midrule
\multicolumn{1}{c|}{\textsc{$2\times$}}  & 0.156                & 0.054                & 0.031                & 0.037                & \textbf{0.025}
\\
\multicolumn{1}{c|}{\textsc{$4\times$}} &  0.190                & 0.228                & 0.141                & 0.113                & \textbf{0.082} 
\\
\multicolumn{1}{c|}{\textsc{$8\times$}} & 0.235                & 0.360                & 0.250                & 0.293                & \textbf{0.170} \\
\midrule

\multicolumn{6}{c}{\texttt{CelebA-HQ}}
\\ 
\midrule
\multirow{2}{*}{Scale Factor}
& \multicolumn{1}{c}{\textsc{Dps}}
& \multicolumn{1}{c}{\textsc{Ddrm}}
& \multicolumn{1}{c}{\textsc{Ddnm}}
& \multicolumn{1}{c}{\textsc{CoPaint}} 
& \multicolumn{1}{c}{\textsc{CoPaint-TT}}
\\
& \texttt{LPIPS}$\downarrow$  
& \texttt{LPIPS}$\downarrow$
& \texttt{LPIPS}$\downarrow$
& \texttt{LPIPS}$\downarrow$
& \texttt{LPIPS}$\downarrow$
\\ 
\midrule
\multicolumn{1}{c|}{\textsc{$2\times$}}  & 0.417                & 0.121                & 0.113                & 0.063                & \textbf{0.042}  \\
\multicolumn{1}{c|}{\textsc{$4\times$}} & 0.483                & 0.345                & 0.328                & 0.252                & \textbf{0.204}   \\
\multicolumn{1}{c|}{\textsc{$8\times$}} & 0.531                & 0.480                & 0.528                & 0.511                & \textbf{0.423}    \\
\toprule
\bottomrule
\end{tabular}
\end{threeparttable}
}
\end{table*}

\clearpage

\clearpage 

\section{Failure Case Study}
\label{app:fail}
\begin{figure*}[h!]
\begin{center}
\resizebox{\textwidth}{!}{
\begin{tabular}{ccccc|cccc}
    &
    \multicolumn{4}{c}{\texttt{Incorerct details}} & 
    \multicolumn{4}{c}{\texttt{Large Mask}} 
    \\
    \hspace{-5mm} \rotatebox[origin=c]{90}{\footnotesize Input } \hspace{-4mm}
    & \multicolumn{1}{m{2cm}}{\includegraphics[width=2cm,height=2cm]{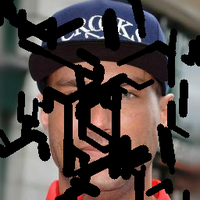}} \hspace{-4mm}
    & \multicolumn{1}{m{2cm}}{\includegraphics[width=2cm,height=2cm]{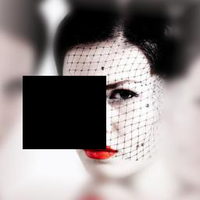}} \hspace{-4mm}
    & \multicolumn{1}{m{2cm}}{\includegraphics[width=2cm,height=2cm]{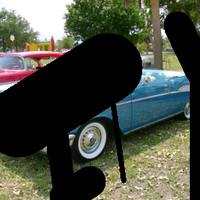}} \hspace{-4mm}
    & \multicolumn{1}{m{2cm}}{\includegraphics[width=2cm,height=2cm]{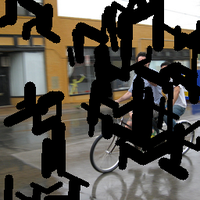}} \hspace{-4mm}
    & \multicolumn{1}{m{2cm}}{\includegraphics[width=2cm,height=2cm]{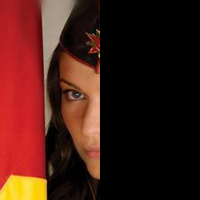}} \hspace{-4mm}
    & \multicolumn{1}{m{2cm}}{\includegraphics[width=2cm,height=2cm]{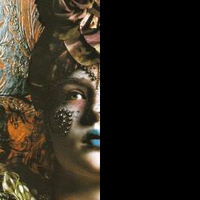}} \hspace{-4mm}
    & \multicolumn{1}{m{2cm}}{\includegraphics[width=2cm,height=2cm]{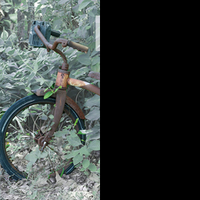}} \hspace{-4mm}
    & \multicolumn{1}{m{2cm}}{\includegraphics[width=2cm,height=2cm]{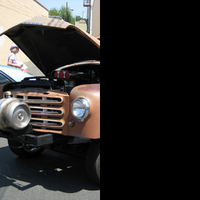}} \hspace{-4mm}
    \\
    \hspace{-5mm} \rotatebox[origin=c]{90}{\footnotesize \textsc{Blended} } \hspace{-4mm}
    & \multicolumn{1}{m{2cm}}{\includegraphics[width=2cm,height=2cm]{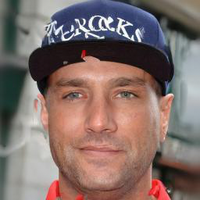}} \hspace{-4mm}
    & \multicolumn{1}{m{2cm}}{\includegraphics[width=2cm,height=2cm]{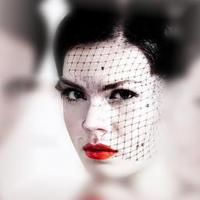}} \hspace{-4mm}
    & \multicolumn{1}{m{2cm}}{\includegraphics[width=2cm,height=2cm]{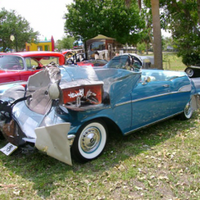}} \hspace{-4mm}
    & \multicolumn{1}{m{2cm}}{\includegraphics[width=2cm,height=2cm]{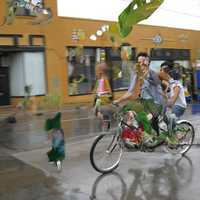}} \hspace{-4mm}
    & \multicolumn{1}{m{2cm}}{\includegraphics[width=2cm,height=2cm]{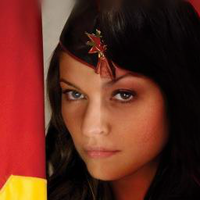}} \hspace{-4mm}
    & \multicolumn{1}{m{2cm}}{\includegraphics[width=2cm,height=2cm]{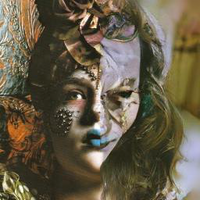}} \hspace{-4mm}
    & \multicolumn{1}{m{2cm}}{\includegraphics[width=2cm,height=2cm]{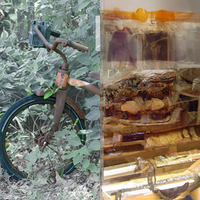}} \hspace{-4mm}
    & \multicolumn{1}{m{2cm}}{\includegraphics[width=2cm,height=2cm]{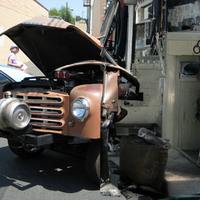}} \hspace{-4mm}
    \\
    \hspace{-5mm} \rotatebox[origin=c]{90}{\footnotesize \textsc{Resampling} } \hspace{-4mm}
    & \multicolumn{1}{m{2cm}}{\includegraphics[width=2cm,height=2cm]{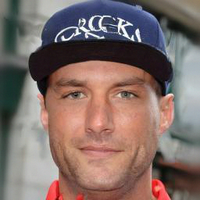}} \hspace{-4mm}
    & \multicolumn{1}{m{2cm}}{\includegraphics[width=2cm,height=2cm]{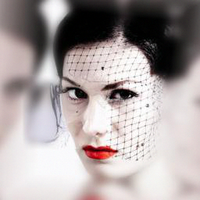}} \hspace{-4mm}
    & \multicolumn{1}{m{2cm}}{\includegraphics[width=2cm,height=2cm]{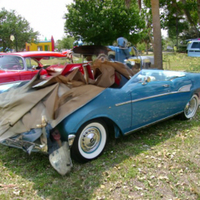}} \hspace{-4mm}
    & \multicolumn{1}{m{2cm}}{\includegraphics[width=2cm,height=2cm]{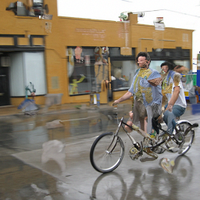}} \hspace{-4mm}
    & \multicolumn{1}{m{2cm}}{\includegraphics[width=2cm,height=2cm]{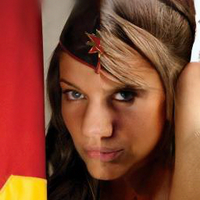}} \hspace{-4mm}
    & \multicolumn{1}{m{2cm}}{\includegraphics[width=2cm,height=2cm]{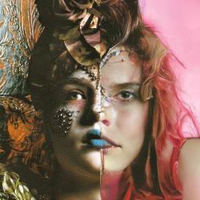}} \hspace{-4mm}
    & \multicolumn{1}{m{2cm}}{\includegraphics[width=2cm,height=2cm]{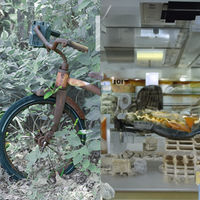}} \hspace{-4mm}
    & \multicolumn{1}{m{2cm}}{\includegraphics[width=2cm,height=2cm]{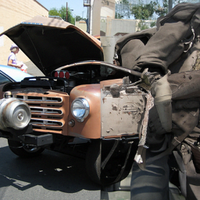}} \hspace{-4mm}
    \\
    \hspace{-5mm} \rotatebox[origin=c]{90}{\footnotesize \textsc{Ddrm} } \hspace{-4mm}
    & \multicolumn{1}{m{2cm}}{\includegraphics[width=2cm,height=2cm]{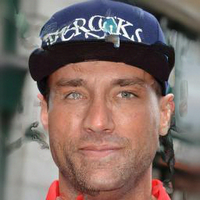}} \hspace{-4mm}
    & \multicolumn{1}{m{2cm}}{\includegraphics[width=2cm,height=2cm]{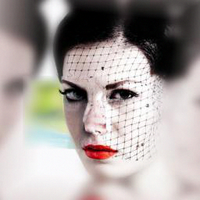}} \hspace{-4mm}
    & \multicolumn{1}{m{2cm}}{\includegraphics[width=2cm,height=2cm]{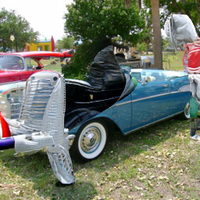}} \hspace{-4mm}
    & \multicolumn{1}{m{2cm}}{\includegraphics[width=2cm,height=2cm]{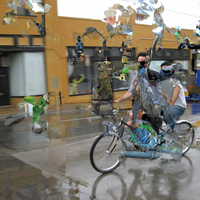}} \hspace{-4mm}
    & \multicolumn{1}{m{2cm}}{\includegraphics[width=2cm,height=2cm]{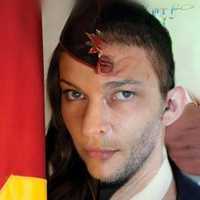}} \hspace{-4mm}
    & \multicolumn{1}{m{2cm}}{\includegraphics[width=2cm,height=2cm]{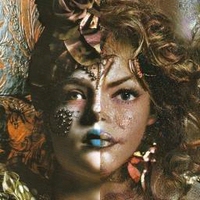}} \hspace{-4mm}
    & \multicolumn{1}{m{2cm}}{\includegraphics[width=2cm,height=2cm]{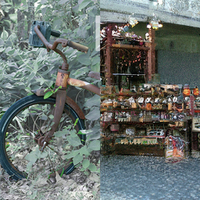}} \hspace{-4mm}
    & \multicolumn{1}{m{2cm}}{\includegraphics[width=2cm,height=2cm]{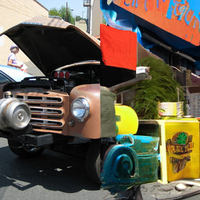}} \hspace{-4mm}
    \\
    \hspace{-5mm} \rotatebox[origin=c]{90}{\footnotesize \textsc{RePaint} } \hspace{-4mm}
    & \multicolumn{1}{m{2cm}}{\includegraphics[width=2cm,height=2cm]{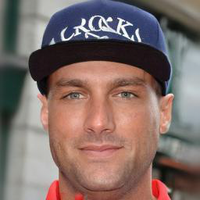}} \hspace{-4mm}
    & \multicolumn{1}{m{2cm}}{\includegraphics[width=2cm,height=2cm]{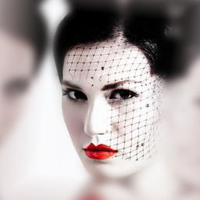}} \hspace{-4mm}
    & \multicolumn{1}{m{2cm}}{\includegraphics[width=2cm,height=2cm]{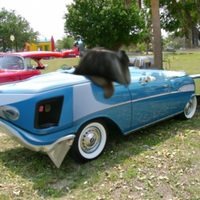}} \hspace{-4mm}
    & \multicolumn{1}{m{2cm}}{\includegraphics[width=2cm,height=2cm]{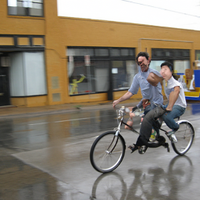}} \hspace{-4mm}
    & \multicolumn{1}{m{2cm}}{\includegraphics[width=2cm,height=2cm]{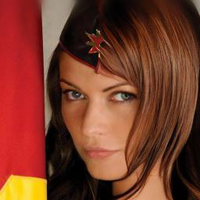}} \hspace{-4mm}
    & \multicolumn{1}{m{2cm}}{\includegraphics[width=2cm,height=2cm]{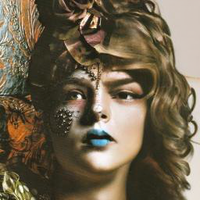}} \hspace{-4mm}
    & \multicolumn{1}{m{2cm}}{\includegraphics[width=2cm,height=2cm]{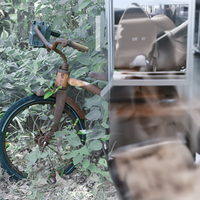}} \hspace{-4mm}
    & \multicolumn{1}{m{2cm}}{\includegraphics[width=2cm,height=2cm]{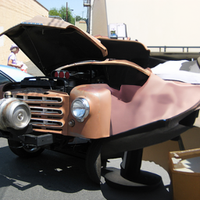}} \hspace{-4mm}
    \\
    \hspace{-5mm} \rotatebox[origin=c]{90}{\footnotesize \textsc{Ddnm} } \hspace{-4mm}
    & \multicolumn{1}{m{2cm}}{\includegraphics[width=2cm,height=2cm]{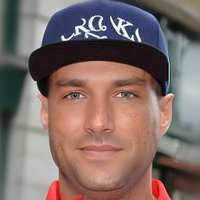}} \hspace{-4mm}
    & \multicolumn{1}{m{2cm}}{\includegraphics[width=2cm,height=2cm]{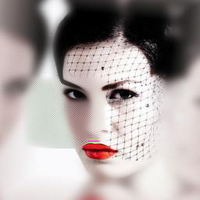}} \hspace{-4mm}
    & \multicolumn{1}{m{2cm}}{\includegraphics[width=2cm,height=2cm]{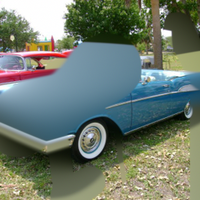}} \hspace{-4mm}
    & \multicolumn{1}{m{2cm}}{\includegraphics[width=2cm,height=2cm]{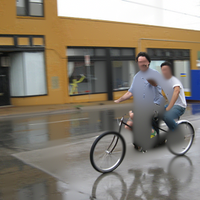}} \hspace{-4mm}
    & \multicolumn{1}{m{2cm}}{\includegraphics[width=2cm,height=2cm]{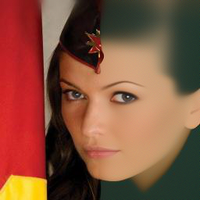}} \hspace{-4mm}
    & \multicolumn{1}{m{2cm}}{\includegraphics[width=2cm,height=2cm]{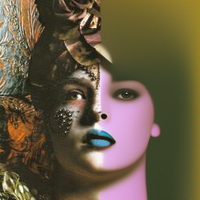}} \hspace{-4mm}
    & \multicolumn{1}{m{2cm}}{\includegraphics[width=2cm,height=2cm]{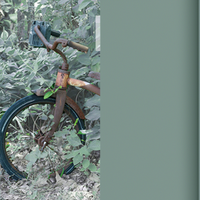}} \hspace{-4mm}
    & \multicolumn{1}{m{2cm}}{\includegraphics[width=2cm,height=2cm]{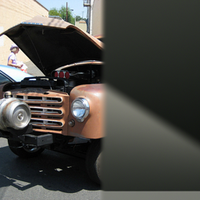}} \hspace{-4mm}
    \\
    \hspace{-5mm} \rotatebox[origin=c]{90}{\footnotesize \textsc{Dps} } \hspace{-4mm}
    & \multicolumn{1}{m{2cm}}{\includegraphics[width=2cm,height=2cm]{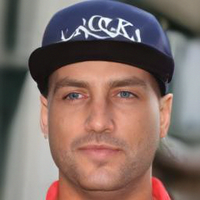}} \hspace{-4mm}
    & \multicolumn{1}{m{2cm}}{\includegraphics[width=2cm,height=2cm]{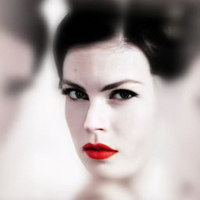}} \hspace{-4mm}
    & \multicolumn{1}{m{2cm}}{\includegraphics[width=2cm,height=2cm]{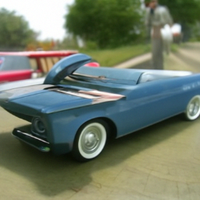}} \hspace{-4mm}
    & \multicolumn{1}{m{2cm}}{\includegraphics[width=2cm,height=2cm]{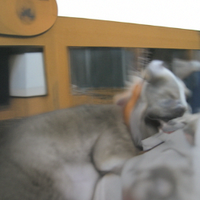}} \hspace{-4mm}
    & \multicolumn{1}{m{2cm}}{\includegraphics[width=2cm,height=2cm]{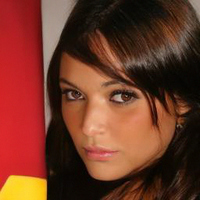}} \hspace{-4mm}
    & \multicolumn{1}{m{2cm}}{\includegraphics[width=2cm,height=2cm]{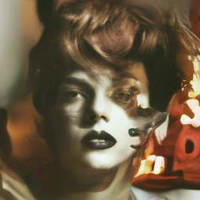}} \hspace{-4mm}
    & \multicolumn{1}{m{2cm}}{\includegraphics[width=2cm,height=2cm]{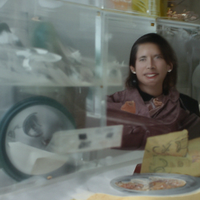}} \hspace{-4mm}
    & \multicolumn{1}{m{2cm}}{\includegraphics[width=2cm,height=2cm]{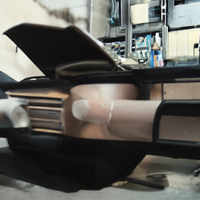}} \hspace{-4mm}
    \\
    \hspace{-5mm} \rotatebox[origin=c]{90}{\footnotesize \bf \textsc{CoPaint-Fast} } \hspace{-4mm}
    & \multicolumn{1}{m{2cm}}{\includegraphics[width=2cm,height=2cm]{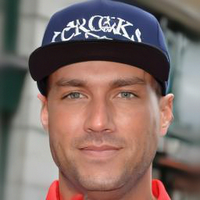}} \hspace{-4mm}
    & \multicolumn{1}{m{2cm}}{\includegraphics[width=2cm,height=2cm]{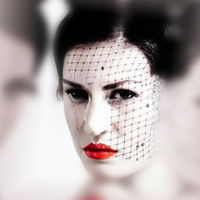}} \hspace{-4mm}
    & \multicolumn{1}{m{2cm}}{\includegraphics[width=2cm,height=2cm]{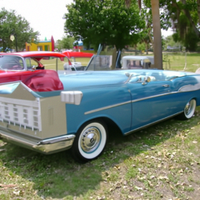}} \hspace{-4mm}
    & \multicolumn{1}{m{2cm}}{\includegraphics[width=2cm,height=2cm]{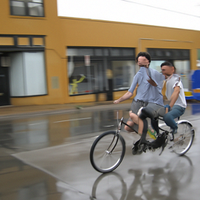}} \hspace{-4mm}
    & \multicolumn{1}{m{2cm}}{\includegraphics[width=2cm,height=2cm]{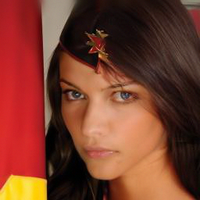}} \hspace{-4mm}
    & \multicolumn{1}{m{2cm}}{\includegraphics[width=2cm,height=2cm]{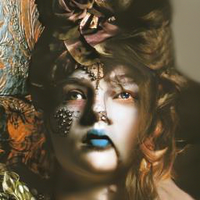}} \hspace{-4mm}
    & \multicolumn{1}{m{2cm}}{\includegraphics[width=2cm,height=2cm]{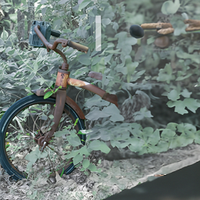}} \hspace{-4mm}
    & \multicolumn{1}{m{2cm}}{\includegraphics[width=2cm,height=2cm]{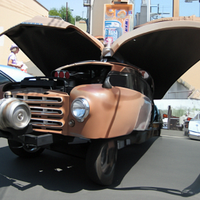}} \hspace{-4mm}
    \\
    \hspace{-5mm} \rotatebox[origin=c]{90}{\footnotesize \bf \textsc{CoPaint} } \hspace{-4mm}
    & \multicolumn{1}{m{2cm}}{\includegraphics[width=2cm,height=2cm]{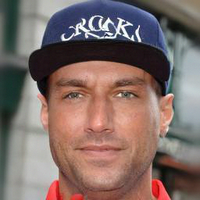}} \hspace{-4mm}
    & \multicolumn{1}{m{2cm}}{\includegraphics[width=2cm,height=2cm]{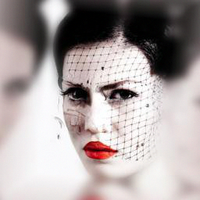}} \hspace{-4mm}
    & \multicolumn{1}{m{2cm}}{\includegraphics[width=2cm,height=2cm]{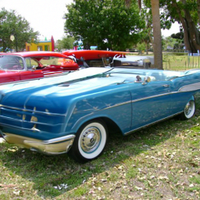}} \hspace{-4mm}
    & \multicolumn{1}{m{2cm}}{\includegraphics[width=2cm,height=2cm]{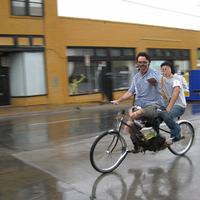}} \hspace{-4mm}
    & \multicolumn{1}{m{2cm}}{\includegraphics[width=2cm,height=2cm]{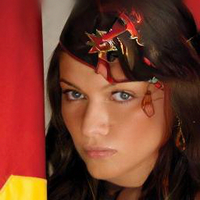}} \hspace{-4mm}
    & \multicolumn{1}{m{2cm}}{\includegraphics[width=2cm,height=2cm]{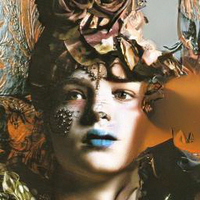}} \hspace{-4mm}
    & \multicolumn{1}{m{2cm}}{\includegraphics[width=2cm,height=2cm]{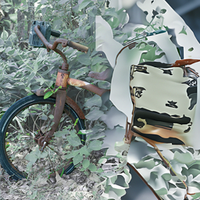}} \hspace{-4mm}
    & \multicolumn{1}{m{2cm}}{\includegraphics[width=2cm,height=2cm]{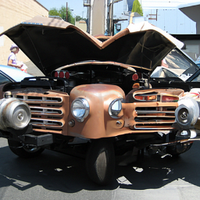}} \hspace{-4mm}
    \\
    \hspace{-5mm} \rotatebox[origin=c]{90}{\footnotesize \bf \textsc{CoPaint-TT} } \hspace{-4mm}
    & \multicolumn{1}{m{2cm}}{\includegraphics[width=2cm,height=2cm]{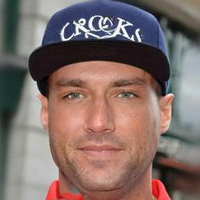}} \hspace{-4mm}
    & \multicolumn{1}{m{2cm}}{\includegraphics[width=2cm,height=2cm]{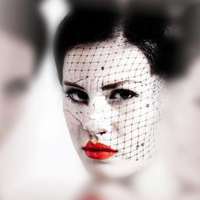}} \hspace{-4mm}
    & \multicolumn{1}{m{2cm}}{\includegraphics[width=2cm,height=2cm]{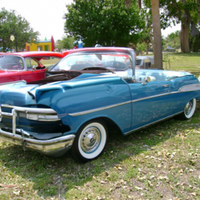}} \hspace{-4mm}
    & \multicolumn{1}{m{2cm}}{\includegraphics[width=2cm,height=2cm]{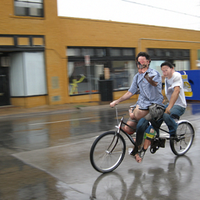}} \hspace{-4mm}
    & \multicolumn{1}{m{2cm}}{\includegraphics[width=2cm,height=2cm]{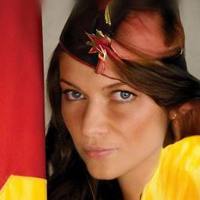}} \hspace{-4mm}
    & \multicolumn{1}{m{2cm}}{\includegraphics[width=2cm,height=2cm]{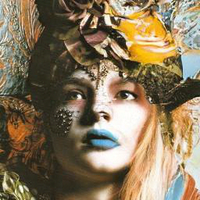}} \hspace{-4mm}
    & \multicolumn{1}{m{2cm}}{\includegraphics[width=2cm,height=2cm]{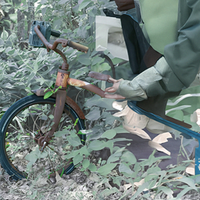}} \hspace{-4mm}
    & \multicolumn{1}{m{2cm}}{\includegraphics[width=2cm,height=2cm]{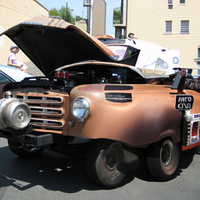}} \hspace{-4mm}
    \\
\end{tabular}}
\end{center}
\vskip -0.1in
\caption{\footnotesize Fail-cases of our method}
\label{fig:failcase}
\vskip -0.2in
\end{figure*}

We present a failure case study of our methods, \textsc{CoPaint} and \textsc{CoPaint-TT}, which can be found in Figure~\ref{fig:failcase}. 
Our findings indicate that these methods are susceptible to failure when it comes to inpainting image details. 
For instance, in the first column, while the inpainted area appears coherent and natural, the text on the hat does not blend well with the surrounding region. Other baselines exhibit similar issues. We attribute this to the deficiency of diffusion models in generating image details, particularly text, and plan to address this in future work. Additionally, we demonstrate that all methods, including ours, are likely to fail for large masked regions where the revealed surrounding information is inadequate for inpainting, resulting in unnatural images. An example of this is shown in the last column.
\vspace{-1in}

\clearpage
\section{Potential Societal Impacts}
\label{app:societal}
Despite the recent success in image generation with diffusion models, these models are prone to the biases exhibited in data~\citep{rombachHighResolution2022} and thus could generate biased images for downstream tasks. 
In line with other diffusion inpainting works, our method heavily relies on the pre-trained diffusion models and thus could exhibit or even amplify the biases existing in the models. 
For example, as shown in Figure~\ref{fig:coherence_short}, \textsc{Blended}~\citep{Song2019GenerativeMB,blended} inpaint a blond-haired woman for the reference image with a black-haired woman, which aligns with a known bias in \texttt{CelebA-HQ} dataset~\citep{Liu2021JustTT}.
The underlying reason lies in that, the replacement operation used by \textsc{Blended} only enforces the inpainting constraint on the revealed part of the generated image, while the unrevealed part is not directly modified and has to rely more on prior knowledge learned from data.
By contrast, in this paper, we introduce a Bayesian framework to jointly modify both the revealed and unrevealed parts of intermediate variables in each time step. 
This would enforce better coherence between the revealed and unrevealed parts, making our method less susceptible to biases.
As shown in Figure~\ref{fig:coherence_short}, our method \textsc{CoPaint} successfully completes the image with a black-haired woman.
On the other hand, however, due to the suboptimal greedy optimization and one-step approximation error, we note that there are still some imperfections in our method. Therefore, some bias may still persist, particularly when the revealed part contains too little information.
Besides, our method might be used in generating fake content and other malicious images to deceive humans and spread misinformation. 
In practice, our method should be appropriately used with careful checks on potential risks.

\clearpage

\end{document}